\documentclass[10pt,twocolumn,letterpaper]{article}

\usepackage[pagenumers]{cvpr} 

\newcommand{\ood}{OoD}
\newcommand{\Agg}[1]{AggS}
\newcommand{\Aggs}[1]{AggSs}
\newcommand{\OurAgg}[1]{GMM-All}

\newcommand{\mycomment}[1]{}
\definecolor{gold}{HTML}{B59410}

\usepackage{graphicx}

\usepackage{pifont}
\newcommand{\cmark}{\textcolor{OliveGreen}{\ding{51}}}  
\newcommand{\xmark}{\textcolor{lightgray}{\ding{55}}}    

\usepackage{etoolbox} 
\usepackage{array}
\usepackage[table]{xcolor} 
\usepackage{pgfplots}
\pgfplotsset{compat=1.17}
\usepackage{pgf}  
\usepackage{multirow} 
\usepackage{tabularx,ragged2e}
\usepackage{amsmath}
\usepackage{amssymb}
\usepackage{booktabs}
\usepackage{tikz} 
\usepackage{mathrsfs}

\usepackage{chngcntr}

\definecolor{cvprblue}{rgb}{0.21,0.49,0.74}
\usepackage[pagebackref,breaklinks,colorlinks,allcolors=cvprblue]{hyperref}
\usepackage{marvosym}

\def\paperID{*****} 
\def\confName{CVPR}
\def\confYear{2026}

\title{Better than Average: Spatially-Aware Aggregation of Segmentation Uncertainty Improves Downstream Performance}

\author{
  \begin{tabular}{c}
    Vanessa Emanuela Guarino$^{1,2,5,*,\text{\Letter}}$ \quad 
    Claudia Winklmayr$^{1,2,5,*}$ \quad 
    Jannik Franzen$^{1,2,3,5,*}$ \\
    Josef Lorenz Rumberger$^{1,2,3,4}$ \quad 
    Manuel Pfeuffer$^{4}$ \quad 
    Sonja Greven$^{4}$ \quad 
    Klaus Maier-Hein$^{2,6,7}$ \\
    Carsten T. Lüth$^{2,6,7,\text{\textdagger}}$ \quad 
    Christoph Karg$^{1,2,\text{\textdagger}}$ \quad
    Dagmar Kainmueller$^{1,2,5,\text{\textdagger},\text{\Letter}}$ 
  \end{tabular}
  \and
  \begin{tabular}{c}
    $^1$ Max-Delbrück-Center (MDC) \quad 
    $^2$ Helmholtz Imaging \quad
    $^3$ Charité Universitätsmedizin \\
    $^4$ Humboldt-Universität zu Berlin \quad  
    $^5$ University of Potsdam \\
    $^6$ German Cancer Research Center (DKFZ) \quad
    $^7$ Heidelberg University \\
  \end{tabular}
    \and
   \begin{tabular}{c}
   {$^\text{\Letter}$ \tt \small
   \{firstnames.surname\}@mdc-berlin.de} \quad $^{*/\text{\textdagger}}$ equal contribution 
   \end{tabular}
}

\begin{document}
\maketitle
\begin{abstract}
Uncertainty Quantification (UQ) is crucial for ensuring the reliability of automated image segmentations in safety-critical domains like biomedical image analysis or autonomous driving. In segmentation, UQ generates pixel-wise uncertainty scores that must be aggregated into image-level scores for downstream tasks like Out-of-Distribution (\ood{}) or failure detection. Despite routine use of aggregation strategies, their properties and impact on downstream task performance have not yet been comprehensively studied. Global Average is the default choice, yet it does not account for spatial and structural features of segmentation uncertainty. Alternatives like patch-, class- and threshold-based strategies exist, but lack systematic comparison, leading to inconsistent reporting and unclear best practices. We address this gap by (1) formally analyzing properties, limitations, and pitfalls of common strategies; (2) proposing novel strategies that incorporate spatial uncertainty structure and (3) benchmarking their performance on \ood{} and failure detection across ten datasets that vary in image geometry and structure. We find that aggregators leveraging spatial structure yield stronger performance in both downstream tasks studied. However, the performance of individual aggregators depends heavily on dataset characteristics, so we (4) propose a meta-aggregator that integrates multiple aggregators and performs robustly across datasets.
\end{abstract}    
\section{Introduction}
\label{sec:intro}
\begin{figure*}[ht]
    \centering
    \includegraphics[width=\textwidth]{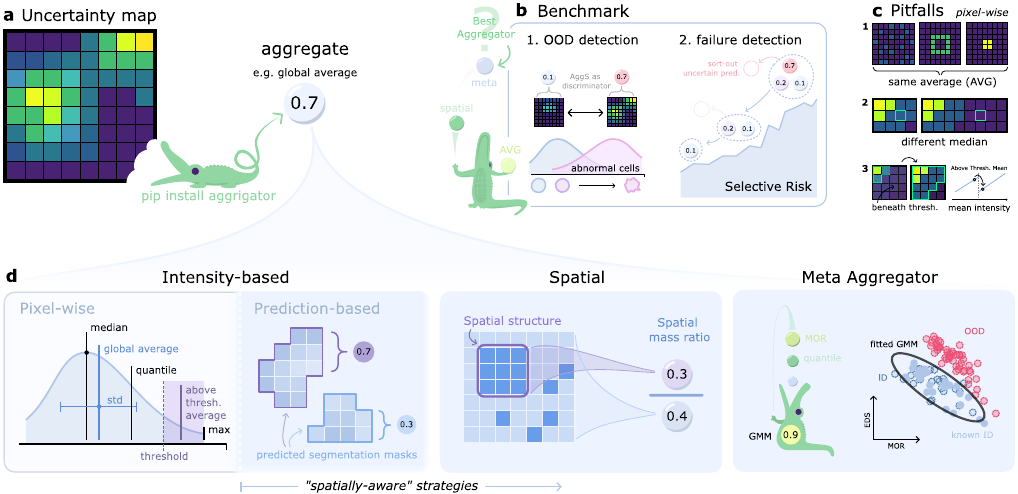}
    \caption{\textbf{Aggregation strategies (\Agg{}). (a)} An \Agg{} reduces an uncertainty map to a single scalar score.
    \textbf{(b) Empirical evaluation of \Aggs{}.} Choosing an appropriate \Agg{} is challenging and highly task-dependent. Hence, we benchmark different strategies in the context of \ood{}- and failure detection over different datasets. 
    \textbf{(c) Limitations of individual \Aggs{}.}
     Pixel-wise strategies, in particular, have key shortcomings \eg 1. AVG ignores \emph{spatial structure}; 2. AQA lacks \emph{proportional invariance}; and 3. ATA is not \emph{monotonic} (see Supp. \ref{app:formal_properties} for more details). \textbf{(d) Subtypes of \Aggs{}.} Beyond pixel-wise \Aggs{}, we also explore spatially-aware approaches—such as prediction-based and spatial \Aggs{}, which measure the fraction of uncertainty mass within structured regions of the uncertainty map. Finally, we consider Meta-\Aggs{}, which combine intensity-based and spatial strategies by fitting a Gaussian Mixture Model (GMM).}
    \label{fig:aggregation_strategies}
\end{figure*}

Reliable image segmentation is essential in safety-critical domains such as biomedicine~\cite{jungo2020analyzing, lambert2024trustworthy} and autonomous driving~\cite{kendall2017uncertainties,meyer2020learning, wang2025reliable}, where inaccurate predictions can have severe consequences. A key strategy to enhance reliability is estimating a model’s confidence for any given segmentation, enabling the detection of highly uncertain samples that may require expert review, further training, or exclusion from deployment. In segmentation, Uncertainty Quantification (UQ) methods produce pixelwise uncertainty scores \cite{kendall2017uncertainties, kohl2018probabilistic, monteiro2020stochastic}, which must be aggregated into image-level scalar values to support downstream decision-making. 
In the following, we refer to any function performing this aggregation as an \textit{Aggregation Strategy} (\Agg{}). \noindent 
Two of the most commonly performed and essential tasks requiring the aggregation of pixelwise uncertainty are \textit{Out-of-Distribution (\ood{}) Detection}, where the goal is to identify samples that fall outside the distribution of the training data~\cite{kahl2024values, zenk2025comparative}, and \textit{Failure Detection} (FD), which flags segmentations that may require manual correction or exclusion \cite{geifman2018biasreduced}. In both cases, the underlying hypothesis is that risky samples can be identified via a high aggregated uncertainty score. In practice, the aggregation of pixelwise uncertainties has received limited attention, with the global average (AVG) across all pixels being the default choice ~\cite{yang2017activelearning,ozdemir2021ALbayesianqueries,gaillochet2023medicalAL,Mittal2025realisticevaluationAL}. In the context of segmentation, however, a sample’s \ood{}-ness or susceptibility to prediction errors is often reflected in localized uncertainty patterns, for example in regions containing unseen object classes or ambiguous boundary delineations. Such local variations are prone to being overlooked when uncertainty is aggregated via simple pixelwise averaging, underscoring the need for targeted \Aggs{} that account for image context or spatial structure. \newline\noindent
Although multiple studies in object detection, medical imaging, and autonomous driving have highlighted the importance of targeted \Agg{}s, most proposed methods are narrowly tailored to specific tasks, lack formal theoretical support, and frequently rely on intermediate model outputs or ad hoc definitions of “relevance”, limiting their generalizability ~\cite{pocevivciute2024benefits,jungo2020analyzing,prasanna2024perception,valiuddin2024review}. To date, a systematic evaluation of how different \Agg{}s impact downstream performance is still missing, and so are respective best practices. \newline\noindent 
To address the lack of systematic studies on uncertainty aggregation, we first identify the most commonly used \Aggs{} in the literature. As none of these methods explicitly account for the spatial distribution of uncertainty, we subsequently propose three novel, spatially informed \Aggs{}, grounded in theoretically well-established measures for analyzing spatial relationships in grayscale images, including Moran’s I \cite{Moran1948TheIO}, Shannon Entropy \cite{shannon}, and the Edge Density score \cite{Pereira2014Edge}. Wherever possible, we discuss theoretical properties of the \Aggs{} and then empirically benchmark their behavior in \ood{} and Failure Detection. The benchmark comprises 10 datasets spanning diverse modalities, including street scenes, medical imaging, and synthetic data, covering both grayscale and multi-channel inputs, and encompassing semantic and instance segmentation tasks. \newline\noindent
Our evaluation confirms that global averaging is suboptimal. More broadly, we find that no single \Agg{} consistently outperforms others across all experimental settings; rather, the effectiveness of an \Agg{} depends strongly on the dataset and task. To address this limitation, we propose a meta-aggregation strategy (\OurAgg{}), which combines multiple individual \Aggs{} and demonstrates robust performance across diverse scenarios, making it a reliable default choice when there is insufficient knowledge about the relevant uncertainty characteristics in \ood{}- or potential failure-cases to make an informed choice. \newline\noindent 
In summary, our work provides the first systematic comparison of \Agg{}s and establishes a foundation for optimizing downstream task performance through principled uncertainty aggregation in segmentation. \newline\noindent 
\noindent To summarize our \textbf{Contributions:} 
\begin{itemize}  
\item We extensively benchmark the behavior of common \Aggs{} on a wide variety of practically relevant datasets.
\item We introduce spatial \Aggs{} that are able to capture the spatial patterns of uncertainty estimates.
\item We propose  \OurAgg{}, a new all-purpose \Agg{} combining  intensity- and spatial aspects of uncertainty maps. \OurAgg{} performs competitively on  downstream tasks, outperforming baselines for \ood{}- and failure detection on many datasets. 
\item We provide an open-source Python library for uncertainty aggregation: \url{https://github.com/Kainmueller-Lab/aggrigator}. 
\end{itemize}

\section{Related Work}

\paragraph{Pixel-score Uncertainty Quantification}
Segmentation models take an input image \( I \) of size $m\times n$ and predict for each pixel a probability distribution \( p(c) \) over a set of classes \( c \in \{1, \dots, K\} \). Assigning each pixel to its most probable class yields a segmentation mask \( M \in \{1, \dots, K\}^{m \times n} \). Depending on the task, classes may represent semantic categories (\eg, road, building, person) or distinguish individual object instances, as in \emph{semantic}, \emph{instance}, or \emph{panoptic} segmentation.
\noindent UQ methods for segmentation \cite{valiuddin2024review, gawlikowski2023survey} typically assign an uncertainty score to each pixel which reflects the model's confidence in the predicted class or instance, resulting in an \emph{uncertainty map} $U$.
%


\noindent\textbf{Pixel-score Out-of-Distribution Detection} 
\ood{} detection identifies inputs that deviate from a model's training distribution, relying on the assumption that models should express low confidence on unfamiliar samples. Many approaches for \ood{} detection were first developed for classification with Maximum Softmax Probability (MSP) as the initial measure of unreliability \cite{hendrycks2017a}. This was extended by sample-based Bayesian methods leveraging predictive uncertainty \cite{gal2016dropout, lakshminarayanan2017simple, osawa2019practical}, forming the dominant output-based paradigm later applied to segmentation \cite{kendall2017uncertainties, BMVC2017_57, lakshminarayanan2017simple, hendrycks2017a, mukhoti2018evaluating}. Other approaches include distance-based metrics in latent space \cite{lee2018simple, sun2022knnood}, density-based likelihood estimation \cite{Ancha2024, Mukhoti_2023_CVPR, postels2021, arnez2024latent}, and reconstruction errors from auxiliary encoder-decoder models \cite{xia2020synthesize, denouden2018improving}, mostly suitable for anomaly segmentation \cite{chan2021segmentmeifyoucan}. Our framework combines \textit{output}- and \textit{density-based} concepts, operating \textit{post-hoc} on features derived from pixelwise uncertainty maps. This makes it model-agnostic, applicable to alternative uncertainty representations such as reconstruction-based generative model ~\cite{chan2021pixelwise}, without relying on latent spaces or specialized training schemes~\cite{lee2018simple, Mukhoti_2023_CVPR, arnez2024latent}. 
\newline 
\noindent \textbf{Pixel-score Failure Detection} FD assigns a confidence score per image (\iid{} or \ood{}) to flag potential segmentation errors, and when this score is used to abstain or retain predictions, it constitutes selective classification (SC). Standard baselines rely on MSP, refined via thresholding for user-specified risk levels \cite{geifman2017selective}. Classification-based confidence methods have also been adapted to FD, \eg, using ensembles \cite{rabanser2023training} or joint prediction–rejection training \cite{huang2020selfadaptative}. Unlike classification, however, ideal confidence functions for segmentation remain less established. While existing image-level FD methods primarily focus on segmentation quality estimation \cite{kohlberger2012evaluating, wang2020deepMRI} or distribution shift detection, the SC literature has historically emphasized evaluation metrics, improving the accuracy-coverage tradeoff \cite{geifman2017selective} with recent extensions to selective accuracy across all coverage levels \cite{traub2024}. A newer trend empirically adapts classification-based scores to generate \textit{pixel}-level or \textit{component}-level uncertainty estimates \cite{nair2020exploring, molchanova2023novel}, yet partial rejections offer limited benefit in practice, as specialists must then review the image. While some works evaluate subsets of \Agg{}s in biomedical settings \cite{jungo2020analyzing, zenk2025comparative}, our work is the first to systematically compare \Aggs{} and introduce a meta-strategy that enables the effective empirical use of SC.



\section{Methodology}

\subsection{Common aggregation strategies}
\label{sec:common_agg}
An \emph{aggregation strategy} $f$ summarizes a 2D uncertainty map \( U \) into a single scalar \( f(U) \in\mathbb{R}\).
Common \Aggs{} are \emph{intensity-based}, quantifying the magnitude of uncertainty in \( U \), typically through averaging. 
Among these, \emph{pixelwise} methods utilize only the uncertainty values, while \emph{prediction-based} methods leverage segmentation outputs to compute class-aware uncertainty scores, allowing for better sensitivity to localized uncertainty. This approach also applies to other contextual inputs such as ground truth labels, error maps, or depth cues. Pixelwise \Aggs{} are among the most widely used approaches due to their simplicity and ease of implementation \cite{kahl2024values, luo2024uncertaintyguidedtieredselftrainingframework,yang2017activelearning,pocevivciute2024benefits}.
\noindent
\textit{Global Average} (AVG) is the average score taken across all pixelwise uncertainty scores $u_{i}$. \textit{Patch-Level Maximum Average} (PLM) operates by sliding a fixed-size patch across the image, computing the average pixel intensity within each patch, and subsequently taking the maximum value over all patch positions.
\textit{Above-Threshold Average} (ATA) is defined as the average across all $u_{i}> T$ with a fixed threshold $T>0$. If no values surpass $T$, the aggregated score is defined as zero. \textit{Above-Quantile Average} (AQA) is the average across all uncertainty values above a fixed $q$-quantile for $q\in(0,1)$. In other words, it averages the top $1-q$ values, independently of magnitude.
\\
Using a predicted segmentation mask $M \in \{1,2,...K\}^{m\times n}$ a class-level average $\alpha_c$ can be calculated  as the average uncertainty score taken across all pixels assigned to class $c$ by the mask $M$.
For a weight vector $w\in[0,1]^K$ satisfying $\sum_{c=1}^{K}  w_c = 1$ we further define the \textit{Weighted Class Average} as $\sum_{c=1}^{K} w_c\alpha_c$. Note, that all class-level averages studied here exclude the background-class. For specific choices of $w$ we obtain the following common prediction-based \Agg{} \cite{Wang_2020,Camarasa_2021}:

\noindent
\textit{Balanced Class Average} (BCA) weighs each class average equally, i.e. by setting all weights $w_c:=1/K$. \textit{Imbalanced Class Average} (ICA) weighs each class score by the relative proportion of the class, i.e. by setting all weights $w_c:=A_c / \sum_{c=1}^{K} A_c$, where $A_c$ is the number of pixels assigned to class $c$. \textit{Above-Quantile Average for Foreground Ratio} (QFR)  is defined as the average uncertainty among the top $q_{FG}$ uncertain pixels. It uses the predicted segmentation mask to calculate the foreground (FG) ratio $q_{FG}=\frac{\text{number of FG pixels}}{\text{total number of pixels}}$.
This ensures that the average is computed over the most uncertain pixels, in a proportion that matches the relative size of the foreground \cite{kahl2024values}. 

\subsection{Pitfalls of common Aggregation Strategies}
Intensity-based \Aggs{} are susceptible to the following pitfalls
as illustrated in \Cref{fig:aggregation_strategies}c:
(1) \Aggs{} that are not \emph{sensitive to spatial structure} assign identical scores to distinct spatial patterns, such as uniform low-uncertainty versus compact regions of high uncertainty. This includes all averaging \Aggs{}, notably AVG, which are therefore unsuitable for detecting small but relevant uncertain regions or uncertainty concentrated along object boundaries.
(2) \Aggs{} lacking \emph{proportion invariance} yield scores that vary with the relative amount of low-uncertainty background. For example, AQA produces higher aggregated uncertainty when background pixels are cropped away, even though the actual high-uncertainty regions remain unchanged.
(3) \Aggs{} which are not \textit{monotonic}, as is the case for ATA, fail to capture a global increase of pixelwise uncertainty values, and more importantly, may even \emph{reduce} the resulting score.
We provide formal definitions and a deeper analysis of these properties as well as a detailed discussion of the respective pitfalls in Supp. \ref{app:formal_properties}.


\subsection{Spatial aggregation strategies}
\label{sec:spatial_agg}

In order to address the lack of spatial awareness in intensity-based \Aggs{} and explicitly study spatial patterns in uncertainty maps we draw on the following established methods designed to capture structural patterns such as clustering, noise, or edges in grayscale images. 
\emph{Moran's I} \cite{Moran1948TheIO} measures spatial autocorrelation by quantifying how similar intensity values are in a local neighborhood;
\emph{Edge Density Score} \cite{Pereira2014Edge} captures the presence of spatial boundaries by computing the density of edges in a local window;
\emph{(Shannon) Entropy} \cite{shannon} reflects the local randomness or heterogeneity of intensity values.

\noindent
We propose to leverage spatial measures as "aggregators" of uncertainty by calculating a \emph{spatial mass ratio (SMR)} i.e. the fraction of total uncertainty mass that is concentrated in regions with high local spatial structure. SMR is determined by weighting the uncertainty map with pixelwise spatial scores computed from a sliding window and dividing the average uncertainty in regions with high spatial measure by the overall average uncertainty, resulting in an interpretable score between $0$ and $1$ (see Supp. \ref{sec:spatial_mass_ratio}): 
\\
$\text{SMR}_{\text{Moran}}$ (or MOR) equals 0 if all uncertainty lies in noise-like regions, and 1 if it is entirely concentrated in clusters.
\\
$\text{SMR}_{\text{EDS}}$ (or EDS) equals 0 if all uncertainty lies in flat regions, and 1 if it is entirely concentrated along edges.
\\
$\text{SMR}_{\text{Entropy}}$ (or ENT) equals 0 if all uncertainty lies in regions of almost constant uncertainty, and 1 if it is fully concentrated in areas with high variability.
See \Cref{fig:datasets} for an illustration  of how MOR and EDS capture different spatial uncertainty patterns across dataset samples.

\subsection{Meta-aggregation}
\label{sec:meta_agg}

To unify spatial and intensity-based \Aggs{} and faithfully capture heterogeneous aspects of uncertainty maps, we model aggregated scores resulting from different \Aggs{}
as descriptive features of an uncertainty map $U$.
Specifically, given a set of \( d \) aggregation functions \( f_1, \dots, f_d \), we represent each uncertainty map \( U \) as a feature vector \( f_U = (f_1(U), \dots, f_d(U)) \in \mathbb{R}^d \), where $f_i(U)$ represents a single aggregated score.

\noindent
The key idea is that \emph{in-distribution (\iid{})} samples occupy a specific region
in this feature space, and uncertainty maps whose feature vectors lie outside this region are likely to be \ood{} or failures (see \Cref{fig:aggregation_strategies}).

\noindent
We assume that the \iid{} feature vectors follow a multivariate distribution which we describe 
with a Gaussian Mixture Model (GMM) \cite{dempster1977maximum}. We use a subset of known \iid{} uncertainty maps \( \mathcal{U}_{\text{\iid{}}}^{fit} \) (separate from the set for evaluation) to fit and estimate the optimal number of GMM-modes based on the Bayesian Information Criterion \cite{schwarz1978estimating} leading to our final density function \( p_{\text{GMM}}(u) \) (see Supp. \ref{sec:supplement_gmm} for details). 
We then define our \emph{meta-aggregation strategy} as the negative log-likelihood (NLL) of the features:
\[
f_{\text{meta}}(U; p_{\text{GMM}}) := -\ln p_{\text{GMM}}(f_U).
\]
The definition of the meta-aggregator depends on the set of \Aggs{} used as input features. We consider three main variants: (a) \emph{GMM-Spa}, which includes only spatial measures; (b) \emph{GMM-Int}, based solely on intensity-based strategies; and (c) \emph{GMM-All}, which combines both spatial and intensity-based features.
For a detailed discussion of the selection of strategies and the corresponding ablation studies, see Supp.~\ref{sec:supplement_gmm}.

\section{Experimental Setups \& Results}
\subsection{Implementation Details}
We generate segmentations across diverse datasets using a range of network architecture, including U-Net 3D, U-Net, HoVer-NeXt, HRNet and DeepLabv3+ (for details on model training, see Supp.~\ref{app:data}). To estimate pixel-wise uncertainty, we focus on the widely used and computationally efficient Monte Carlo Dropout~(MCD)~\cite{gal2016dropout}. By performing stochastic forward passes with dropout activated during inference, MCD captures epistemic uncertainty, which is particularly relevant for downstream tasks such as \ood{} detection and FD \cite{kendall2017uncertainties}. Furthermore, we explicitly normalize all uncertainty maps such that $U \in [0,1]^{m \times n}$ (cf.  Supp.~\ref{app:supp_uq_maps}).

\noindent While alternative UQ methods or segmentation backbones may yield different uncertainty distributions, the aggregation behavior we investigate remains largely unaffected. To support this claim, we replicate our experiments using additional UQ methods, using other UQ methods, namely MSP, Deep Ensembles \cite{lakshminarayanan2017simple} and Test Time Augmentations \cite{moshkov2020test}. The corresponding results are reported in Supp.~\ref{app:mean_rank_tables}, where we observe consistent trends across methods.

\subsection{Datasets}

\begin{figure}[t]
    \centering
    \includegraphics[width=0.99\linewidth]{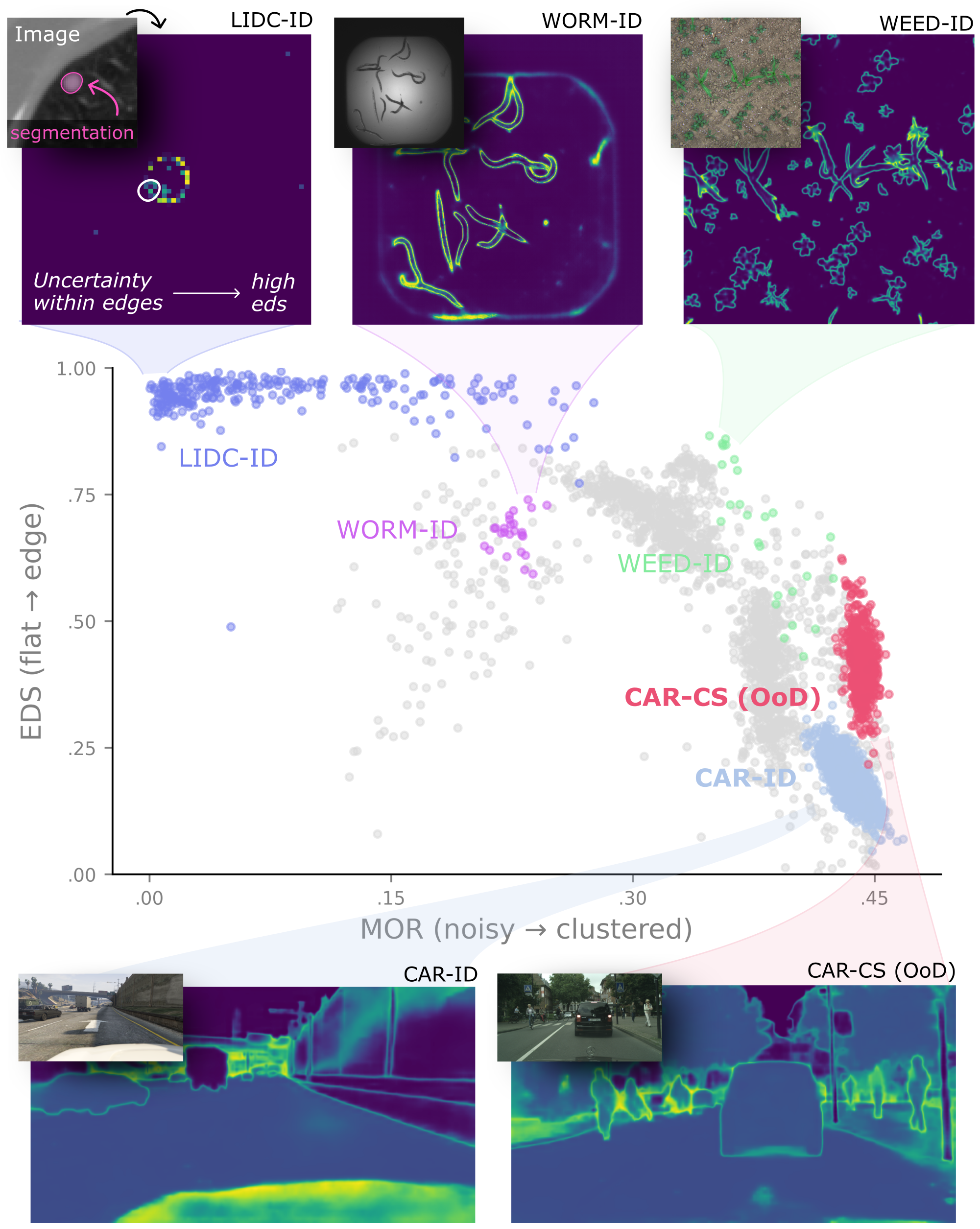}
    \caption{
    \textbf{Structural diversity of datasets.}
    The scatterplot illustrates structural diversity by projecting uncertainty maps into the space of two SMR scores: MOR and EDS. High MOR indicates clustered uncertainty (low: noise), while high EDS reflects edge-localized uncertainty (low: flat regions). Notably, EDS effectively identifies \ood{} samples in CAR-ID/CAR-CS.
    }
    \label{fig:datasets}
\end{figure}

In order to assess whether \Aggs{} capture relevant aspects of uncertainty, we require not only structurally diverse uncertainty maps but also multiple \ood{} settings encompassing both covariate and semantic shifts \cite{MORENOTORRES2012521}).
We briefly summarize the tasks, datasets, and \iid{}/\ood{} setups below; full details are provided in Supp.~\ref{app:data}. \newline \noindent
\textbf{Segmentation of nuclei in pathology images:} We use the synthetic Arctique (ARC) histopathology dataset~\cite{franzen2024arctique} as \iid{} data and derive two perturbed variants as \ood{}: (1) ARC-Nuc, where nuclei intensities are reduced (analyzed for instance segmentation), and (2) ARC-BC, where the red blood cell count is increased (analyzed for semantic segmentation). In ARC-Nuc, uncertainty is expected around the boundaries of "hard-to-identify" nuclei; in ARC-BC, it is concentrated in localized regions (“blobs”) where red blood cells may be semantically mistaken for eosinophils. \newline 
We also use the Lizard dataset~\cite{graham2021lizard} taking data from five medical centers as \iid{} (LIZ) and from a sixth center as \ood{} (LIZ-G). We evaluate both semantic and instance segmentation, denoting the \ood{} sets as LIZ-SG (semantic) and LIZ-IG (instance). Here, the shift arises from variations in the recording protocol,1 affecting tissue appearance and increasing overall object uncertainty. \newline \noindent
\textbf{Binary segmentation of lung nodules in CT volumes:} 
Following~\cite{kahl2024values}, the LIDC dataset~\cite{armato2011lung} is partitioned into subsets based on (1)  textural changes in lung nodules (LIDC-Tex) and (2) high-grade malignant nodules (LIDC-Mal), both used as \ood{} data. The task involves binary segmentation of tumors versus background. 
Both \ood{} shifts increase intra-tumoral uncertainty due to texture variations; additionally, larger malignant tumors increase the surface area of elevated uncertainty in the 2D slices. \newline\noindent
%
\textbf{Binary segmentation of microorganisms in microscopy images:} The C. Elegans live and dead assay~\cite{ljosa2012annotated} (WORM) is used as the \iid{} dataset. Two subsets of the SinfNet dataset~\cite{Sabban_SinfNet_Microorganism_image_2023}, depicting different breeds of microorganisms, are used as \ood{} data: (1) the Protist subset (WORM-Pro) and (2) the Nematode subset (WORM-Nem). The task is foreground–background segmentation, where the \ood{} datasets contain individual microorganisms whose morphological shapes differ substantially from C. Elegans, leading to localized, clustered regions of elevated uncertainty. \newline \noindent
\textbf{Semantic segmentation of urban street scenes:} The synthetic GTA dataset~\cite{Richter_2016_ECCV}, derived from video game footage, is used as the \iid{} dataset (CAR-ID), and the Cityscapes dataset~\cite{Cordts2016Cityscapes}, consisting of real-world urban scenes filmed from a car’s perspective, as the \ood{} dataset (CAR-CS). The task is semantic segmentation, where the transition from synthetic to real-world imagery leads to an overall increase in boundary uncertainty, with uncertainty in CAR-CS appearing more spatially diffuse. Notably, in the CAR-ID data, total uncertainty is high yet localized, particularly around unlabeled or underrepresented classes (\eg the front of the ego vehicle), emphasizing the necessity of spatially informed aggregation (cf. \cref{fig:datasets}). \newline \noindent 
\textbf{Semantic segmentation of multispectral crop images:}
We use the Weedsgalore dataset~\cite{Celikkan_2025_WACV} (WEED), containing drone images of maize and several weed species, as \iid{} data. The Crop and Weed dataset~\cite{steininger2023cropandweed}, captured under diverse soil conditions with handheld cameras, serves as the \ood{} dataset (WEED-Hand). The task is 3-class semantic segmentation (crops, weeds, background). The \ood{} set includes additional weed species with previously unseen shapes and textures, leading to elevated uncertainty in regions with unfamiliar morphologies. \newline \noindent
\textbf{Dataset Diversity according to Spatial Measures\,}
In \Cref{fig:datasets}, we show the aggregated scores of uncertainty maps projected into the space defined by the spatial \Aggs{} EDS and MOR. Scores from the same dataset are highlighted in color, illustrating that they occupy distinct regions in feature space. In this visualization, CAR-ID and CAR-CS are clearly separable along the EDS dimension.
The example images reveal the underlying cause: the unfamiliar CAR-CS images produce noticeably “blurrier” uncertainty patterns, resulting in lower EDS values. Additional examples 
can be found in Supp.~\ref{app:data}. 

\subsection{On \ood{} Detection Performance} \label{sec:ood_res}

\begin{figure}[t]
    \centering
    \includegraphics[width=\linewidth]{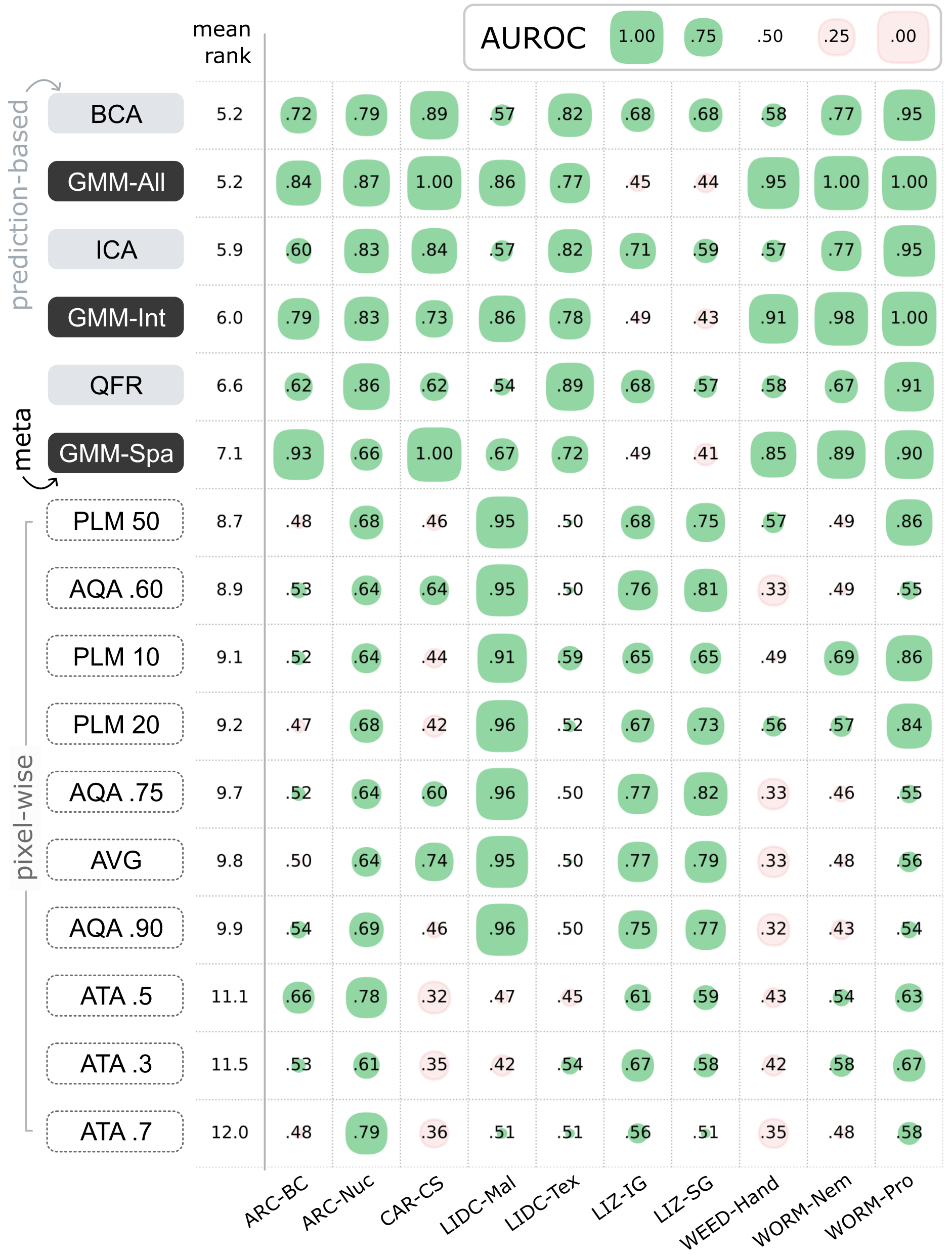}
    \caption{
    \textbf{Performance on \ood{} Detection.} Higher AUROC (0.6–1.0) indicates better \iid{}–\ood{} separation. \Aggs{} are ranked by their mean AUROC, computed over 500 bootstrap samples per dataset and then averaged across datasets for stability. Ranking robustness is assessed via one-sided Wilcoxon signed-rank tests at 5\%, showing that \textit{prediction-based} BCA and ICA, along with GMM-based methods, form a statistically dominant tier ($p < 0.05$). Numbers after \Agg{} labels indicate method-specific parameters (cf. Supp.~\ref{app:aggs-params}). Full analysis, including additional UQ methods and confidence intervals, is in Supp.~\ref{app:mean_rank_tables}.
    }
    \label{fig:ood_results}
\end{figure}

\begin{figure*}[t]
    \includegraphics[width=0.945\textwidth]{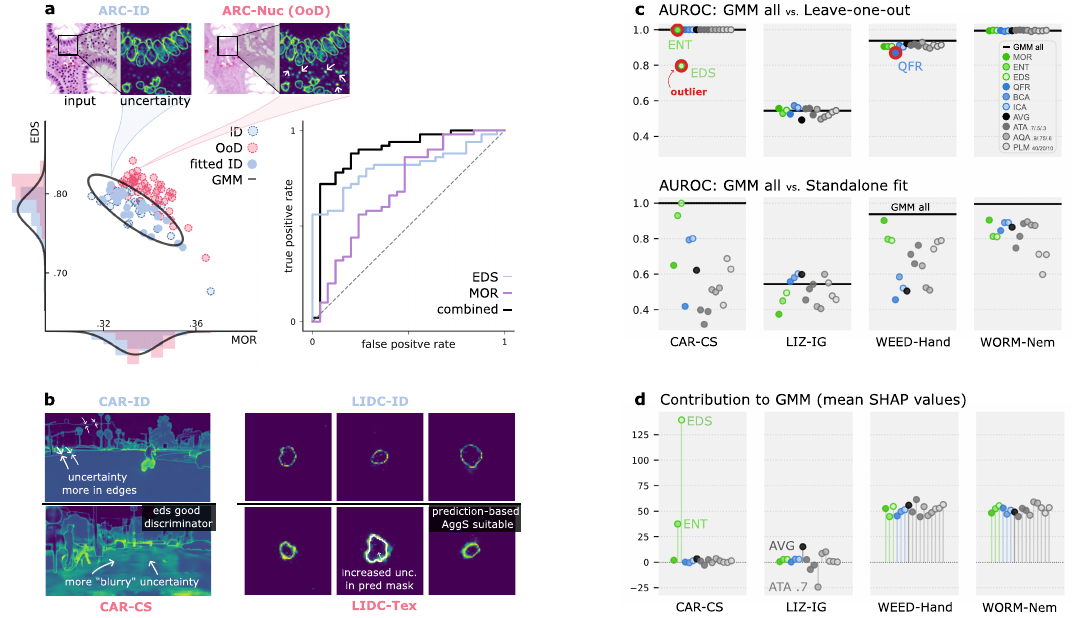}
    \caption{\textbf{GMM Robustness for \ood{} Detection.} (a) Fitting a GMM enables spatial methods to act as \Aggs{}, capturing subtle non-linear changes in uncertainty maps. (b) Qualitative examples showing that effective \Aggs{} for \iid{}–\ood{} separation are data-dependent. (c) \textbf{\emph{GMM-All} AUROC: leave-one-out (top) and individual fitting (bottom) on a subset (CAR-CS, LIZ-IG, WEED-Hand, and WORM-Nem split).} Using all \Aggs{} generally matches or outperforms individual ones in 3 cases, with minimal impact from removing specific ones. Exceptions occur when a feature dominates (\eg, EDS for CAR-CS, marked as Tukey outlier) or when features lack discriminative power. (d) \textbf{\emph{GMM-All} AUROC: absolute SHAP values.} Positive values indicate that an \Agg{} aids \emph{GMM-All} in separating \iid{}–\ood{}; bi-directional or null contributions reduce performance (e.g., in LIZ-IG). SHAP values are shown for the same subsets as in (c).}   
    \label{fig:ood_qual}
\end{figure*}

For an \Agg{} to be effective in \ood{} detection, it should assign higher aggregated uncertainty to \ood{} samples than to \iid{} samples. Following \cite{hendrycks2017a}, we construct balanced sets of \iid{} and \ood{} samples and evaluate performance using the Area Under the Receiver Operating Characteristic curve (AUROC)~\cite{davis2006relationship} (cf. Supp.~\ref{app:supplement_downstream_tasks} for further details). 
\Cref{fig:ood_results} summarizes our benchmark results, with \Aggs{} ranked by their mean performance computed over 500 bootstrap samples from each test dataset, to account for variability in the evaluation data.
Top performers by average rank are \textit{prediction-based} BCA, ICA, and our \emph{GMM-All}, supporting our claim that it offers robust performance across datasets when no single \Agg{} is clearly preferred. Ranking robustness is further assessed via one-sided Wilcoxon signed-rank tests at 5\%, confirming that while no \Agg{} is universally dominant, BCA, ICA, and \emph{GMM-All} form a statistically significant top tier ($p < 0.05$) \cite{wilcoxon1945individual}. Full confidence analyses are provided in Supp.~\ref{app:mean_rank_tables}.
\newline\noindent
In 6 of 10 scenarios, AVG performs poorly, approaching random guessing. Notably, it excels on LIDC-Mal, where larger malignant nodules and elevated intra-nodule uncertainty favor prediction-unaware aggregators, and performs competitively on LIZ-IG and LIZ-SG, due to high background noise and dense ring-like structures in the \ood{} scene, as well as on CAR-CS, where the presence of semantic classes not learned from in CAR-ID leads to increase in the total uncertainty. 
Although AQA and PLM often correlate with AVG, PLM outperforms the latter in WORM-Pro, where clustered uncertainty introduced by protists dominates the background. Finally, ATA methods consistently underperform, ranking last on average, possibly due to their \textit{non-monotonic} relationship with increased pixelwise uncertainty. Notably, \emph{GMM-All}, \emph{GMM-Int}, and \emph{GMM-Spa} struggle with \iid{}/\ood{} separation in LIZ-IG and LIZ-SG, though marginal-based separation succeed. We hypothesize that similar marginal feature performance prevents the GMM from finding a clear separation in high-dimensional space.
\newline\noindent To further evaluate robustness, \Cref{fig:ood_qual}c–d compares the AUROC of \emph{GMM-All} (flat black line) against leave-one-out variants (top panel) and fits on individual \Aggs{} (bottom panel) across CAR-CS, LIZ-IG, WEED-Hand and WORM-Nem (cf. Supp.~\ref{app:gmm-ablations} for details and results). Using all \Aggs{} generally matches or outperforms individual features in 3 of 4 cases, with minimal impact from removing specific ones. Exceptions occur when a feature dominates (\eg, EDS for CAR-CS, marked as Tukey outlier ~\cite{tukey1977exploratory}). \Cref{fig:ood_qual}d complements the ablations by showing SHAP values \cite{NIPS2017_7062}, averaged across samples (i.e., absolute), computed from the NLL values of \emph{GMM-All} fitted on all \Aggs{}: while the \emph{GMM-All}’s ability to detect \ood{} samples is largely driven by EDS in CAR, all \Aggs{} contribute positively for WEED and WORM. As hypothesized, in LIZ-IG, \emph{GMM-All} fails to find a clear separation for some \Aggs{}, and even when separation exists, feature variability across \iid{} and \ood{} samples prevents concentrated, unidirectional contributions.




\begin{figure*}[t]
    \centering
    \includegraphics[width=0.95\textwidth]{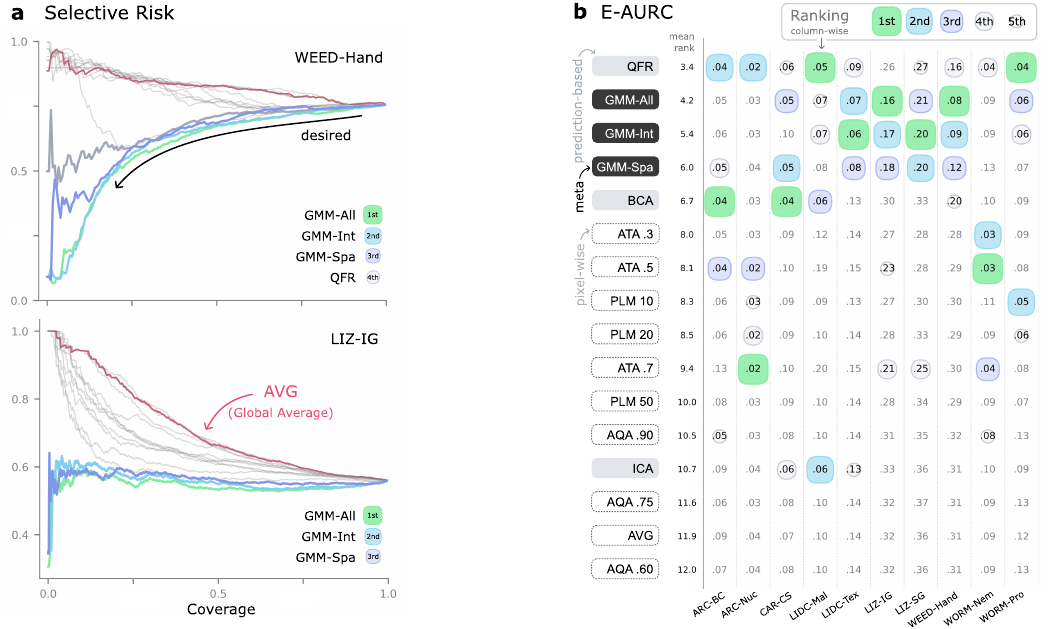}
    \caption{
    \textbf{Performance on FD.}
    (a) \textbf{Exemplary Selective Risk-Coverage curves}. 
    (b) \textbf{E-AURC scores}. Lower values indicate better alignment between uncertainty and prediction errors. \Aggs{} are ranked by their mean E-AURC, computed over 500 bootstrap samples per dataset and then averaged across datasets for stability. Ranking robustness is assessed via one-sided Wilcoxon signed-rank tests at 5\%, showing that \textit{prediction-based} QFR shows statistically significant improvement over all other \Aggs{} ($p < 0.001$), followed by BCA and GMM-based methods ($p < 0.05$). Full analysis, including additional UQ methods and confidence intervals, is in Supp. \ref{app:mean_rank_tables}.
    }
    \label{fig:fd_results}
\end{figure*}

\subsection{On Failure Detection Performance}
\label{sec:fd_res}


We evaluate FD using SC metrics, discarding predictions based on aggregated uncertainty and measuring the resulting missegmentation risk with a task-specific metric ~\cite{geifman2018biasreduced}. Following~\citet{kahl2024values}, we adopt $(1 - \textnormal{Dice})$, analogous to the misclassification rate in classification. The Selective Risk–Coverage curve relates the fraction of retained predictions (coverage) to the corresponding selective risk. Since we compare \Aggs{} only, keeping the model and absolute risk fixed, we assess reliability using the Excess Area Under the Risk–Coverage Curve (E-AURC, ~\cite{geifman2018biasreduced, jaeger2023reflectevaluationpracticesfailure}), thereby focusing on ranking quality (cf. Supp.~\ref{app:supplement_downstream_tasks} for its derivation). \newline\noindent 
\Cref{fig:fd_results} shows selected risk-curves together with the mean rank performance. As for \ood{} Detection, the GMM is fitted on \iid{} features and we use the same bootstrapping procedure for statistical robustness (cf. Supp. \cref{app:mean_rank_tables}). 
Evaluation of all \Aggs{} is performed on \ood{} and \iid{} data combined, for conciseness we refer to the combined dataset by the \ood{} label. 
The difference between the exemplary selective risk-coverage curves in \Cref{fig:fd_results}a is striking: on LIZ, segmentation suffers from \textit{silent failures}, i.e. samples confidently misclassified, but \emph{GMM-All}, \emph{GMM-Spa}, and \emph{GMM-Int} mitigate this by filtering uncertain samples without raising misclassification risk. On WEED-Hand, these scores minimize E-AURC by effectively removing the most uncertain, error-prone predictions. \newline\noindent
The curves also highlight AVG’s generally poor performance, ranking lowest (see Supp. ~\ref{app:mean_rank_tables} for mean rank tables and detailed comparisons) due to underestimating uncertainty on fully misclassified samples, except for synthetic datasets like CAR and ARC, where silent failures are rare and misclassified instances mostly correspond to \ood{} samples. 
In contrast, although ATA ranks low for \ood{} detection, it performs well for FD, particularly on WORM-Nem and ARC-Nuc, where segmentation errors concentrate around high-uncertainty object boundaries, allowing even low-threshold ATA to align closely with the error.
Across datasets, \emph{GMM-All}, \emph{GMM-Spa}, and \emph{GMM-Int} scores perform comparably to the top-ranked QFR, which surpasses BCA and ICA by applying an object-size–independent threshold that targets the most uncertain regions (typically object boundaries), achieving the highest statistical performance ($p < 0.01$, Wilcoxon test; cf. Supp. \ref{app:mean_rank_tables}).
ICA's poor performance is due to segmentation errors which frequently arise in small objects.
Notably, E-AURC rankings for all GMM-based scores invert relative to AUROC on LIZ-IG, LIZ-SG, and WORM-Nem. This reversal occurs because samples on the outer GMM iso-density contours correspond to high segmentation errors but not to \ood{} samples, affecting E-AURC but not AUROC. Similar to ARC, this drop is mainly due to \emph{GMM-Spa}’s wide support (see WORM's SMRs in \Cref{fig:datasets}), which weakens the correlation between distance to the mode and segmentation error.

\subsection{On Overall Downstream Performance}\label{sec:overall-res}
Overall, we observe substantial variability in the performance of different \Aggs{}, confirming that the choice of \Agg{} strongly depends on both the dataset and the task. \textit{Prediction-based} aggregators consistently rank among the top performers, underscoring the importance of the \textit{proportion-invariance} property that they satisfy. Our proposed \emph{GMM-All}, \emph{GMM-Spa}, and \emph{GMM-Int} also perform strongly, affirming
their role as robust general aggregation strategies. Overall, their performance is highly correlated across datasets. 
\emph{GMM-All} outperforms its ablated versions (\emph{GMM-Spa} and \emph{GMM-Int}) in both tasks, confirming that combining intensity and structural-based features of uncertainty maps enhances performance. Conversely, pixel-wise \Aggs{} such as AVG, AQA, PLM, and ATA show less stable performance across both tasks and datasets. Notably, AVG, the baseline, ranks among the lowest-performing methods for both \ood{} detection and FD, which supports the claim that it should not be the default choice for aggregation.

\section{Conclusion}
Our results indicate that the effectiveness of intensity-based aggregators strongly depends on dataset properties such as object size, structure, and foreground proportion, and that the commonly used AVG can obscure critical information, making it a suboptimal default for both \ood{} detection and FD. Aggregators that explicitly incorporate spatial features consistently outperform  by preserving structural cues that intensity-only methods would otherwise discard. Consequently, selecting an \Agg{} that reflects spatial and structural characteristics is crucial. When this is not feasible, our meta-aggregation strategy offers a robust alternative by combining diverse aggregators and leveraging spatial structure to capture a richer set of uncertainty features. This approach, however, assumes that the aggregated scores follow a suitable distribution (\eg a GMM) and requires a sufficiently large and representative \iid{} set to ensure stable performance (See Supp.~\ref{app:limitations} for a detailed discussion).

\section*{Acknowledgements}
We would like to express our gratitude to Jeremias Traub for helpful discussion and insights regarding failure detection metrics and to Kim-Celine Kahl for her support with the LIDC/LIDC-Mal \& -Tex and CAR-ID/CAR-CS experiments.
This work was funded by the German Research Foundation (DFG) Research Training Group CompCancer (RTG2424), DFG Research Unit DeSBi (KI-FOR 5363, project no.\ 459422098), DFG Collaborative Research Center FONDA (SFB 1404, project no.\ 414984028), DFG Individual Research Grant UMDISTO (project no.\ 498181230), Synergy Unit of the Helmholtz Foundation Model Initiative, and supported by the Helmholtz Einstein International Berlin Research School In Data Science (HEIBRiDS).

\section*{Contribution Statement}
VEG, CW, JF, CK, CL, and DK collaboratively conceptualized the research question, designed the methodology and the principal experiments. VEG conducted the experiments on OoD-detection and FD and handled statistical evaluation. CW collected and defined aggregation strategies and devised spatial measures. JF co-prototyped and ablated the meta-aggregation strategy, led result reporting and visualization, and released the package. CK analyzed key properties and pitfalls of aggregation strategies and developed the aggregation code-base. VEG, CW, and CK collected datasets and evaluated pixelwise uncertainty maps. CL provided code and data for the CAR and LIDC datasets. MP and SG provided valuable feedback on the design of the GMM and the statistical evaluation. JLR contributed to the code and prepared the repository for publication. VEG, CW, JF, JLR, CK, CL and DK contributed to manuscript preparation. VEG, CK, CL, and DK supervised the project and KMH provided overall guidance. All authors reviewed and approved the final version of the manuscript.

{
    \small

}

\clearpage
\setcounter{section}{0}

\renewcommand{\thesection}{\Alph{section}}
\renewcommand{\thefigure}{\Alph{section}\arabic{table}} 
\renewcommand{\thetable}{\Alph{section}\arabic{table}}  
\counterwithin{table}{section}
\counterwithin{figure}{section}

\twocolumn[{%
 \centering
 \LARGE Supplementary Material\\[1.5em]
}]

\section{Formal properties of intensity-based aggregation strategies}\label{app:formal_properties}

In this section we give formal definitions of relevant properties of intensity-based \Aggs{}. The properties \textit{Monotonicity} and \textit{Proportion invariance} have already been mentioned in the main text where we showed how they impact downstream performance. Here, we additionally discuss \textit{Parameter independence} and \textit{Locality}.

When an intesity-based \Agg{} exhibits a given property, we provide a formal proof or justification. Conversely, if an \Agg{} does not satisfy a property, we present a counterexample. The counterexamples are derived from experiments on simplified toy datasets—designed to be idealized test cases. If an \Agg{} fails under these conditions, it is likely to be even more unreliable in real-world applications.

\begin{table}[h!]
    \centering
    \begin{tabular}{l|c|c|c|c}
        & \textbf{PF} & \textbf{M} & \textbf{PI} &  \textbf{L}\\
        \hline
        Global Average (AVG)  & \cmark & \cmark & \xmark & \xmark \\
        Above-Threshold Average (ATA) & \xmark & \xmark & \cmark & \xmark \\
        Above-Quantile Average (AQA) & \xmark & \cmark & \xmark & \xmark \\
        Patch-Level Maximum (PLM) & \xmark & \cmark & \cmark & \cmark \\
        Weighted Class Average (WCA) & \xmark & \cmark & \xmark$^*$  & \xmark \\
    \end{tabular}
    \caption{\textbf{Overview of key properties of selected \Aggs{}.}  PF: Parameter-free, M: Monotonic, PI: Proportion invariant, L: Local.
    $^*$WCA becomes proportion invariant when the background class is excluded in specific use cases where foreground classes exhibit high uncertainty and the background class has low uncertainty.}
    \label{tab:comparison_suppl_mat}
\end{table}

\subsection{Parameter-fee property}\label{app:aggs-params}
AVG clearly is parameter-free.
ATA depends on the fixed threshold $T>0$ and is only applicable if $T$ is less than 
the pixelwise maximum of the heatmap $U$. 

\noindent AQA depends on the fixed portion $q\in(0,1)$ of lower uncertainty values above which all uncertainties are averaged. 

\noindent PLM depends on the choice of patch size. While too small patch sizes are prone to outliers (\eg for a $1\times1$ patch size PLM is equivalent to the pixelwise maximum), too large patch sizes approximate AVG and thus might return lower averages than expected.

\noindent Weighted Class Average (WCA) technically depends on the choice of weights $w_c$ for the classes. However, the most important examples Single-Class Average (SCA), Balanced Class Average (BCA) and Imbalanced Class Average (ICA) are parameter-free since the choice of weights is fixed.

\subsection{Monotonicity}
We call an \Agg{} \textit{monotonic} if an increase in all pixelwise uncertainty scores leads to an increase in the aggregated score. More formally, \Agg{} $f$ is monotonic if and only if $f(U)\leq f(V)$ for any uncertainty maps $U, V$ satisfying $u_i\leq v_i$ for each pixel $i=1,\dots,mn$.

\noindent This property ensures that the aggregated scores reliably and intuitively track increases in pixelwise uncertainty. Suppose, \eg that an \ood{} sample would lead to a slight increase in uncertainty distributed equally across all pixels of the uncertainty map. A non-monotonic \Agg{} might assign similar aggregated scores to both the \iid{} and the \ood{} sample making them impossible to distinguish despite a clear difference in pixelwise uncertainty.



\begin{figure}
    \centering
    \includegraphics[width=\linewidth]{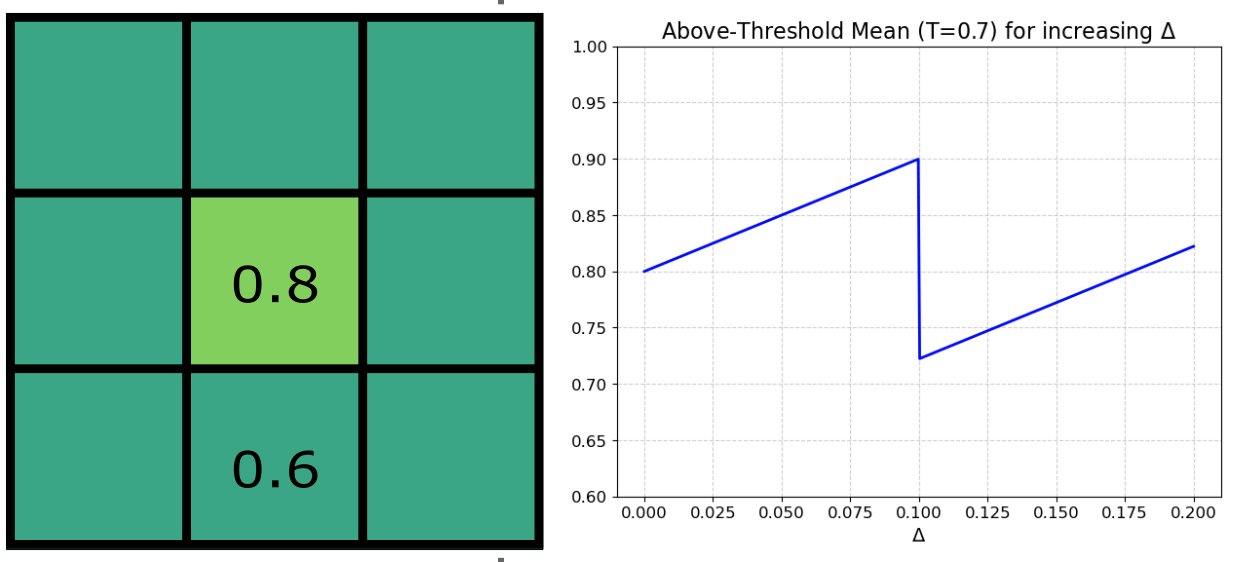}
    \caption{\textbf{Non-monotonicity of ATA.} Left is shown a $3\times3$ uncertainty map. If we increase each pixel value by $\Delta$ the ATA score (threshold $T=0.7$) will drop at a value of $\Delta=0.1$, since the amount of pixels affecting the average suddenly increases as their values pass the threshold.}
    \label{fig:non_mono}
\end{figure}

\noindent AVG is monotonic since if we have two uncertainty maps $U, V$ of shape $m\times n$ satisfying $u_{i} \leq v_{i}$ for all pixels $i=1,\dots,mn$ then for the AVG holds
\begin{equation}
\text{AVG}(U) = \frac{1}{mn}\sum_{i} u_{i} \leq \frac{1}{mn}\sum_{i} v_{i} = \text{AVG}(V)
\end{equation}
This implies that PLM is also monotonic, since it computes the average w.r.t.\ to patches of fixed size.

\noindent To prove that AQA is monotonic, consider the lists of pixelwise uncertainty values of $U$ resp. $V$ \emph{sorted ascendingly}: $u^1\leq\dots\leq u^{mn}$ resp. $v^1\leq \dots\leq v^{mn}$. Our assumption that each pixelwise value of $V$ is increased compared to $U$ implies that for both lists we have $u^i\leq v^i$ for all $i=1,\dots,mn$. Therefore, the $q$-quantile w.r.t.\ the values of $V$ must be greater or equal to the $q$-quantile w.r.t.\  $U$. Furthermore, since both lists have the same number of uncertainty values lying above the $q$-quantile the average of those values for $V$ will be greater or equal to the average for $U$.

\noindent For each choice of class weights $w_c$ the WCA is monotonic as well: If $u_i \geq v_i$ for all pixels $i$ for uncertainty maps $U, V$, then in particular for the class-wise average we have $\alpha_c^U \geq \alpha_c^V$ for each class $c$. For a fixed choice of weights $w_c$ this implies 
\begin{equation}
    \text{WCA}(U) =  \sum_c w_c \alpha_c^U \geq =  \sum_c w_c \alpha_c^V = \text{WCA}(V)
\end{equation}

\noindent In contrast, ATA is not monotonic, a counterexample is illustrated in \cref{fig:aggregation_strategies}b and \cref{fig:non_mono}.

\subsection{Proportion invariance}
Consider idealized binary uncertainty maps having only low-uncertainty regions with uncertainty values approximately $0$ (\eg irrelevant background) and high-uncertainty regions with uncertainty values approximately $1$ (\eg  relevant foreground).

\noindent An \Agg{} $f$ is \textit{proportion invariant} if $f(U)= f(V)$ for any such uncertainty maps $U, V$ only differing by the area proportions of the low- resp.\ high-uncertainty regions. A situation where this property would be desirable is the cropping of uncertainty maps: in this case we might want the aggregated uncertainty score to be unaffected when irrelevant low-uncertainty regions in the background are removed.

\noindent By this definition AVG is highly dependent on the present area-proportion, as it divides the sum over all pixelwise uncertainty by the total number of pixels. 

\noindent AQA also depends on the proportion of low-uncertainty pixels as it considers the $q$ highest uncertainty values in $U$ 
If the proportion of high-uncertainty values drops below $q$, the selection of the top $q$ values will inevitably include low-uncertainty pixels, resulting in a lower score. This effect is illustrated in \cref{fig:aggregation_strategies} (b), where the presence of low-uncertainty pixels influences the final aggregation outcome.

\noindent In contrast, PLM only depends on the uncertainty values within the maximal patch. Increasing or decreasing low-uncertainty proportion does not affect this patch and thus PLM remains unaffected.

\noindent Similarly, ATA only depends on high uncertainty values. Increasing or decreasing low-uncertainty  proportion does not affect this set of high values (assuming that the threshold $T$ is chosen sufficiently high to not capture low background uncertainty). As a result, ATA is proportion invariant.

\noindent The WCA is generally not proportion invariant, as changes in class proportions influence individual class averages, thereby affecting the overall weighted sum. However, in scenarios where high uncertainty is typically concentrated in foreground classes and low uncertainty is prevalent in the background
the BCA and ICA
remain proportion invariant in these settings, as they are unaffected by background proportion changes.

\subsection{Locality}
AVG is non-local as it is computed across all pixels. 
Similarly, ATA and AQA are non-local \Aggs{} since the relevant uncertainty values above threshold $T$ or $q$-quantile may occur at pixels across the whole image 
In contrast, PLM is local as its resulting score only depends on the uncertainty values of pixels within the maximal patch.

\noindent The WCA is generally not a local aggregation strategy, as it computes a weighted sum of class-wise uncertainty averages across all classes. Consequently, if all class weights are nonzero, every pixel in $U$ contributes to the final score, making it inherently global rather than local.
However, in the specific case of the SCA applied to a class that is a priori known to have spatially localized instances—meaning it occupies only a small proportion of the image—SCA can indeed exhibit local behavior. 

\paragraph{Applicability to spatial measures}
We do not apply these properties to spatial measures because their values reflect local structural patterns in uncertainty maps and should not directly be compared to uncertainty intensities.
Consequently, properties like monotonicity are irrelevant for spatial aggregators, as the increase of spatial measures is not always related to an increase of uncertainty. 
For instance, a global pixelwise increase of uncertainty can potentially decrease the entropy across an uncertainty map, in particular if the uncertainty values after the increase all lie within the same bin.


\section{Details on spatial aggregation strategies}
\label{sec:spatial_mass_ratio}

\begin{figure}[t]
    \centering
    \includegraphics[width=\linewidth]{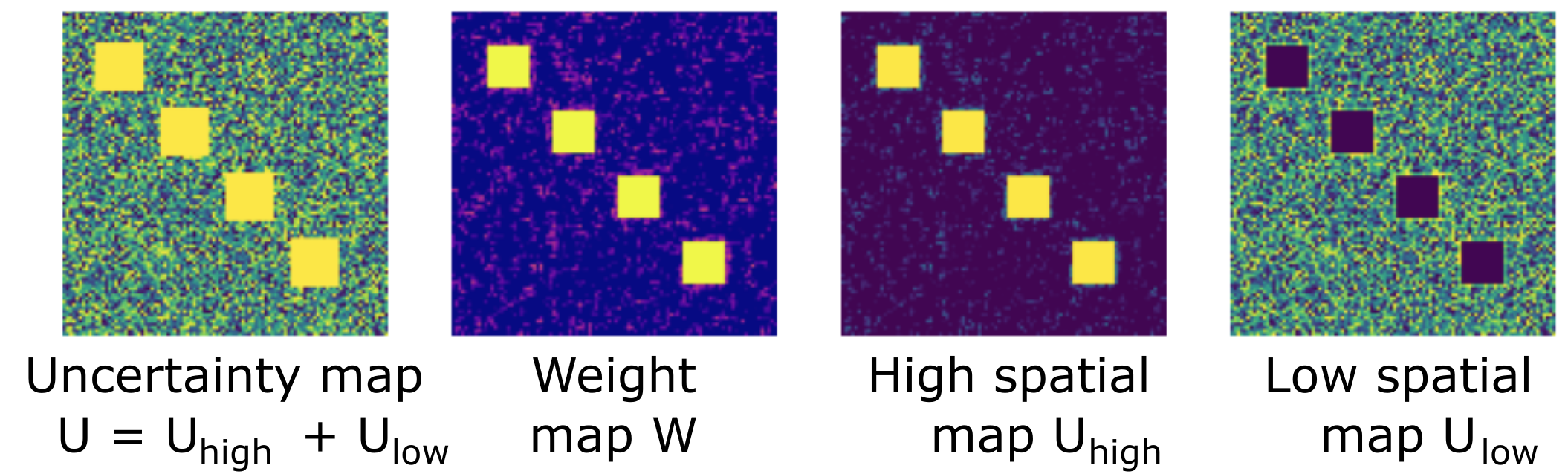}
    \caption{\textbf{Spatial decomposition using Moran's I.} For an uncertainty map $U$ we compute the pixelwise local spatial measure w.r.t. to Moran's I which captures noise ($I=0$) and clusters ($I=1$). The local spatial measures can be interpreted as weights of a weight matrix $W$ where high weights correspond to high spatial auto-correlation. The pixelwise product of $U$ and $W$ is $U_{\text{high}}$ while the pixelwise product of $U$ and $1-W$ is $U_{\text{low}}$, which yields a matrix decomposition of the original uncertainty map.}
    \label{fig:spatial_decomposition}
\end{figure}

\subsection{Selection of spatial measures}
For our experiments, we use the following spatial measures: Moran’s I \cite{Moran1948TheIO}, Edge Density Score \cite{Pereira2014Edge}, and Shannon Entropy \cite{shannon}.

\paragraph{Moran’s I} It measures spatial autocorrelation by comparing the similarity of values at neighboring pixels. A high positive value indicates that similar uncertainty values cluster together, while a value near zero implies spatial randomness. In its original form, Moran’s I ranges from $-1$ (negative correlation) to $1$ (positive correlation).
For our analysis we adapt Moran’s I by capping negative values at 0, as negative spatial correlation (\eg checkerboard patterns) is not to be expected in our data.

\paragraph{Edge Density Score} It quantifies the presence of edges by computing the proportion of pixels within a local window whose gradient magnitude exceeds a specified threshold. It reflects how much spatial variation (\eg sharp transitions or boundaries) is present in the uncertainty map.

\paragraph{Shannon Entropy} It measures the local variability of uncertainty values by computing the Shannon entropy over discretized uncertainty levels within a local window. Higher entropy indicates more heterogeneous or noisy uncertainty, while lower values correspond to more uniform regions. 
\newline

\noindent Both the Edge Density Score and entropy involve additional hyperparameters. The Edge Density Score uses a gradient threshold $\tau$ to classify pixels as edge-like; we set $\tau = 0.2$. Entropy is computed using $b = 4$ bins to discretize local uncertainty values. Another viable spatial measure is Geary’s C \cite{geary1954contiguity}, which captures local dissimilarity, with values close to 1 indicating randomness, below 1 indicating positive autocorrelation, and above 1 indicating negative autocorrelation. However, we omit it, as it provides information similar to Moran’s I.

\subsection{Spatial mass ratio}
Given an uncertainty map \( U \) and a spatial measure with values between 0 and 1, we propose to calculate a \emph{Spatial Mass Ratio (SMR)} which reflects how much of the total uncertainty is concentrated in spatially structured regions of the image. \newline \noindent 
For each pixel \( u \) of the (1-padded) uncertainty map \( U \) we first compute the pixelwise local spatial measure \( w_u \) within a sliding window of size \( 3 \times 3 \) centered at \( u \). 
The resulting matrix \( W \in [0,1]^{m \times n} \) contains the local spatial measures for all pixels $u_i\in U$. \newline \noindent 
Using $W$ we then compute a filtered uncertainty map \( U^{\text{high}} \) by pixelwise multiplying \( U \) with \( W \), retaining only the uncertainty mass in regions of high local spatial measure. 
Similarly, multiplying \( U \) with \( 1 - W \) yields \( U^{\text{low}} \), capturing the mass in regions of low local spatial measure. This results in a spatial decomposition of the uncertainty (see ~\Cref{fig:spatial_decomposition}):
\[
U = U^{\text{high}} + U^{\text{low}}.
\]
Finally, we define the \emph{Spatial Mass Ratio (SMR)} as the portion of uncertainty mass located in pixels with high spatial measure:
\[
\text{SMR} := \frac{\sum_{i}^{nm} u_i w_i}{\sum_{i}^{nm} u_i} \in [0,1].
\]
The behavior of the SMR can be intuitively understood by considering its extreme cases:
For Moran's I, an SMR of 0 indicates that all uncertainty is located in noise-like, uncorrelated regions, while a value of 1 reflects fully clustered uncertainty. For Edge Density Score, SMR equals 0 when uncertainty lies entirely in flat areas and reaches 1 when it is concentrated along edges. For Entropy, an SMR of 0 means local uncertainty values fall within a single bin (i.e., low variability), whereas a value of 1 corresponds to a uniform distribution across bins, indicating high local randomness.

\section{Details on Benchmarking Results}
\label{app:mean_rank_tables}

\begin{table}[t]
    \centering
    \setlength{\tabcolsep}{2.3pt}
    \renewcommand{\arraystretch}{1.75}
    \scriptsize
    \newcommand{\headerbox}[1]{%
        \tikz[baseline=(X.base)] \node[draw=gray, fill=white, rounded corners=4pt, line width=0.75pt, inner sep=3pt, minimum width=5.5em,] (X) {\strut#1};%
    }
    \scalebox{0.62}{%
\begin{tabular}{rcccccccccc|c}
\multicolumn{1}{c}{} & \multicolumn{10}{l}{\textbf{a \qquad AUROC}} & \\
\textbf{BCA}\, & \heatmapAUROC{.72}{.47}{.93} $\pm$ .06 & \heatmapAUROC{.79}{.61}{.87} $\pm$ .05 & \heatmapAUROC{.89}{.32}{1.00} $\pm$ .01 & \heatmapAUROC{.57}{.42}{.96} $\pm$ .05 & \heatmapAUROC{.82}{.45}{.89} $\pm$ .07 & \heatmapAUROC{.68}{.45}{.77} $\pm$ .02 & \heatmapAUROC{.68}{.41}{.82} $\pm$ .02 & \heatmapAUROC{.58}{.32}{.95} $\pm$ .06 & \heatmapAUROC{.77}{.43}{1.00} $\pm$ .06 & \heatmapAUROC{.95}{.54}{1.00} $\pm$ .02 & \textbf{5.2} \\
\textbf{GMM-All}\, & \heatmapAUROC{.84}{.47}{.93} $\pm$ .05 & \heatmapAUROC{.87}{.61}{.87} $\pm$ .05 & \heatmapAUROC{1.00}{.32}{1.00} $\pm$ .00 & \heatmapAUROC{.86}{.42}{.96} $\pm$ .03 & \heatmapAUROC{.77}{.45}{.89} $\pm$ .05 & \heatmapAUROC{.45}{.45}{.77} $\pm$ .03 & \heatmapAUROC{.44}{.41}{.82} $\pm$ .03 & \heatmapAUROC{.95}{.32}{.95} $\pm$ .03 & \heatmapAUROC{1.00}{.43}{1.00} $\pm$ .00 & \heatmapAUROC{1.00}{.54}{1.00} $\pm$ .00 & \textbf{5.2} \\
\textbf{ICA}\, & \heatmapAUROC{.60}{.47}{.93} $\pm$ .07 & \heatmapAUROC{.83}{.61}{.87} $\pm$ .05 & \heatmapAUROC{.84}{.32}{1.00} $\pm$ .02 & \heatmapAUROC{.57}{.42}{.96} $\pm$ .05 & \heatmapAUROC{.82}{.45}{.89} $\pm$ .07 & \heatmapAUROC{.71}{.45}{.77} $\pm$ .02 & \heatmapAUROC{.59}{.41}{.82} $\pm$ .03 & \heatmapAUROC{.57}{.32}{.95} $\pm$ .06 & \heatmapAUROC{.77}{.43}{1.00} $\pm$ .06 & \heatmapAUROC{.95}{.54}{1.00} $\pm$ .02 & \textbf{5.9 }\\
\textbf{GMM-Int}\, & \heatmapAUROC{.79}{.47}{.93} $\pm$ .06 & \heatmapAUROC{.83}{.61}{.87} $\pm$ .06 & \heatmapAUROC{.73}{.32}{1.00} $\pm$ .02 & \heatmapAUROC{.86}{.42}{.96} $\pm$ .03 & \heatmapAUROC{.78}{.45}{.89} $\pm$ .05 & \heatmapAUROC{.49}{.45}{.77} $\pm$ .03 & \heatmapAUROC{.43}{.41}{.82} $\pm$ .03 & \heatmapAUROC{.91}{.32}{.95} $\pm$ .04 & \heatmapAUROC{.98}{.43}{1.00} $\pm$ .01 & \heatmapAUROC{1.00}{.54}{1.00} $\pm$ .00 & \textbf{6.0} \\
\textbf{QFR}\, & \heatmapAUROC{.62}{.47}{.93} $\pm$ .07 & \heatmapAUROC{.86}{.61}{.87} $\pm$ .04 & \heatmapAUROC{.62}{.32}{1.00} $\pm$ .02 & \heatmapAUROC{.54}{.42}{.96} $\pm$ .05 & \heatmapAUROC{.89}{.45}{.89} $\pm$ .05 & \heatmapAUROC{.68}{.45}{.77} $\pm$ .03 & \heatmapAUROC{.57}{.41}{.82} $\pm$ .03 & \heatmapAUROC{.58}{.32}{.95} $\pm$ .06 & \heatmapAUROC{.67}{.43}{1.00} $\pm$ .06 & \heatmapAUROC{.91}{.54}{1.00} $\pm$ .03 & \textbf{6.6} \\
\textbf{GMM-Spa}\, & \heatmapAUROC{.93}{.47}{.93} $\pm$ .03 & \heatmapAUROC{.66}{.61}{.87} $\pm$ .07 & \heatmapAUROC{1.00}{.32}{1.00} $\pm$ .00 & \heatmapAUROC{.67}{.42}{.96} $\pm$ .04 & \heatmapAUROC{.72}{.45}{.89} $\pm$ .07 & \heatmapAUROC{.49}{.45}{.77} $\pm$ .03 & \heatmapAUROC{.41}{.41}{.82} $\pm$ .03 & \heatmapAUROC{.85}{.32}{.95} $\pm$ .05 & \heatmapAUROC{.89}{.43}{1.00} $\pm$ .04 & \heatmapAUROC{.90}{.54}{1.00} $\pm$ .03 & \textbf{7.1} \\
\addlinespace[3pt]
PLM 50\, & \heatmapAUROC{.48}{.47}{.93} $\pm$ .08 & \heatmapAUROC{.68}{.61}{.87} $\pm$ .07 & \heatmapAUROC{.46}{.32}{1.00} $\pm$ .02 & \heatmapAUROC{.95}{.42}{.96} $\pm$ .02 & \heatmapAUROC{.50}{.45}{.89} $\pm$ .07 & \heatmapAUROC{.68}{.45}{.77} $\pm$ .02 & \heatmapAUROC{.75}{.41}{.82} $\pm$ .02 & \heatmapAUROC{.57}{.32}{.95} $\pm$ .05 & \heatmapAUROC{.49}{.43}{1.00} $\pm$ .07 & \heatmapAUROC{.86}{.54}{1.00} $\pm$ .04 & 8.7 \\
AQA 0.60\, & \heatmapAUROC{.53}{.47}{.93} $\pm$ .07 & \heatmapAUROC{.64}{.61}{.87} $\pm$ .07 & \heatmapAUROC{.64}{.32}{1.00} $\pm$ .02 & \heatmapAUROC{.95}{.42}{.96} $\pm$ .02 & \heatmapAUROC{.50}{.45}{.89} $\pm$ .08 & \heatmapAUROC{.76}{.45}{.77} $\pm$ .02 & \heatmapAUROC{.81}{.41}{.82} $\pm$ .02 & \heatmapAUROC{.33}{.32}{.95} $\pm$ .05 & \heatmapAUROC{.49}{.43}{1.00} $\pm$ .07 & \heatmapAUROC{.55}{.54}{1.00} $\pm$ .05 & 8.9 \\
PLM 10\, & \heatmapAUROC{.52}{.47}{.93} $\pm$ .08 & \heatmapAUROC{.64}{.61}{.87} $\pm$ .07 & \heatmapAUROC{.44}{.32}{1.00} $\pm$ .03 & \heatmapAUROC{.91}{.42}{.96} $\pm$ .03 & \heatmapAUROC{.59}{.45}{.89} $\pm$ .09 & \heatmapAUROC{.65}{.45}{.77} $\pm$ .02 & \heatmapAUROC{.65}{.41}{.82} $\pm$ .03 & \heatmapAUROC{.49}{.32}{.95} $\pm$ .04 & \heatmapAUROC{.69}{.43}{1.00} $\pm$ .06 & \heatmapAUROC{.86}{.54}{1.00} $\pm$ .04 & 9.1 \\
PLM 20\, & \heatmapAUROC{.47}{.47}{.93} $\pm$ .08 & \heatmapAUROC{.68}{.61}{.87} $\pm$ .07 & \heatmapAUROC{.42}{.32}{1.00} $\pm$ .03 & \heatmapAUROC{.96}{.42}{.96} $\pm$ .02 & \heatmapAUROC{.52}{.45}{.89} $\pm$ .08 & \heatmapAUROC{.67}{.45}{.77} $\pm$ .02 & \heatmapAUROC{.73}{.41}{.82} $\pm$ .02 & \heatmapAUROC{.56}{.32}{.95} $\pm$ .04 & \heatmapAUROC{.57}{.43}{1.00} $\pm$ .07 & \heatmapAUROC{.84}{.54}{1.00} $\pm$ .04 & 9.2 \\
AQA.75\, & \heatmapAUROC{.52}{.47}{.93} $\pm$ .07 & \heatmapAUROC{.64}{.61}{.87} $\pm$ .07 & \heatmapAUROC{.60}{.32}{1.00} $\pm$ .02 & \heatmapAUROC{.96}{.42}{.96} $\pm$ .02 & \heatmapAUROC{.50}{.45}{.89} $\pm$ .08 & \heatmapAUROC{.77}{.45}{.77} $\pm$ .02 & \heatmapAUROC{.82}{.41}{.82} $\pm$ .02 & \heatmapAUROC{.33}{.32}{.95} $\pm$ .05 & \heatmapAUROC{.46}{.43}{1.00} $\pm$ .07 & \heatmapAUROC{.55}{.54}{1.00} $\pm$ .05 & 9.7\\
AVG\, & \heatmapAUROC{.50}{.47}{.93} $\pm$ .07 & \heatmapAUROC{.64}{.61}{.87} $\pm$ .07 & \heatmapAUROC{.74}{.32}{1.00} $\pm$ .02 & \heatmapAUROC{.95}{.42}{.96} $\pm$ .02 & \heatmapAUROC{.50}{.45}{.89} $\pm$ .08 & \heatmapAUROC{.77}{.45}{.77} $\pm$ .02 & \heatmapAUROC{.79}{.41}{.82} $\pm$ .02 & \heatmapAUROC{.33}{.32}{.95} $\pm$ .06 & \heatmapAUROC{.48}{.43}{1.00} $\pm$ .07 & \heatmapAUROC{.56}{.54}{1.00} $\pm$ .05 & 9.8 \\
AQA .9\, & \heatmapAUROC{.54}{.47}{.93} $\pm$ .07 & \heatmapAUROC{.69}{.61}{.87} $\pm$ .07 & \heatmapAUROC{.46}{.32}{1.00} $\pm$ .03 & \heatmapAUROC{.96}{.42}{.96} $\pm$ .02 & \heatmapAUROC{.50}{.45}{.89} $\pm$ .08 & \heatmapAUROC{.75}{.45}{.77} $\pm$ .02 & \heatmapAUROC{.77}{.41}{.82} $\pm$ .02 & \heatmapAUROC{.32}{.32}{.95} $\pm$ .05 & \heatmapAUROC{.43}{.43}{1.00} $\pm$ .07 & \heatmapAUROC{.54}{.54}{1.00} $\pm$ .05 & 9.9 \\
ATA .5\, & \heatmapAUROC{.66}{.47}{.93} $\pm$ .07 & \heatmapAUROC{.78}{.61}{.87} $\pm$ .06 & \heatmapAUROC{.32}{.32}{1.00} $\pm$ .02 & \heatmapAUROC{.47}{.42}{.96} $\pm$ .05 & \heatmapAUROC{.45}{.45}{.89} $\pm$ .07 & \heatmapAUROC{.61}{.45}{.77} $\pm$ .03 & \heatmapAUROC{.59}{.41}{.82} $\pm$ .03 & \heatmapAUROC{.43}{.32}{.95} $\pm$ .05 & \heatmapAUROC{.54}{.43}{1.00} $\pm$ .07 & \heatmapAUROC{.63}{.54}{1.00} $\pm$ .05 & 11.1 \\
ATA .3\, & \heatmapAUROC{.53}{.47}{.93} $\pm$ .07 & \heatmapAUROC{.61}{.61}{.87} $\pm$ .07 & \heatmapAUROC{.35}{.32}{1.00} $\pm$ .02 & \heatmapAUROC{.42}{.42}{.96} $\pm$ .05 & \heatmapAUROC{.54}{.45}{.89} $\pm$ .07 & \heatmapAUROC{.67}{.45}{.77} $\pm$ .02 & \heatmapAUROC{.58}{.41}{.82} $\pm$ .03 & \heatmapAUROC{.42}{.32}{.95} $\pm$ .04 & \heatmapAUROC{.58}{.43}{1.00} $\pm$ .06 & \heatmapAUROC{.67}{.54}{1.00} $\pm$ .05 & 11.5 \\
ATA .7\, & \heatmapAUROC{.48}{.47}{.93} $\pm$ .08 & \heatmapAUROC{.79}{.61}{.87} $\pm$ .06 & \heatmapAUROC{.36}{.32}{1.00} $\pm$ .02 & \heatmapAUROC{.51}{.42}{.96} $\pm$ .06 & \heatmapAUROC{.51}{.45}{.89} $\pm$ .07 & \heatmapAUROC{.56}{.45}{.77} $\pm$ .03 & \heatmapAUROC{.51}{.41}{.82} $\pm$ .03 & \heatmapAUROC{.35}{.32}{.95} $\pm$ .04 & \heatmapAUROC{.48}{.43}{1.00} $\pm$ .07 & \heatmapAUROC{.58}{.54}{1.00} $\pm$ .05 & 12.0 \\
\addlinespace[5pt]
\multicolumn{1}{c}{} & \multicolumn{10}{l}{\textbf{b \qquad E-AURC}} & \\
\textbf{QFR}\, & \heatmapEAURC{.04}{.04}{.13} $\pm$ .01 & \heatmapEAURC{.02}{.02}{.04} $\pm$ .00 & \heatmapEAURC{.06}{.04}{.10} $\pm$ .01 & \heatmapEAURC{.05}{.05}{.20} $\pm$ .01 & \heatmapEAURC{.09}{.06}{.15} $\pm$ .04 & \heatmapEAURC{.26}{.16}{.33} $\pm$ .01 & \heatmapEAURC{.27}{.20}{.37} $\pm$ .01 & \heatmapEAURC{.16}{.08}{.32} $\pm$ .02 & \heatmapEAURC{.04}{.03}{.13} $\pm$ .01 & \heatmapEAURC{.04}{.04}{.13} $\pm$ .01 & \textbf{3.4} \\
\textbf{GMM-All}\, & \heatmapEAURC{.05}{.04}{.13} $\pm$ .01 & \heatmapEAURC{.03}{.02}{.04} $\pm$ .00 & \heatmapEAURC{.05}{.04}{.10} $\pm$ .01 & \heatmapEAURC{.07}{.05}{.20} $\pm$ .01 & \heatmapEAURC{.07}{.06}{.15} $\pm$ .01 & \heatmapEAURC{.16}{.16}{.33} $\pm$ .01 & \heatmapEAURC{.21}{.20}{.37} $\pm$ .01 & \heatmapEAURC{.08}{.08}{.32} $\pm$ .01 & \heatmapEAURC{.09}{.03}{.13} $\pm$ .01 & \heatmapEAURC{.06}{.04}{.13} $\pm$ .01 & \textbf{4.2} \\
\textbf{GMM-Int}\, & \heatmapEAURC{.06}{.04}{.13} $\pm$ .01 & \heatmapEAURC{.03}{.02}{.04} $\pm$ .01 & \heatmapEAURC{.10}{.04}{.10} $\pm$ .01 & \heatmapEAURC{.07}{.05}{.20} $\pm$ .01 & \heatmapEAURC{.06}{.06}{.15} $\pm$ .01 & \heatmapEAURC{.17}{.16}{.33} $\pm$ .01 & \heatmapEAURC{.20}{.20}{.37} $\pm$ .01 & \heatmapEAURC{.09}{.08}{.32} $\pm$ .02 & \heatmapEAURC{.09}{.03}{.13} $\pm$ .01 & \heatmapEAURC{.06}{.04}{.13} $\pm$ .01 & \textbf{5.4} \\
\textbf{GMM-Spa}\, & \heatmapEAURC{.05}{.04}{.13} $\pm$ .01 & \heatmapEAURC{.04}{.02}{.04} $\pm$ .01 & \heatmapEAURC{.05}{.04}{.10} $\pm$ .00 & \heatmapEAURC{.08}{.05}{.20} $\pm$ .01 & \heatmapEAURC{.08}{.06}{.15} $\pm$ .01 & \heatmapEAURC{.18}{.16}{.33} $\pm$ .01 & \heatmapEAURC{.20}{.20}{.37} $\pm$ .01 & \heatmapEAURC{.12}{.08}{.32} $\pm$ .02 & \heatmapEAURC{.13}{.03}{.13} $\pm$ .03 & \heatmapEAURC{.07}{.04}{.13} $\pm$ .01 & \textbf{6.0} \\
\textbf{BCA}\, & \heatmapEAURC{.04}{.04}{.13} $\pm$ .01 & \heatmapEAURC{.03}{.02}{.04} $\pm$ .00 & \heatmapEAURC{.04}{.04}{.10} $\pm$ .01 & \heatmapEAURC{.06}{.05}{.20} $\pm$ .01 & \heatmapEAURC{.13}{.06}{.15} $\pm$ .04 & \heatmapEAURC{.30}{.16}{.33} $\pm$ .01 & \heatmapEAURC{.33}{.20}{.37} $\pm$ .01 & \heatmapEAURC{.20}{.08}{.32} $\pm$ .02 & \heatmapEAURC{.10}{.03}{.13} $\pm$ .03 & \heatmapEAURC{.09}{.04}{.13} $\pm$ .01 & \textbf{6.7} \\
\addlinespace[3pt]
ATA .3\, & \heatmapEAURC{.05}{.04}{.13} $\pm$ .01 & \heatmapEAURC{.03}{.02}{.04} $\pm$ .01 & \heatmapEAURC{.09}{.04}{.10} $\pm$ .01 & \heatmapEAURC{.12}{.05}{.20} $\pm$ .03 & \heatmapEAURC{.14}{.06}{.15} $\pm$ .04 & \heatmapEAURC{.27}{.16}{.33} $\pm$ .01 & \heatmapEAURC{.28}{.20}{.37} $\pm$ .01 & \heatmapEAURC{.28}{.08}{.32} $\pm$ .02 & \heatmapEAURC{.03}{.03}{.13} $\pm$ .01 & \heatmapEAURC{.09}{.04}{.13} $\pm$ .01 & 8.0 \\
ATA .5\, & \heatmapEAURC{.04}{.04}{.13} $\pm$ .01 & \heatmapEAURC{.02}{.02}{.04} $\pm$ .00 & \heatmapEAURC{.10}{.04}{.10} $\pm$ .01 & \heatmapEAURC{.19}{.05}{.20} $\pm$ .04 & \heatmapEAURC{.15}{.06}{.15} $\pm$ .04 & \heatmapEAURC{.23}{.16}{.33} $\pm$ .01 & \heatmapEAURC{.28}{.20}{.37} $\pm$ .01 & \heatmapEAURC{.29}{.08}{.32} $\pm$ .02 & \heatmapEAURC{.03}{.03}{.13} $\pm$ .01 & \heatmapEAURC{.08}{.04}{.13} $\pm$ .01 & 8.1 \\
PLM 10\, & \heatmapEAURC{.06}{.04}{.13} $\pm$ .01 & \heatmapEAURC{.03}{.02}{.04} $\pm$ .00 & \heatmapEAURC{.09}{.04}{.10} $\pm$ .01 & \heatmapEAURC{.09}{.05}{.20} $\pm$ .03 & \heatmapEAURC{.13}{.06}{.15} $\pm$ .04 & \heatmapEAURC{.27}{.16}{.33} $\pm$ .01 & \heatmapEAURC{.30}{.20}{.37} $\pm$ .01 & \heatmapEAURC{.30}{.08}{.32} $\pm$ .02 & \heatmapEAURC{.11}{.03}{.13} $\pm$ .04 & \heatmapEAURC{.05}{.04}{.13} $\pm$ .01 & 8.3 \\
PLM 20\, & \heatmapEAURC{.06}{.04}{.13} $\pm$ .01 & \heatmapEAURC{.02}{.02}{.04} $\pm$ .01 & \heatmapEAURC{.09}{.04}{.10} $\pm$ .01 & \heatmapEAURC{.10}{.05}{.20} $\pm$ .03 & \heatmapEAURC{.14}{.06}{.15} $\pm$ .04 & \heatmapEAURC{.28}{.16}{.33} $\pm$ .01 & \heatmapEAURC{.33}{.20}{.37} $\pm$ .01 & \heatmapEAURC{.29}{.08}{.32} $\pm$ .02 & \heatmapEAURC{.09}{.03}{.13} $\pm$ .03 & \heatmapEAURC{.06}{.04}{.13} $\pm$ .01 & 8.5 \\
ATA .7\, & \heatmapEAURC{.13}{.04}{.13} $\pm$ .02 & \heatmapEAURC{.02}{.02}{.04} $\pm$ .00 & \heatmapEAURC{.10}{.04}{.10} $\pm$ .01 & \heatmapEAURC{.20}{.05}{.20} $\pm$ .04 & \heatmapEAURC{.15}{.06}{.15} $\pm$ .04 & \heatmapEAURC{.21}{.16}{.33} $\pm$ .01 & \heatmapEAURC{.25}{.20}{.37} $\pm$ .01 & \heatmapEAURC{.30}{.08}{.32} $\pm$ .02 & \heatmapEAURC{.04}{.03}{.13} $\pm$ .01 & \heatmapEAURC{.08}{.04}{.13} $\pm$ .01 & 9.4 \\
PLM 50\, & \heatmapEAURC{.08}{.04}{.13} $\pm$ .01 & \heatmapEAURC{.03}{.02}{.04} $\pm$ .00 & \heatmapEAURC{.09}{.04}{.10} $\pm$ .01 & \heatmapEAURC{.10}{.05}{.20} $\pm$ .03 & \heatmapEAURC{.14}{.06}{.15} $\pm$ .04 & \heatmapEAURC{.28}{.16}{.33} $\pm$ .01 & \heatmapEAURC{.34}{.20}{.37} $\pm$ .01 & \heatmapEAURC{.29}{.08}{.32} $\pm$ .02 & \heatmapEAURC{.09}{.03}{.13} $\pm$ .03 & \heatmapEAURC{.07}{.04}{.13} $\pm$ .01 & 10.0 \\
AQA .9\, & \heatmapEAURC{.05}{.04}{.13} $\pm$ .01 & \heatmapEAURC{.03}{.02}{.04} $\pm$ .00 & \heatmapEAURC{.08}{.04}{.10} $\pm$ .01 & \heatmapEAURC{.10}{.05}{.20} $\pm$ .03 & \heatmapEAURC{.14}{.06}{.15} $\pm$ .04 & \heatmapEAURC{.31}{.16}{.33} $\pm$ .01 & \heatmapEAURC{.35}{.20}{.37} $\pm$ .01 & \heatmapEAURC{.32}{.08}{.32} $\pm$ .03 & \heatmapEAURC{.08}{.03}{.13} $\pm$ .02 & \heatmapEAURC{.13}{.04}{.13} $\pm$ .02 & 10.5 \\
ICA\, & \heatmapEAURC{.09}{.04}{.13} $\pm$ .01 & \heatmapEAURC{.04}{.02}{.04} $\pm$ .01 & \heatmapEAURC{.06}{.04}{.10} $\pm$ .01 & \heatmapEAURC{.06}{.05}{.20} $\pm$ .01 & \heatmapEAURC{.13}{.06}{.15} $\pm$ .04 & \heatmapEAURC{.33}{.16}{.33} $\pm$ .01 & \heatmapEAURC{.36}{.20}{.37} $\pm$ .01 & \heatmapEAURC{.31}{.08}{.32} $\pm$ .03 & \heatmapEAURC{.10}{.03}{.13} $\pm$ .03 & \heatmapEAURC{.09}{.04}{.13} $\pm$ .01 & 10.7 \\
AQA .75\, & \heatmapEAURC{.06}{.04}{.13} $\pm$ .01 & \heatmapEAURC{.03}{.02}{.04} $\pm$ .01 & \heatmapEAURC{.08}{.04}{.10} $\pm$ .01 & \heatmapEAURC{.10}{.05}{.20} $\pm$ .03 & \heatmapEAURC{.14}{.06}{.15} $\pm$ .04 & \heatmapEAURC{.32}{.16}{.33} $\pm$ .01 & \heatmapEAURC{.37}{.20}{.37} $\pm$ .01 & \heatmapEAURC{.31}{.08}{.32} $\pm$ .03 & \heatmapEAURC{.09}{.03}{.13} $\pm$ .03 & \heatmapEAURC{.13}{.04}{.13} $\pm$ .02 & 11.6 \\
AVG\, & \heatmapEAURC{.09}{.04}{.13} $\pm$ .01 & \heatmapEAURC{.04}{.02}{.04} $\pm$ .01 & \heatmapEAURC{.07}{.04}{.10} $\pm$ .01 & \heatmapEAURC{.10}{.05}{.20} $\pm$ .03 & \heatmapEAURC{.14}{.06}{.15} $\pm$ .04 & \heatmapEAURC{.32}{.16}{.33} $\pm$ .01 & \heatmapEAURC{.36}{.20}{.37} $\pm$ .01 & \heatmapEAURC{.31}{.08}{.32} $\pm$ .03 & \heatmapEAURC{.09}{.03}{.13} $\pm$ .03 & \heatmapEAURC{.12}{.04}{.13} $\pm$ .01 & 11.9 \\
AQA .6\, & \heatmapEAURC{.07}{.04}{.13} $\pm$ .01 & \heatmapEAURC{.04}{.02}{.04} $\pm$ .01 & \heatmapEAURC{.08}{.04}{.10} $\pm$ .01 & \heatmapEAURC{.10}{.05}{.20} $\pm$ .03 & \heatmapEAURC{.14}{.06}{.15} $\pm$ .04 & \heatmapEAURC{.32}{.16}{.33} $\pm$ .01 & \heatmapEAURC{.36}{.20}{.37} $\pm$ .01 & \heatmapEAURC{.31}{.08}{.32} $\pm$ .03 & \heatmapEAURC{.09}{.03}{.13} $\pm$ .03 & \heatmapEAURC{.13}{.04}{.13} $\pm$ .01 & 12.0 \\
\addlinespace[29pt]
\multicolumn{1}{c}{} & \multicolumn{1}{c}{\smash{\rotatebox[origin=lt]{55}{ARC-BC}}} & \multicolumn{1}{c}{\smash{\rotatebox[origin=lt]{55}{ARC-Nuc}}} & \multicolumn{1}{c}{\smash{\rotatebox[origin=lt]{55}{CAR-CS}}} & \multicolumn{1}{c}{\smash{\rotatebox[origin=lt]{55}{LIDC-Mal}}} & \multicolumn{1}{c}{\smash{\rotatebox[origin=lt]{55}{LIDC-Tex}}} & \multicolumn{1}{c}{\smash{\rotatebox[origin=lt]{55}{LIZ-IG}}} & \multicolumn{1}{c}{\smash{\rotatebox[origin=lt]{55}{LIZ-SG}}} & \multicolumn{1}{c}{\smash{\rotatebox[origin=lt]{55}{WEED-Hand}}} & \multicolumn{1}{c}{\smash{\rotatebox[origin=lt]{55}{WORM-Nem}}} & \multicolumn{1}{c}{\smash{\rotatebox[origin=lt]{55}{WORM-Pro}}} & \\
\end{tabular}%
}
    \caption{\textbf{Performance of \Aggs{} on uncertainty maps generated with MC Dropout (MCD) in \ood{} and failure detection.} Columns are color-coded with a red-to-green gradient, where white represents the mean (for AUROC this is equivalent to random guessing). (a) Higher AUROC values (greener cells, 0.6–1.0) indicate better \iid{} vs. \ood{} separation, while (b) lower E-AURC values (greener) indicate better alignment between uncertainty estimates and prediction errors. \Aggs{} are ranked from best (top) to worst (bottom) based on their average metric, first computed across 500 bootstrap samples per dataset, and then averaged across datasets to ensure stable rankings (rightmost column). Each value is reported with the standard deviation across bootstrap samples. These tables complement Figures~\ref{fig:ood_results} and \ref{fig:fd_results}b.}
    \label{tab:mdc_bootstrap_results}
\end{table}

\noindent Table~\ref{tab:mdc_bootstrap_results} provides a more detailed analysis
of the results shown in Figures~\ref{fig:ood_results} and~\ref{fig:fd_results}b, reporting mean scores and standard deviations computed over 500 bootstrap samples of the evaluation data. This approach reduces the risk that observed performance is driven by sample variability in the test sets. As expected, standard deviations are larger when fewer evaluation samples are available (cf. Supp.~\ref{app:train}); however, the relative ranking of \Aggs{} remains stable across datasets. \newline \noindent 
To assess the statistical significance of performance differences, we conducted a one-sided Wilcoxon signed-rank test \cite{wilcoxon1945individual} on the AUROC and E-AURC scores obtained from the bootstrapped samples across the 10 datasets. The null hypothesis ($\mathscr{H}_0$) assumes that aggregator $f_A$ does not outperform $f_B$, while the alternative hypothesis ($\mathscr{H}_1$) asserts that $f_A$ performs significantly better (i.e., higher AUROC or lower E-AURC), with a significance level of $\alpha = 0.05$. Full p-value matrices are omitted here for clarity but are available for \Aggs{} on MCD heatmaps at \url{https://github.com/Kainmueller-Lab/aggrigator_experiments}. \newline \noindent 
To substantiate our claim that the variability in the performance of different \Aggs{} arises from the task and dataset properties rather than the UQ methods used to generate the pixelwise uncertainties we have reproduced our main benchmarking results for additional UQ methods beyond Monte Carlo Dropout (MCD, \cite{gal2016dropout}): Test Time Augmentation (TTA, \cite{moshkov2020test}), Maximum Softmax Probability (MSP, \cite{hendrycks2017a}) and two ensembling approaches: Deep Ensembles (DE, \cite{lakshminarayanan2017simple},  for CAR-CS, LIDC-Mal, and LIDC-Tex) and the computationally lighter Checkpoint Ensemble (CE, \cite{chen2017} for ARC-BC and ARC-Nuc.) \newline \noindent 
TTA selects augmentations at test time that best suit each dataset, avoiding to affect the model’s predictive capacity and improving its ability to generalize. Tables~\ref{tab:ensemble_bootstrapped_results} and~\ref{tab:tta_bootstrapped_results} follow the same statistical protocol as Table~\ref{tab:mdc_bootstrap_results}. Consistent with previous observations, standard deviations increase when fewer evaluation samples are available. Additionally, \emph{GMM-All} and its ablated variants exhibit reduced performance on MSP heatmaps of LIZ-SG and LIZ-IG, suggesting that the underlying data modes are insufficiently separable in high-dimensional feature space. \newline \noindent 
While CE, DE, and TTA are available only for a subset of datasets, MSP provides a comprehensive overview that allows fair comparison with MCD. Table \ref{tab:msp_boostrapped_results} shows that even when the uncertainty estimate is heuristic (as in MSP, which relies solely on softmax scores and thus cannot capture \ood{} uncertainty), the relative performance of \Aggs{} remains largely consistent with that observed under MCD. This suggests, at least empirically, that best practices for aggregation are better identified by evaluating \Aggs{} across diverse datasets rather than across uncertainty methods.

\subsection{On the \Aggs{} performance in \ood{} Detection}
As shown in Table~\ref{tab:msp_boostrapped_results}a, the top performers, based on their average rank, are the \textit{prediction-based} aggregators, as well as the \emph{GMM-All}, \emph{GMM-Int} and \emph{GMM-Spa} applied to MSP uncertainty maps. The conclusions drawn in Section~\ref{sec:ood_res} remain consistent, with the notable exception of the default aggregation strategy AVG: its mean rank improves from 12th to 7th due to stronger performance on ARC-BC, ARC-Nuc, WEED-Pro, and WEED-Nem. This behavior is largely driven by the synthetic nature of the ARC datasets, where MSP is more sensitive to induced and controlled noise, and by the label shift present in the WEED \ood{} variants, which likely reduces model confidence. In LIDC-Tex, MSP fails to detect \ood{} instances: the increased transparency of tumors should create elevated uncertainty along borders within the predicted masks, but the overconfident MSP does not capture this, leading to missed detections (with a consequent increase in E-AURC; cf. Table~\ref{tab:msp_boostrapped_results}b). \newline \noindent 
When the analysis is repeated for CE and DE, as reported in Table~\ref{tab:ensemble_bootstrapped_results}a, the ranking aligns closely with that observed for MCD, reflecting the greater suitability of these techniques for capturing epistemic uncertainty. Results for TTA are available only for a smaller subset of datasets (cf. Table~\ref{tab:tta_bootstrapped_results}a); therefore, it is not possible to determine whether uncertainty increases in \ood{} samples extend to the extreme tail of the uncertainty distribution, potentially affecting the performance of AQA 0.90.  Nonetheless, we continue to observe the dominance of \textit{prediction-based} aggregators and the \emph{GMM-All} \Agg{}, while \emph{GMM-Spa} performs poorly in scenarios where spatial augmentations produce similar uncertainty maps for both \iid{} and \ood{} samples. This is expected, for instance, in the three-label task learned on ARC-Nuc, where removing nuclei intensity leads to fragmented border predictions; for both \iid{} and \ood{} samples, the resulting uncertainty structure resembles the effects of rotations or crops. \newline \noindent 
Across UQ methods, the p-value matrices show that no single aggregator consistently dominates in AUROC, supporting our hypothesis that dataset-specific factors (\eg object count, class imbalance) strongly affect the choice of \Agg{} and no universally optimal method can be determined. Instead, a statistically significant top tier emerges: the \emph{GMM}-based strategies and the weighted averaging approaches (BCA, ICA) significantly outperform all others ($p<0.05$), although no method within this group is uniformly superior. While \textit{GMM-All} often leads, its advantage over BCA or other GMM variants is not always significant. Lower-tier methods, such as threshold-based approaches (ATA, AQA), patch-based PLM, and the baseline AVG, are significantly outperformed in most comparisons, confirming their status as suboptimal choices. \newline \noindent 
Extending the analysis to MSP, TTA, and CE/DE largely reinforces these conclusions. The MSP baseline mirrors the MCD results, showing a clear separation ($p \ll 0.01$) between top-tier \Aggs{} (\emph{GMM}-based ones, BCA, ICA) and lower-tier methods. Interestingly, QFR, while dominant in Failure Detection, ranks lower here but still outperforms most lower-tier methods ($p < 0.05$), and under ensemble- and TTA-based uncertainty it becomes statistically indistinguishable from (or even superior to) \emph{GMM-All} and BCA, suggesting that its foreground–background ratio strategy benefits from targeted augmentations and increased model diversity. Across MSP, TTA, and ensembles, GMM-based \Aggs{} (\emph{GMM-All, GMM-Int, GMM-Spa}) remain consistently strong, rarely showing significant disadvantages. Overall, this analysis supports our core hypothesis that, as a general rule, a stable top group of aggregators exists (the \textit{prediction-aware} strategies and the meta-aggregators family), but that the optimal choice for unexplored cases strongly depends on the specific dataset properties and \ood{} perturbation. 

\begin{table}[t]
    \centering
    \setlength{\tabcolsep}{2.3pt}
    \renewcommand{\arraystretch}{1.75}
    \scriptsize
    \newcommand{\headerbox}[1]{%
        \tikz[baseline=(X.base)] \node[draw=gray, fill=white, rounded corners=4pt, line width=0.75pt, inner sep=3pt, minimum width=5.5em,] (X) {\strut#1};%
    }
    \scalebox{0.62}{%
\begin{tabular}{rcccccccccc|c}
\multicolumn{1}{c}{} & \multicolumn{10}{l}{\textbf{a \qquad AUROC}} & \\
\textbf{GMM-All}\, & \heatmapAUROC{.81}{.47}{.93} $\pm$ .05 & \heatmapAUROC{.85}{.61}{.87} $\pm$ .05 & \heatmapAUROC{1.00}{.32}{1.00} $\pm$ .00 & \heatmapAUROC{.92}{.42}{.96} $\pm$ .03 & \heatmapAUROC{.55}{.45}{.89} $\pm$ .08 & \heatmapAUROC{.55}{.45}{.77} $\pm$ .03 & \heatmapAUROC{.45}{.41}{.82} $\pm$ .03 & \heatmapAUROC{.93}{.32}{.95} $\pm$ .03 & \heatmapAUROC{.99}{.43}{1.00} $\pm$ .01 & \heatmapAUROC{1.00}{.54}{1.00} $\pm$ .00 & \textbf{5.4} \\
\textbf{BCA}\, & \heatmapAUROC{.87}{.47}{.93} $\pm$ .04 & \heatmapAUROC{.75}{.61}{.87} $\pm$ .06 & \heatmapAUROC{.90}{.32}{1.00} $\pm$ .01 & \heatmapAUROC{.60}{.42}{.96} $\pm$ .05 & \heatmapAUROC{.64}{.45}{.89} $\pm$ .09 & \heatmapAUROC{.66}{.45}{.77} $\pm$ .02 & \heatmapAUROC{.73}{.41}{.82} $\pm$ .02 & \heatmapAUROC{.62}{.32}{.95} $\pm$ .06 & \heatmapAUROC{.69}{.43}{1.00} $\pm$ .06 & \heatmapAUROC{.86}{.54}{1.00} $\pm$ .04 & \textbf{5.6} \\
\textbf{ICA}\, & \heatmapAUROC{.83}{.47}{.93} $\pm$ .05 & \heatmapAUROC{.78}{.61}{.87} $\pm$ .06 & \heatmapAUROC{.80}{.32}{1.00} $\pm$ .02 & \heatmapAUROC{.61}{.42}{.96} $\pm$ .05 & \heatmapAUROC{.64}{.45}{.89} $\pm$ .09 & \heatmapAUROC{.70}{.45}{.77} $\pm$ .02 & \heatmapAUROC{.66}{.41}{.82} $\pm$ .02 & \heatmapAUROC{.61}{.32}{.95} $\pm$ .06 & \heatmapAUROC{.68}{.43}{1.00} $\pm$ .06 & \heatmapAUROC{.86}{.54}{1.00} $\pm$ .03 & \textbf{5.9} \\
\textbf{QFR}\, & \heatmapAUROC{.85}{.47}{.93} $\pm$ .05 & \heatmapAUROC{.81}{.61}{.87} $\pm$ .06 & \heatmapAUROC{.56}{.32}{1.00} $\pm$ .02 & \heatmapAUROC{.60}{.42}{.96} $\pm$ .05 & \heatmapAUROC{.62}{.45}{.89} $\pm$ .08 & \heatmapAUROC{.73}{.45}{.77} $\pm$ .02 & \heatmapAUROC{.71}{.41}{.82} $\pm$ .02 & \heatmapAUROC{.60}{.32}{.95} $\pm$ .06 & \heatmapAUROC{.66}{.43}{1.00} $\pm$ .06 & \heatmapAUROC{.89}{.54}{1.00} $\pm$ .03 & \textbf{5.9} \\
\textbf{GMM-Int}\, & \heatmapAUROC{.81}{.47}{.93} $\pm$ .05 & \heatmapAUROC{.76}{.61}{.87} $\pm$ .06 & \heatmapAUROC{.80}{.32}{1.00} $\pm$ .02 & \heatmapAUROC{.91}{.42}{.96} $\pm$ .03 & \heatmapAUROC{.58}{.45}{.89} $\pm$ .07 & \heatmapAUROC{.52}{.45}{.77} $\pm$ .03 & \heatmapAUROC{.46}{.41}{.82} $\pm$ .03 & \heatmapAUROC{.94}{.32}{.95} $\pm$ .03 & \heatmapAUROC{.99}{.43}{1.00} $\pm$ .01 & \heatmapAUROC{1.00}{.54}{1.00} $\pm$ .00 & \textbf{6.3} \\
\textbf{GMM-Spa}\, & \heatmapAUROC{.84}{.47}{.93} $\pm$ .06 & \heatmapAUROC{.73}{.61}{.87} $\pm$ .07 & \heatmapAUROC{1.00}{.32}{1.00} $\pm$ .00 & \heatmapAUROC{.70}{.42}{.96} $\pm$ .05 & \heatmapAUROC{.58}{.45}{.89} $\pm$ .07 & \heatmapAUROC{.51}{.45}{.77} $\pm$ .02 & \heatmapAUROC{.52}{.41}{.82} $\pm$ .03 & \heatmapAUROC{.89}{.32}{.95} $\pm$ .03 & \heatmapAUROC{.82}{.43}{1.00} $\pm$ .05 & \heatmapAUROC{.87}{.54}{1.00} $\pm$ .04 & \textbf{6.7} \\
\addlinespace[3pt]
AVG\, & \heatmapAUROC{.62}{.47}{.93} $\pm$ .08 & \heatmapAUROC{.60}{.61}{.87} $\pm$ .08 & \heatmapAUROC{.74}{.32}{1.00} $\pm$ .02 & \heatmapAUROC{.96}{.42}{.96} $\pm$ .02 & \heatmapAUROC{.40}{.45}{.89} $\pm$ .07 & \heatmapAUROC{.77}{.45}{.77} $\pm$ .02 & \heatmapAUROC{.78}{.41}{.82} $\pm$ .02 & \heatmapAUROC{.32}{.32}{.95} $\pm$ .06 & \heatmapAUROC{.50}{.43}{1.00} $\pm$ .07 & \heatmapAUROC{.65}{.54}{1.00} $\pm$ .05 & 8.7 \\
AQA .6\, & \heatmapAUROC{.68}{.47}{.93} $\pm$ .07 & \heatmapAUROC{.60}{.61}{.87} $\pm$ .07 & \heatmapAUROC{.60}{.32}{1.00} $\pm$ .02 & \heatmapAUROC{.96}{.42}{.96} $\pm$ .02 & \heatmapAUROC{.40}{.45}{.89} $\pm$ .07 & \heatmapAUROC{.77}{.45}{.77} $\pm$ .02 & \heatmapAUROC{.77}{.41}{.82} $\pm$ .02 & \heatmapAUROC{.33}{.32}{.95} $\pm$ .05 & \heatmapAUROC{.47}{.43}{1.00} $\pm$ .07 & \heatmapAUROC{.63}{.54}{1.00} $\pm$ .05 & 9.1 \\
PLM 20\, & \heatmapAUROC{.66}{.47}{.93} $\pm$ .06 & \heatmapAUROC{.62}{.61}{.87} $\pm$ .07 & \heatmapAUROC{.37}{.32}{1.00} $\pm$ .02 & \heatmapAUROC{.96}{.42}{.96} $\pm$ .02 & \heatmapAUROC{.40}{.45}{.89} $\pm$ .08 & \heatmapAUROC{.68}{.45}{.77} $\pm$ .02 & \heatmapAUROC{.68}{.41}{.82} $\pm$ .02 & \heatmapAUROC{.57}{.32}{.95} $\pm$ .04 & \heatmapAUROC{.62}{.43}{1.00} $\pm$ .07 & \heatmapAUROC{.80}{.54}{1.00} $\pm$ .05 & 9.2 \\
AQA .9\, & \heatmapAUROC{.72}{.47}{.93} $\pm$ .06 & \heatmapAUROC{.61}{.61}{.87} $\pm$ .07 & \heatmapAUROC{.44}{.32}{1.00} $\pm$ .03 & \heatmapAUROC{.96}{.42}{.96} $\pm$ .02 & \heatmapAUROC{.41}{.45}{.89} $\pm$ .08 & \heatmapAUROC{.76}{.45}{.77} $\pm$ .02 & \heatmapAUROC{.79}{.41}{.82} $\pm$ .02 & \heatmapAUROC{.31}{.32}{.95} $\pm$ .05 & \heatmapAUROC{.42}{.43}{1.00} $\pm$ .07 & \heatmapAUROC{.60}{.54}{1.00} $\pm$ .05 & 9.3 \\
PLM 50\, & \heatmapAUROC{.55}{.47}{.93} $\pm$ .08 & \heatmapAUROC{.61}{.61}{.87} $\pm$ .07 & \heatmapAUROC{.42}{.32}{1.00} $\pm$ .02 & \heatmapAUROC{.96}{.42}{.96} $\pm$ .02 & \heatmapAUROC{.41}{.45}{.89} $\pm$ .07 & \heatmapAUROC{.69}{.45}{.77} $\pm$ .02 & \heatmapAUROC{.69}{.41}{.82} $\pm$ .03 & \heatmapAUROC{.56}{.32}{.95} $\pm$ .05 & \heatmapAUROC{.48}{.43}{1.00} $\pm$ .07 & \heatmapAUROC{.78}{.54}{1.00} $\pm$ .04 & 9.3 \\
AQA .75\, & \heatmapAUROC{.67}{.47}{.93} $\pm$ .07 & \heatmapAUROC{.61}{.61}{.87} $\pm$ .07 & \heatmapAUROC{.56}{.32}{1.00} $\pm$ .02 & \heatmapAUROC{.95}{.42}{.96} $\pm$ .02 & \heatmapAUROC{.41}{.45}{.89} $\pm$ .07 & \heatmapAUROC{.77}{.45}{.77} $\pm$ .02 & \heatmapAUROC{.78}{.41}{.82} $\pm$ .02 & \heatmapAUROC{.32}{.32}{.95} $\pm$ .05 & \heatmapAUROC{.45}{.43}{1.00} $\pm$ .07 & \heatmapAUROC{.62}{.54}{1.00} $\pm$ .05 & 9.4 \\
PLM 10\, & \heatmapAUROC{.50}{.47}{.93} $\pm$ .07 & \heatmapAUROC{.58}{.61}{.87} $\pm$ .08 & \heatmapAUROC{.35}{.32}{1.00} $\pm$ .02 & \heatmapAUROC{.90}{.42}{.96} $\pm$ .03 & \heatmapAUROC{.45}{.45}{.89} $\pm$ .08 & \heatmapAUROC{.66}{.45}{.77} $\pm$ .02 & \heatmapAUROC{.59}{.41}{.82} $\pm$ .03 & \heatmapAUROC{.50}{.32}{.95} $\pm$ .04 & \heatmapAUROC{.69}{.43}{1.00} $\pm$ .06 & \heatmapAUROC{.81}{.54}{1.00} $\pm$ .04 & 10.5 \\
ATA .5\, & \heatmapAUROC{.88}{.47}{.93} $\pm$ .04 & \heatmapAUROC{.82}{.61}{.87} $\pm$ .05 & \heatmapAUROC{.31}{.32}{1.00} $\pm$ .02 & \heatmapAUROC{.50}{.42}{.96} $\pm$ .00 & \heatmapAUROC{.50}{.45}{.89} $\pm$ .00 & \heatmapAUROC{.52}{.45}{.77} $\pm$ .03 & \heatmapAUROC{.63}{.41}{.82} $\pm$ .02 & \heatmapAUROC{.46}{.32}{.95} $\pm$ .04 & \heatmapAUROC{.43}{.43}{1.00} $\pm$ .06 & \heatmapAUROC{.35}{.54}{1.00} $\pm$ .05 & 10.5 \\
ATA .3\, & \heatmapAUROC{.87}{.47}{.93} $\pm$ .04 & \heatmapAUROC{.72}{.61}{.87} $\pm$ .06 & \heatmapAUROC{.52}{.32}{1.00} $\pm$ .02 & \heatmapAUROC{.65}{.42}{.96} $\pm$ .05 & \heatmapAUROC{.34}{.45}{.89} $\pm$ .06 & \heatmapAUROC{.60}{.45}{.77} $\pm$ .02 & \heatmapAUROC{.61}{.41}{.82} $\pm$ .02 & \heatmapAUROC{.44}{.32}{.95} $\pm$ .04 & \heatmapAUROC{.48}{.43}{1.00} $\pm$ .07 & \heatmapAUROC{.42}{.54}{1.00} $\pm$ .05 & 10.9 \\
ATA .7\, & \heatmapAUROC{.56}{.47}{.93} $\pm$ .08 & \heatmapAUROC{.50}{.61}{.87} $\pm$ .00 & \heatmapAUROC{.28}{.32}{1.00} $\pm$ .02 & \heatmapAUROC{.50}{.42}{.96} $\pm$ .00 & \heatmapAUROC{.50}{.45}{.89} $\pm$ .00 & \heatmapAUROC{.50}{.45}{.77} $\pm$ .00 & \heatmapAUROC{.64}{.41}{.82} $\pm$ .03 & \heatmapAUROC{.39}{.32}{.95} $\pm$ .04 & \heatmapAUROC{.50}{.43}{1.00} $\pm$ .00 & \heatmapAUROC{.50}{.54}{1.00} $\pm$ .00 & 12.9 \\
\addlinespace[5pt]
\multicolumn{1}{c}{} & \multicolumn{10}{l}{\textbf{b \qquad E-AURC}} & \\
\textbf{GMM-All}\, & \heatmapEAURC{.06}{.04}{.13} $\pm$ .01 & \heatmapEAURC{.03}{.02}{.04} $\pm$ .00 & \heatmapEAURC{.05}{.04}{.10} $\pm$ .01 & \heatmapEAURC{.07}{.05}{.20} $\pm$ .01 & \heatmapEAURC{.10}{.06}{.15} $\pm$ .02 & \heatmapEAURC{.19}{.16}{.33} $\pm$ .01 & \heatmapEAURC{.21}{.20}{.37} $\pm$ .01 & \heatmapEAURC{.12}{.08}{.32} $\pm$ .02 & \heatmapEAURC{.08}{.03}{.13} $\pm$ .01 & \heatmapEAURC{.04}{.04}{.13} $\pm$ .01 & \textbf{4.4} \\
\textbf{QFR}\, & \heatmapEAURC{.05}{.04}{.13} $\pm$ .01 & \heatmapEAURC{.02}{.02}{.04} $\pm$ .00 & \heatmapEAURC{.07}{.04}{.10} $\pm$ .01 & \heatmapEAURC{.06}{.05}{.20} $\pm$ .01 & \heatmapEAURC{.13}{.06}{.15} $\pm$ .04 & \heatmapEAURC{.28}{.16}{.33} $\pm$ .01 & \heatmapEAURC{.32}{.20}{.37} $\pm$ .01 & \heatmapEAURC{.16}{.08}{.32} $\pm$ .02 & \heatmapEAURC{.04}{.03}{.13} $\pm$ .01 & \heatmapEAURC{.04}{.04}{.13} $\pm$ .01 & \textbf{4.5} \\
\textbf{GMM-Int}\, & \heatmapEAURC{.07}{.04}{.13} $\pm$ .01 & \heatmapEAURC{.04}{.02}{.04} $\pm$ .01 & \heatmapEAURC{.08}{.04}{.10} $\pm$ .01 & \heatmapEAURC{.07}{.05}{.20} $\pm$ .01 & \heatmapEAURC{.11}{.06}{.15} $\pm$ .02 & \heatmapEAURC{.16}{.16}{.33} $\pm$ .01 & \heatmapEAURC{.22}{.20}{.37} $\pm$ .01 & \heatmapEAURC{.12}{.08}{.32} $\pm$ .02 & \heatmapEAURC{.08}{.03}{.13} $\pm$ .01 & \heatmapEAURC{.04}{.04}{.13} $\pm$ .01 & \textbf{5.1} \\
\textbf{BCA}\, & \heatmapEAURC{.05}{.04}{.13} $\pm$ .01 & \heatmapEAURC{.02}{.02}{.04} $\pm$ .00 & \heatmapEAURC{.03}{.04}{.10} $\pm$ .01 & \heatmapEAURC{.07}{.05}{.20} $\pm$ .01 & \heatmapEAURC{.14}{.06}{.15} $\pm$ .04 & \heatmapEAURC{.26}{.16}{.33} $\pm$ .01 & \heatmapEAURC{.35}{.20}{.37} $\pm$ .01 & \heatmapEAURC{.23}{.08}{.32} $\pm$ .02 & \heatmapEAURC{.07}{.03}{.13} $\pm$ .02 & \heatmapEAURC{.07}{.04}{.13} $\pm$ .01 & \textbf{5.7} \\
\textbf{GMM-Spa}\, & \heatmapEAURC{.05}{.04}{.13} $\pm$ .01 & \heatmapEAURC{.04}{.02}{.04} $\pm$ .01 & \heatmapEAURC{.04}{.04}{.10} $\pm$ .01 & \heatmapEAURC{.09}{.05}{.20} $\pm$ .01 & \heatmapEAURC{.09}{.06}{.15} $\pm$ .01 & \heatmapEAURC{.17}{.16}{.33} $\pm$ .01 & \heatmapEAURC{.24}{.20}{.37} $\pm$ .01 & \heatmapEAURC{.12}{.08}{.32} $\pm$ .02 & \heatmapEAURC{.09}{.03}{.13} $\pm$ .02 & \heatmapEAURC{.07}{.04}{.13} $\pm$ .01 & \textbf{6.1} \\
\addlinespace[3pt]
ATA .5\, & \heatmapEAURC{.04}{.04}{.13} $\pm$ .01 & \heatmapEAURC{.02}{.02}{.04} $\pm$ .00 & \heatmapEAURC{.10}{.04}{.10} $\pm$ .01 & \heatmapEAURC{.15}{.05}{.20} $\pm$ .01 & \heatmapEAURC{.11}{.06}{.15} $\pm$ .01 & \heatmapEAURC{.21}{.16}{.33} $\pm$ .01 & \heatmapEAURC{.29}{.20}{.37} $\pm$ .01 & \heatmapEAURC{.34}{.08}{.32} $\pm$ .02 & \heatmapEAURC{.05}{.03}{.13} $\pm$ .01 & \heatmapEAURC{.08}{.04}{.13} $\pm$ .01 & 7.1 \\
ATA .3\, & \heatmapEAURC{.04}{.04}{.13} $\pm$ .01 & \heatmapEAURC{.03}{.02}{.04} $\pm$ .01 & \heatmapEAURC{.10}{.04}{.10} $\pm$ .01 & \heatmapEAURC{.11}{.05}{.20} $\pm$ .01 & \heatmapEAURC{.15}{.06}{.15} $\pm$ .03 & \heatmapEAURC{.24}{.16}{.33} $\pm$ .01 & \heatmapEAURC{.30}{.20}{.37} $\pm$ .01 & \heatmapEAURC{.34}{.08}{.32} $\pm$ .02 & \heatmapEAURC{.03}{.03}{.13} $\pm$ .01 & \heatmapEAURC{.08}{.04}{.13} $\pm$ .01 & 8.1 \\
PLM 20\, & \heatmapEAURC{.08}{.04}{.13} $\pm$ .01 & \heatmapEAURC{.03}{.02}{.04} $\pm$ .01 & \heatmapEAURC{.10}{.04}{.10} $\pm$ .01 & \heatmapEAURC{.07}{.05}{.20} $\pm$ .01 & \heatmapEAURC{.15}{.06}{.15} $\pm$ .04 & \heatmapEAURC{.28}{.16}{.33} $\pm$ .01 & \heatmapEAURC{.31}{.20}{.37} $\pm$ .01 & \heatmapEAURC{.34}{.08}{.32} $\pm$ .02 & \heatmapEAURC{.09}{.03}{.13} $\pm$ .03 & \heatmapEAURC{.05}{.04}{.13} $\pm$ .01 & 8.8 \\
PLM 10\, & \heatmapEAURC{.12}{.04}{.13} $\pm$ .01 & \heatmapEAURC{.03}{.02}{.04} $\pm$ .00 & \heatmapEAURC{.10}{.04}{.10} $\pm$ .01 & \heatmapEAURC{.06}{.05}{.20} $\pm$ .01 & \heatmapEAURC{.15}{.06}{.15} $\pm$ .04 & \heatmapEAURC{.27}{.16}{.33} $\pm$ .01 & \heatmapEAURC{.29}{.20}{.37} $\pm$ .01 & \heatmapEAURC{.34}{.08}{.32} $\pm$ .02 & \heatmapEAURC{.10}{.03}{.13} $\pm$ .04 & \heatmapEAURC{.05}{.04}{.13} $\pm$ .01 & 9.0 \\
ICA\, & \heatmapEAURC{.10}{.04}{.13} $\pm$ .01 & \heatmapEAURC{.04}{.02}{.04} $\pm$ .01 & \heatmapEAURC{.06}{.04}{.10} $\pm$ .01 & \heatmapEAURC{.07}{.05}{.20} $\pm$ .01 & \heatmapEAURC{.14}{.06}{.15} $\pm$ .04 & \heatmapEAURC{.33}{.16}{.33} $\pm$ .01 & \heatmapEAURC{.36}{.20}{.37} $\pm$ .01 & \heatmapEAURC{.34}{.08}{.32} $\pm$ .02 & \heatmapEAURC{.07}{.03}{.13} $\pm$ .02 & \heatmapEAURC{.07}{.04}{.13} $\pm$ .01 & 9.6 \\
PLM 50\, & \heatmapEAURC{.10}{.04}{.13} $\pm$ .01 & \heatmapEAURC{.03}{.02}{.04} $\pm$ .00 & \heatmapEAURC{.10}{.04}{.10} $\pm$ .01 & \heatmapEAURC{.07}{.05}{.20} $\pm$ .01 & \heatmapEAURC{.16}{.06}{.15} $\pm$ .04 & \heatmapEAURC{.29}{.16}{.33} $\pm$ .01 & \heatmapEAURC{.32}{.20}{.37} $\pm$ .01 & \heatmapEAURC{.34}{.08}{.32} $\pm$ .02 & \heatmapEAURC{.08}{.03}{.13} $\pm$ .03 & \heatmapEAURC{.06}{.04}{.13} $\pm$ .01 & 10.0 \\
AQA .6\, & \heatmapEAURC{.08}{.04}{.13} $\pm$ .01 & \heatmapEAURC{.04}{.02}{.04} $\pm$ .01 & \heatmapEAURC{.08}{.04}{.10} $\pm$ .01 & \heatmapEAURC{.07}{.05}{.20} $\pm$ .01 & \heatmapEAURC{.15}{.06}{.15} $\pm$ .04 & \heatmapEAURC{.33}{.16}{.33} $\pm$ .01 & \heatmapEAURC{.35}{.20}{.37} $\pm$ .01 & \heatmapEAURC{.34}{.08}{.32} $\pm$ .02 & \heatmapEAURC{.06}{.03}{.13} $\pm$ .02 & \heatmapEAURC{.08}{.04}{.13} $\pm$ .01 & 10.7 \\
AQA .9\, & \heatmapEAURC{.08}{.04}{.13} $\pm$ .01 & \heatmapEAURC{.03}{.02}{.04} $\pm$ .00 & \heatmapEAURC{.10}{.04}{.10} $\pm$ .01 & \heatmapEAURC{.07}{.05}{.20} $\pm$ .01 & \heatmapEAURC{.16}{.06}{.15} $\pm$ .04 & \heatmapEAURC{.32}{.16}{.33} $\pm$ .01 & \heatmapEAURC{.36}{.20}{.37} $\pm$ .01 & \heatmapEAURC{.35}{.08}{.32} $\pm$ .02 & \heatmapEAURC{.05}{.03}{.13} $\pm$ .02 & \heatmapEAURC{.09}{.04}{.13} $\pm$ .01 & 10.8 \\
AVG\, & \heatmapEAURC{.10}{.04}{.13} $\pm$ .02 & \heatmapEAURC{.04}{.02}{.04} $\pm$ .01 & \heatmapEAURC{.06}{.04}{.10} $\pm$ .01 & \heatmapEAURC{.07}{.05}{.20} $\pm$ .01 & \heatmapEAURC{.15}{.06}{.15} $\pm$ .04 & \heatmapEAURC{.33}{.16}{.33} $\pm$ .01 & \heatmapEAURC{.36}{.20}{.37} $\pm$ .01 & \heatmapEAURC{.33}{.08}{.32} $\pm$ .02 & \heatmapEAURC{.06}{.03}{.13} $\pm$ .02 & \heatmapEAURC{.08}{.04}{.13} $\pm$ .01 & 10.9 \\
AQA .75\, & \heatmapEAURC{.08}{.04}{.13} $\pm$ .01 & \heatmapEAURC{.04}{.02}{.04} $\pm$ .01 & \heatmapEAURC{.09}{.04}{.10} $\pm$ .01 & \heatmapEAURC{.07}{.05}{.20} $\pm$ .01 & \heatmapEAURC{.15}{.06}{.15} $\pm$ .04 & \heatmapEAURC{.33}{.16}{.33} $\pm$ .01 & \heatmapEAURC{.36}{.20}{.37} $\pm$ .01 & \heatmapEAURC{.32}{.08}{.32} $\pm$ .02 & \heatmapEAURC{.06}{.03}{.13} $\pm$ .02 & \heatmapEAURC{.09}{.04}{.13} $\pm$ .01 & 11.1 \\
ATA .7\, & \heatmapEAURC{.10}{.04}{.13} $\pm$ .02 & \heatmapEAURC{.05}{.02}{.04} $\pm$ .00 & \heatmapEAURC{.11}{.04}{.10} $\pm$ .01 & \heatmapEAURC{.15}{.05}{.20} $\pm$ .01 & \heatmapEAURC{.11}{.06}{.15} $\pm$ .01 & \heatmapEAURC{.17}{.16}{.33} $\pm$ .01 & \heatmapEAURC{.29}{.20}{.37} $\pm$ .01 & \heatmapEAURC{.35}{.08}{.32} $\pm$ .02 & \heatmapEAURC{.11}{.03}{.13} $\pm$ .02 & \heatmapEAURC{.09}{.04}{.13} $\pm$ .01 & 11.1 \\
\addlinespace[29pt]
\multicolumn{1}{c}{} & \multicolumn{1}{c}{\smash{\rotatebox[origin=lt]{55}{ARC-BC}}} & \multicolumn{1}{c}{\smash{\rotatebox[origin=lt]{55}{ARC-Nuc}}} & \multicolumn{1}{c}{\smash{\rotatebox[origin=lt]{55}{CAR-CS}}} & \multicolumn{1}{c}{\smash{\rotatebox[origin=lt]{55}{LIDC-Mal}}} & \multicolumn{1}{c}{\smash{\rotatebox[origin=lt]{55}{LIDC-Tex}}} & \multicolumn{1}{c}{\smash{\rotatebox[origin=lt]{55}{LIZ-IG}}} & \multicolumn{1}{c}{\smash{\rotatebox[origin=lt]{55}{LIZ-SG}}} & \multicolumn{1}{c}{\smash{\rotatebox[origin=lt]{55}{WEED-Hand}}} & \multicolumn{1}{c}{\smash{\rotatebox[origin=lt]{55}{WORM-Nem}}} & \multicolumn{1}{c}{\smash{\rotatebox[origin=lt]{55}{WORM-Pro}}} & \\
\end{tabular}%
}
    \caption{\textbf{\Aggs{} performance on uncertainty maps generated with Maximum Softmax Probability (MSP) in \ood{} and failure detection.} Color coding and metric interpretation are as detailed in Table A.1. (a)-(b) \Aggs{} are ranked from best (top) to worst (bottom) based on their average metric, first computed across 500 bootstrap samples per dataset, and then averaged across datasets to ensure stable rankings. Each value is reported with the std. deviation across bootstrap samples.}
    \label{tab:msp_boostrapped_results}
\end{table}

\subsection{On the \Agg{} performance in Failure Detection}
In line with the numerical and qualitative results discussed in Section \ref{sec:fd_res}, Table \ref{tab:msp_boostrapped_results}b shows similar trends for \Aggs{} applied on MSP uncertainty heatmaps. The top-performing predictors remain QFR and BCA, both \textit{prediction-aware}, and \emph{GMM-} meta-aggregators, while ICA continues to underperform due to repeated segmentation errors in small objects. Table~\ref{tab:ensemble_bootstrapped_results}b and Table ~\ref{tab:tta_bootstrapped_results}b reveal no major deviations in their restricted analysis subsets, with one exception: the patch-based PLM20 ranks among the top five, owing to the localized uncertainty maps from test-time augmentations that closely match segmentation error regions. However, given the limited subset of data analyzed for TTA heatmaps, no general conclusions can be drawn regarding the UQ method.  \newline \noindent 
For TTA, \Aggs{} perform better in the three-label classification setting (\eg ARC-Nuc). For MSP, however, performance is also negatively impacted when uncertainty maps show increased uncertainty starting from the borders and extending within the predicted mask (\eg in LIDC-Tex), where the \ood{} variant exhibits inconsistent patterns. This demonstrates that in this case augmentations alone are insufficient to capture the uncertainty nuances between \iid{} and \ood{} samples, while MSP tends to be overconfident in these regions. \newline \noindent
Analysis of the p-value matrices reveals that QFR emerges as the top-performing aggregator for Failure Detection, showing statistically significant improvements over all other methods ($p < 0.05$ in all pairwise comparisons, with $p < 0.001$ in most cases). The next best group includes the GMM-based scores and BCA, which consistently outperform intensity based aggregators but are statistically indistinguishable from one another, indicating a robust, if slightly weaker, alternative to QFR. By contrast, threshold-based methods (ATA, AQA) and patch-based approaches (PLM) exhibit significantly lower performance in most tests, while the global average (AVG) consistently ranks lowest, confirming it should not be used as a default strategy. \newline \noindent 
These patterns are corroborated when examining \Aggs{} on MSP, TTA, and ensemble-based uncertainty heatmaps. Across all three, QFR’s dominance is consistently confirmed, with pairwise p-values frequently approaching machine precision ($p \ll 0.001$), establishing it as the unequivocal top-tier \Agg{} for FD across all tested uncertainty frameworks. The second tier under MCD, consisting of \textit{GMM}-variants and BCA, remains stable and consistently outperforms lower-tier aggregators - thresholding (AQA, ATA), patch-based (PLM), and averaging (AVG) - which are statistically inferior across nearly all datasets and tests. Collectively, these results validate the rankings reported in ~\Cref{fig:fd_results}b and underscore the importance of \textit{prediction-aware}, structure-sensitive aggregators for effective FD.

\begin{table}[t]
    \centering
    \setlength{\tabcolsep}{2.3pt}
    \renewcommand{\arraystretch}{1.75}
    \scriptsize
    \newcommand{\headerbox}[1]{%
        \tikz[baseline=(X.base)] \node[draw=gray, fill=white, rounded corners=4pt, line width=0.75pt, inner sep=3pt, minimum width=5.5em,] (X) {\strut#1};%
    }
    \scalebox{0.65}{%
\begin{tabular}{rccccc|c}
\multicolumn{1}{c}{} & \multicolumn{5}{l}{\textbf{a \qquad AUROC}} & \\
\textbf{GMM-All}\, & \heatmapAUROC{.83}{.47}{.93} $\pm$ .05 & \heatmapAUROC{.84}{.61}{.87} $\pm$ .06 & \heatmapAUROC{.99}{.32}{1.00} $\pm$ .00 & \heatmapAUROC{.86}{.42}{.96} $\pm$ .03 & \heatmapAUROC{.70}{.45}{.89} $\pm$ .07 & \textbf{4.8} \\
\textbf{QFR}\, & \heatmapAUROC{.90}{.47}{.93} $\pm$ .04 & \heatmapAUROC{.81}{.61}{.87} $\pm$ .06 & \heatmapAUROC{.67}{.32}{1.00} $\pm$ .02 & \heatmapAUROC{.77}{.42}{.96} $\pm$ .04 & \heatmapAUROC{.88}{.45}{.89} $\pm$ .04 & \textbf{4.8} \\
\textbf{GMM-Spa}\, & \heatmapAUROC{.93}{.47}{.93} $\pm$ .03 & \heatmapAUROC{.72}{.61}{.87} $\pm$ .06 & \heatmapAUROC{1.00}{.32}{1.00} $\pm$ .00 & \heatmapAUROC{.83}{.42}{.96} $\pm$ .03 & \heatmapAUROC{.70}{.45}{.89} $\pm$ .06 & \textbf{5.0} \\
\textbf{BCA}\, & \heatmapAUROC{.89}{.47}{.93} $\pm$ .04 & \heatmapAUROC{.76}{.61}{.87} $\pm$ .06 & \heatmapAUROC{.93}{.32}{1.00} $\pm$ .01 & \heatmapAUROC{.71}{.42}{.96} $\pm$ .04 & \heatmapAUROC{.82}{.45}{.89} $\pm$ .06 & \textbf{5.2} \\
\textbf{ICA}\, & \heatmapAUROC{.88}{.47}{.93} $\pm$ .05 & \heatmapAUROC{.79}{.61}{.87} $\pm$ .06 & \heatmapAUROC{.89}{.32}{1.00} $\pm$ .01 & \heatmapAUROC{.70}{.42}{.96} $\pm$ .04 & \heatmapAUROC{.82}{.45}{.89} $\pm$ .07 & \textbf{5.4} \\
\textbf{GMM-Int}\, & \heatmapAUROC{.79}{.47}{.93} $\pm$ .05 & \heatmapAUROC{.76}{.61}{.87} $\pm$ .06 & \heatmapAUROC{.79}{.32}{1.00} $\pm$ .02 & \heatmapAUROC{.89}{.42}{.96} $\pm$ .03 & \heatmapAUROC{.68}{.45}{.89} $\pm$ .06 & \textbf{6.4} \\
\addlinespace[3pt]
AQA .9\, & \heatmapAUROC{.86}{.47}{.93} $\pm$ .05 & \heatmapAUROC{.64}{.61}{.87} $\pm$ .07 & \heatmapAUROC{.42}{.32}{1.00} $\pm$ .03 & \heatmapAUROC{.96}{.42}{.96} $\pm$ .01 & \heatmapAUROC{.57}{.45}{.89} $\pm$ .07 & 8.6 \\
PLM 20\, & \heatmapAUROC{.78}{.47}{.93} $\pm$ .05 & \heatmapAUROC{.63}{.61}{.87} $\pm$ .07 & \heatmapAUROC{.39}{.32}{1.00} $\pm$ .02 & \heatmapAUROC{.96}{.42}{.96} $\pm$ .01 & \heatmapAUROC{.58}{.45}{.89} $\pm$ .07 & 8.8 \\
AQA .75\, & \heatmapAUROC{.75}{.47}{.93} $\pm$ .06 & \heatmapAUROC{.61}{.61}{.87} $\pm$ .07 & \heatmapAUROC{.60}{.32}{1.00} $\pm$ .02 & \heatmapAUROC{.96}{.42}{.96} $\pm$ .01 & \heatmapAUROC{.58}{.45}{.89} $\pm$ .07 & 9.0 \\
AQA .6\, & \heatmapAUROC{.71}{.47}{.93} $\pm$ .06 & \heatmapAUROC{.61}{.61}{.87} $\pm$ .07 & \heatmapAUROC{.67}{.32}{1.00} $\pm$ .02 & \heatmapAUROC{.96}{.42}{.96} $\pm$ .01 & \heatmapAUROC{.57}{.45}{.89} $\pm$ .07 & 9.4 \\
PLM 50\, & \heatmapAUROC{.65}{.47}{.93} $\pm$ .08 & \heatmapAUROC{.63}{.61}{.87} $\pm$ .07 & \heatmapAUROC{.44}{.32}{1.00} $\pm$ .02 & \heatmapAUROC{.96}{.42}{.96} $\pm$ .01 & \heatmapAUROC{.57}{.45}{.89} $\pm$ .07 & 9.8 \\
AVG\, & \heatmapAUROC{.62}{.47}{.93} $\pm$ .06 & \heatmapAUROC{.59}{.61}{.87} $\pm$ .07 & \heatmapAUROC{.78}{.32}{1.00} $\pm$ .02 & \heatmapAUROC{.96}{.42}{.96} $\pm$ .01 & \heatmapAUROC{.57}{.45}{.89} $\pm$ .07 & 10.2 \\
PLM 10\, & \heatmapAUROC{.66}{.47}{.93} $\pm$ .06 & \heatmapAUROC{.58}{.61}{.87} $\pm$ .08 & \heatmapAUROC{.38}{.32}{1.00} $\pm$ .03 & \heatmapAUROC{.93}{.42}{.96} $\pm$ .02 & \heatmapAUROC{.65}{.45}{.89} $\pm$ .07 & 11.2 \\
ATA .5\, & \heatmapAUROC{.84}{.47}{.93} $\pm$ .05 & \heatmapAUROC{.76}{.61}{.87} $\pm$ .06 & \heatmapAUROC{.25}{.32}{1.00} $\pm$ .02 & \heatmapAUROC{.51}{.42}{.96} $\pm$ .06 & \heatmapAUROC{.40}{.45}{.89} $\pm$ .08 & 11.6 \\
ATA .3\, & \heatmapAUROC{.87}{.47}{.93} $\pm$ .05 & \heatmapAUROC{.62}{.61}{.87} $\pm$ .07 & \heatmapAUROC{.30}{.32}{1.00} $\pm$ .02 & \heatmapAUROC{.60}{.42}{.96} $\pm$ .05 & \heatmapAUROC{.55}{.45}{.89} $\pm$ .08 & 11.8 \\
ATA .7\, & \heatmapAUROC{.42}{.47}{.93} $\pm$ .07 & \heatmapAUROC{.74}{.61}{.87} $\pm$ .06 & \heatmapAUROC{.28}{.32}{1.00} $\pm$ .02 & \heatmapAUROC{.42}{.42}{.96} $\pm$ .05 & \heatmapAUROC{.32}{.45}{.89} $\pm$ .06 & 14.0 \\
\addlinespace[5pt]
\multicolumn{1}{c}{} & \multicolumn{5}{l}{\textbf{b \qquad E-AURC}} & \\
\textbf{QFR}\, & \heatmapEAURC{.04}{.04}{.13} $\pm$ .01 & \heatmapEAURC{.02}{.02}{.04} $\pm$ .00 & \heatmapEAURC{.06}{.04}{.10} $\pm$ .01 & \heatmapEAURC{.04}{.05}{.20} $\pm$ .00 & \heatmapEAURC{.04}{.06}{.15} $\pm$ .01 & \textbf{2.9} \\
\textbf{GMM-Spa}\, & \heatmapEAURC{.05}{.04}{.13} $\pm$ .01 & \heatmapEAURC{.04}{.02}{.04} $\pm$ .01 & \heatmapEAURC{.05}{.04}{.10} $\pm$ .00 & \heatmapEAURC{.05}{.05}{.20} $\pm$ .01 & \heatmapEAURC{.06}{.06}{.15} $\pm$ .01 & \textbf{4.4} \\
\textbf{GMM-All}\, & \heatmapEAURC{.05}{.04}{.13} $\pm$ .01 & \heatmapEAURC{.03}{.02}{.04} $\pm$ .00 & \heatmapEAURC{.06}{.04}{.10} $\pm$ .00 & \heatmapEAURC{.07}{.05}{.20} $\pm$ .01 & \heatmapEAURC{.06}{.06}{.15} $\pm$ .01 & \textbf{5.7} \\
\textbf{BCA}\, & \heatmapEAURC{.05}{.04}{.13} $\pm$ .01 & \heatmapEAURC{.03}{.02}{.04} $\pm$ .00 & \heatmapEAURC{.04}{.04}{.10} $\pm$ .00 & \heatmapEAURC{.05}{.05}{.20} $\pm$ .01 & \heatmapEAURC{.08}{.06}{.15} $\pm$ .01 & \textbf{6.4} \\
\textbf{GMM-Int}\, & \heatmapEAURC{.06}{.04}{.13} $\pm$ .01 & \heatmapEAURC{.04}{.02}{.04} $\pm$ .01 & \heatmapEAURC{.08}{.04}{.10} $\pm$ .01 & \heatmapEAURC{.06}{.05}{.20} $\pm$ .01 & \heatmapEAURC{.06}{.06}{.15} $\pm$ .01 & \textbf{7.0} \\
\addlinespace[3pt]
PLM 10\, & \heatmapEAURC{.10}{.04}{.13} $\pm$ .02 & \heatmapEAURC{.02}{.02}{.04} $\pm$ .00 & \heatmapEAURC{.09}{.04}{.10} $\pm$ .01 & \heatmapEAURC{.05}{.05}{.20} $\pm$ .01 & \heatmapEAURC{.07}{.06}{.15} $\pm$ .01 & 7.6 \\
PLM 20\, & \heatmapEAURC{.07}{.04}{.13} $\pm$ .01 & \heatmapEAURC{.03}{.02}{.04} $\pm$ .00 & \heatmapEAURC{.09}{.04}{.10} $\pm$ .01 & \heatmapEAURC{.05}{.05}{.20} $\pm$ .01 & \heatmapEAURC{.08}{.06}{.15} $\pm$ .01 & 8.1 \\
ATA .3\, & \heatmapEAURC{.04}{.04}{.13} $\pm$ .01 & \heatmapEAURC{.03}{.02}{.04} $\pm$ .00 & \heatmapEAURC{.10}{.04}{.10} $\pm$ .01 & \heatmapEAURC{.12}{.05}{.20} $\pm$ .02 & \heatmapEAURC{.12}{.06}{.15} $\pm$ .03 & 9.1 \\
AQA .9\, & \heatmapEAURC{.05}{.04}{.13} $\pm$ .01 & \heatmapEAURC{.03}{.02}{.04} $\pm$ .00 & \heatmapEAURC{.09}{.04}{.10} $\pm$ .01 & \heatmapEAURC{.06}{.05}{.20} $\pm$ .01 & \heatmapEAURC{.08}{.06}{.15} $\pm$ .01 & 9.1 \\
ATA .5\, & \heatmapEAURC{.05}{.04}{.13} $\pm$ .01 & \heatmapEAURC{.02}{.02}{.04} $\pm$ .00 & \heatmapEAURC{.11}{.04}{.10} $\pm$ .01 & \heatmapEAURC{.16}{.05}{.20} $\pm$ .03 & \heatmapEAURC{.17}{.06}{.15} $\pm$ .04 & 9.3 \\
PLM 50\, & \heatmapEAURC{.07}{.04}{.13} $\pm$ .01 & \heatmapEAURC{.03}{.02}{.04} $\pm$ .00 & \heatmapEAURC{.09}{.04}{.10} $\pm$ .01 & \heatmapEAURC{.06}{.05}{.20} $\pm$ .01 & \heatmapEAURC{.08}{.06}{.15} $\pm$ .01 & 9.4 \\
ATA .7\, & \heatmapEAURC{.14}{.04}{.13} $\pm$ .02 & \heatmapEAURC{.01}{.02}{.04} $\pm$ .00 & \heatmapEAURC{.10}{.04}{.10} $\pm$ .01 & \heatmapEAURC{.23}{.05}{.20} $\pm$ .04 & \heatmapEAURC{.18}{.06}{.15} $\pm$ .04 & 10.3 \\
AQA .75\, & \heatmapEAURC{.07}{.04}{.13} $\pm$ .01 & \heatmapEAURC{.03}{.02}{.04} $\pm$ .01 & \heatmapEAURC{.08}{.04}{.10} $\pm$ .01 & \heatmapEAURC{.06}{.05}{.20} $\pm$ .01 & \heatmapEAURC{.08}{.06}{.15} $\pm$ .01 & 10.6 \\
AQA .6\, & \heatmapEAURC{.08}{.04}{.13} $\pm$ .01 & \heatmapEAURC{.04}{.02}{.04} $\pm$ .01 & \heatmapEAURC{.08}{.04}{.10} $\pm$ .01 & \heatmapEAURC{.06}{.05}{.20} $\pm$ .01 & \heatmapEAURC{.08}{.06}{.15} $\pm$ .01 & 10.9 \\
AVG\, & \heatmapEAURC{.10}{.04}{.13} $\pm$ .02 & \heatmapEAURC{.04}{.02}{.04} $\pm$ .01 & \heatmapEAURC{.07}{.04}{.10} $\pm$ .01 & \heatmapEAURC{.06}{.05}{.20} $\pm$ .01 & \heatmapEAURC{.08}{.06}{.15} $\pm$ .01 & 11.3 \\
ICA\, & \heatmapEAURC{.10}{.04}{.13} $\pm$ .02 & \heatmapEAURC{.04}{.02}{.04} $\pm$ .00 & \heatmapEAURC{.06}{.04}{.10} $\pm$ .01 & \heatmapEAURC{.05}{.05}{.20} $\pm$ .01 & \heatmapEAURC{.08}{.06}{.15} $\pm$ .01 & 11.3 \\
\addlinespace[23pt]
\multicolumn{1}{c}{} & \multicolumn{1}{c}{\smash{\rotatebox[origin=lt]{55}{ARC-BC}}} & \multicolumn{1}{c}{\smash{\rotatebox[origin=lt]{55}{ARC-Nuc}}} & \multicolumn{1}{c}{\smash{\rotatebox[origin=lt]{55}{CAR-CS}}} & \multicolumn{1}{c}{\smash{\rotatebox[origin=lt]{55}{LIDC-Mal}}} & \multicolumn{1}{c}{\smash{\rotatebox[origin=lt]{55}{LIDC-Tex}}} & \\
\end{tabular}%
}
    \caption{\textbf{\Aggs{} performance on ensembling heatmaps in \ood{} and failure detection.} Color coding and metric interpretation are as detailed in Table A.1. (a)-(b) \Aggs{} are ranked from best (top) to worst (bottom) based on their average metric, first computed across 500 bootstrap samples per dataset, and then averaged across datasets to ensure stable rankings. Each value is reported with the std. deviation across bootstrap samples.}
    \label{tab:ensemble_bootstrapped_results}
\end{table}

\subsection{On the \Agg{} performance across downstream tasks} 
Mean ranks, metric values, and standard deviations reported in Table \ref{tab:mdc_bootstrap_results}, Table \ref{tab:msp_boostrapped_results}, Table \ref{tab:ensemble_bootstrapped_results} and Table \ref{tab:tta_bootstrapped_results} imply that, in the absence of structural knowledge about the \ood{} dataset, aggregators possessing the theoretical property of \textit{proportion invariance} (cf. Supp.~\ref{app:formal_properties}) and defined as \textit{prediction-aware} can be reliably selected for both \ood{} Detection and Failure Detection tasks (with the exception of ICA in FD). This conclusion also applies to our proposed \emph{GMM-All} and its ablated variants, \emph{GMM-Spa} and \emph{GMM-Int}, which provide a robust alternative across datasets when a dataset- or task-specific choice is uncertain.

\begin{table}[t]
    \centering
    \setlength{\tabcolsep}{2.3pt}
    \renewcommand{\arraystretch}{1.75}
    \scriptsize
    \newcommand{\headerbox}[1]{%
        \tikz[baseline=(X.base)] \node[draw=gray, fill=white, rounded corners=4pt, line width=0.75pt, inner sep=3pt, minimum width=5.5em,] (X) {\strut#1};%
    }
    \scalebox{0.67}{%
\begin{tabular}{rcccc|c}
\multicolumn{1}{c}{} & \multicolumn{4}{l}{\textbf{a \qquad AUROC}} & \\
\textbf{QFR}\, & \heatmapAUROC{.85}{.41}{.90} $\pm$ .05 & \heatmapAUROC{.88}{.60}{.88} $\pm$ .04 & \heatmapAUROC{.68}{.46}{.95} $\pm$ .04 & \heatmapAUROC{.81}{.33}{.88} $\pm$ .07 & \textbf{4.5} \\
\textbf{BCA}\, & \heatmapAUROC{.81}{.41}{.90} $\pm$ .06 & \heatmapAUROC{.87}{.60}{.88} $\pm$ .04 & \heatmapAUROC{.70}{.46}{.95} $\pm$ .04 & \heatmapAUROC{.73}{.33}{.88} $\pm$ .09 & \textbf{5.0} \\
\textbf{GMM-All}\, & \heatmapAUROC{.72}{.41}{.90} $\pm$ .07 & \heatmapAUROC{.88}{.60}{.88} $\pm$ .05 & \heatmapAUROC{.89}{.46}{.95} $\pm$ .03 & \heatmapAUROC{.73}{.33}{.88} $\pm$ .05 & \textbf{5.0} \\
\textbf{ICA}\, & \heatmapAUROC{.74}{.41}{.90} $\pm$ .06 & \heatmapAUROC{.88}{.60}{.88} $\pm$ .04 & \heatmapAUROC{.70}{.46}{.95} $\pm$ .04 & \heatmapAUROC{.72}{.33}{.88} $\pm$ .09 & \textbf{5.5} \\
\textbf{AQA .9}\, & \heatmapAUROC{.81}{.41}{.90} $\pm$ .05 & \heatmapAUROC{.74}{.60}{.88} $\pm$ .06 & \heatmapAUROC{.95}{.46}{.95} $\pm$ .02 & \heatmapAUROC{.51}{.33}{.88} $\pm$ .06 & \textbf{6.5} \\
\addlinespace[3pt]
PLM 20\, & \heatmapAUROC{.73}{.41}{.90} $\pm$ .06 & \heatmapAUROC{.78}{.60}{.88} $\pm$ .06 & \heatmapAUROC{.95}{.46}{.95} $\pm$ .02 & \heatmapAUROC{.50}{.33}{.88} $\pm$ .07 & 6.8 \\
GMM-Int\, & \heatmapAUROC{.65}{.41}{.90} $\pm$ .06 & \heatmapAUROC{.84}{.60}{.88} $\pm$ .05 & \heatmapAUROC{.90}{.46}{.95} $\pm$ .03 & \heatmapAUROC{.72}{.33}{.88} $\pm$ .05 & 7.8 \\
AQA .75\, & \heatmapAUROC{.72}{.41}{.90} $\pm$ .06 & \heatmapAUROC{.66}{.60}{.88} $\pm$ .07 & \heatmapAUROC{.95}{.46}{.95} $\pm$ .02 & \heatmapAUROC{.51}{.33}{.88} $\pm$ .07 & 8.8 \\
AQA .6\, & \heatmapAUROC{.66}{.41}{.90} $\pm$ .07 & \heatmapAUROC{.64}{.60}{.88} $\pm$ .07 & \heatmapAUROC{.95}{.46}{.95} $\pm$ .02 & \heatmapAUROC{.51}{.33}{.88} $\pm$ .07 & 9.8 \\
PLM 10\, & \heatmapAUROC{.44}{.41}{.90} $\pm$ .07 & \heatmapAUROC{.75}{.60}{.88} $\pm$ .06 & \heatmapAUROC{.90}{.46}{.95} $\pm$ .03 & \heatmapAUROC{.60}{.33}{.88} $\pm$ .08 & 9.8 \\
PLM 50\, & \heatmapAUROC{.61}{.41}{.90} $\pm$ .07 & \heatmapAUROC{.75}{.60}{.88} $\pm$ .06 & \heatmapAUROC{.95}{.46}{.95} $\pm$ .02 & \heatmapAUROC{.51}{.33}{.88} $\pm$ .07 & 9.8 \\
GMM-Spa\, & \heatmapAUROC{.69}{.41}{.90} $\pm$ .08 & \heatmapAUROC{.60}{.60}{.88} $\pm$ .08 & \heatmapAUROC{.78}{.46}{.95} $\pm$ .04 & \heatmapAUROC{.68}{.33}{.88} $\pm$ .06 & 10.2 \\
ATA .3\, & \heatmapAUROC{.81}{.41}{.90} $\pm$ .05 & \heatmapAUROC{.67}{.60}{.88} $\pm$ .06 & \heatmapAUROC{.64}{.46}{.95} $\pm$ .05 & \heatmapAUROC{.46}{.33}{.88} $\pm$ .07 & 10.8 \\
AVG\, & \heatmapAUROC{.54}{.41}{.90} $\pm$ .07 & \heatmapAUROC{.62}{.60}{.88} $\pm$ .07 & \heatmapAUROC{.95}{.46}{.95} $\pm$ .02 & \heatmapAUROC{.50}{.33}{.88} $\pm$ .07 & 11.0 \\
ATA .5\, & \heatmapAUROC{.67}{.41}{.90} $\pm$ .07 & \heatmapAUROC{.81}{.60}{.88} $\pm$ .05 & \heatmapAUROC{.55}{.46}{.95} $\pm$ .05 & \heatmapAUROC{.41}{.33}{.88} $\pm$ .06 & 11.8 \\
ATA .7\, & \heatmapAUROC{.41}{.41}{.90} $\pm$ .07 & \heatmapAUROC{.84}{.60}{.88} $\pm$ .05 & \heatmapAUROC{.46}{.46}{.95} $\pm$ .06 & \heatmapAUROC{.33}{.33}{.88} $\pm$ .06 & 13.2 \\
\addlinespace[5pt]
\multicolumn{1}{c}{} & \multicolumn{4}{l}{\textbf{b \qquad E-AURC}} & \\
\textbf{QFR}\, & \heatmapEAURC{.06}{.04}{.13} $\pm$ .01 & \heatmapEAURC{.02}{.02}{.04} $\pm$ .00 & \heatmapEAURC{.04}{.04}{.12} $\pm$ .01 & \heatmapEAURC{.09}{.06}{.15} $\pm$ .04 & \textbf{3.3} \\
\textbf{GMM-All}\, & \heatmapEAURC{.07}{.04}{.13} $\pm$ .01 & \heatmapEAURC{.03}{.02}{.04} $\pm$ .00 & \heatmapEAURC{.07}{.04}{.12} $\pm$ .01 & \heatmapEAURC{.06}{.06}{.15} $\pm$ .01 & \textbf{5.2} \\
\textbf{GMM-Spa}\, & \heatmapEAURC{.07}{.04}{.13} $\pm$ .01 & \heatmapEAURC{.04}{.02}{.04} $\pm$ .01 & \heatmapEAURC{.06}{.04}{.12} $\pm$ .01 & \heatmapEAURC{.06}{.06}{.15} $\pm$ .01 & \textbf{5.8} \\
\textbf{GMM-Int}\, & \heatmapEAURC{.08}{.04}{.13} $\pm$ .01 & \heatmapEAURC{.03}{.02}{.04} $\pm$ .00 & \heatmapEAURC{.07}{.04}{.12} $\pm$ .01 & \heatmapEAURC{.07}{.06}{.15} $\pm$ .01 & \textbf{6.2} \\
\textbf{PLM 20}\, & \heatmapEAURC{.07}{.04}{.13} $\pm$ .01 & \heatmapEAURC{.02}{.02}{.04} $\pm$ .00 & \heatmapEAURC{.06}{.04}{.12} $\pm$ .01 & \heatmapEAURC{.12}{.06}{.15} $\pm$ .04 & \textbf{7.2} \\
\addlinespace[3pt]
BCA\, & \heatmapEAURC{.05}{.04}{.13} $\pm$ .01 & \heatmapEAURC{.04}{.02}{.04} $\pm$ .00 & \heatmapEAURC{.05}{.04}{.12} $\pm$ .01 & \heatmapEAURC{.11}{.06}{.15} $\pm$ .03 & 7.5 \\
PLM 10\, & \heatmapEAURC{.10}{.04}{.13} $\pm$ .01 & \heatmapEAURC{.02}{.02}{.04} $\pm$ .00 & \heatmapEAURC{.05}{.04}{.12} $\pm$ .01 & \heatmapEAURC{.12}{.06}{.15} $\pm$ .04 & 7.2 \\
ATA .5\, & \heatmapEAURC{.05}{.04}{.13} $\pm$ .01 & \heatmapEAURC{.02}{.02}{.04} $\pm$ .00 & \heatmapEAURC{.12}{.04}{.12} $\pm$ .02 & \heatmapEAURC{.14}{.06}{.15} $\pm$ .04 & 8.3 \\
ATA .3\, & \heatmapEAURC{.04}{.04}{.13} $\pm$ .01 & \heatmapEAURC{.04}{.02}{.04} $\pm$ .00 & \heatmapEAURC{.10}{.04}{.12} $\pm$ .01 & \heatmapEAURC{.12}{.06}{.15} $\pm$ .04 & 8.5 \\
PLM 50\, & \heatmapEAURC{.08}{.04}{.13} $\pm$ .01 & \heatmapEAURC{.02}{.02}{.04} $\pm$ .00 & \heatmapEAURC{.06}{.04}{.12} $\pm$ .01 & \heatmapEAURC{.12}{.06}{.15} $\pm$ .04 & 8.8 \\
AQA .9\, & \heatmapEAURC{.06}{.04}{.13} $\pm$ .01 & \heatmapEAURC{.03}{.02}{.04} $\pm$ .00 & \heatmapEAURC{.06}{.04}{.12} $\pm$ .01 & \heatmapEAURC{.12}{.06}{.15} $\pm$ .04 & 8.8 \\
ATA .7\, & \heatmapEAURC{.13}{.04}{.13} $\pm$ .01 & \heatmapEAURC{.02}{.02}{.04} $\pm$ .00 & \heatmapEAURC{.12}{.04}{.12} $\pm$ .01 & \heatmapEAURC{.15}{.06}{.15} $\pm$ .04 & 9.8 \\
ICA\, & \heatmapEAURC{.12}{.04}{.13} $\pm$ .01 & \heatmapEAURC{.04}{.02}{.04} $\pm$ .01 & \heatmapEAURC{.05}{.04}{.12} $\pm$ .01 & \heatmapEAURC{.11}{.06}{.15} $\pm$ .03 & 11.2 \\
AQA .6\, & \heatmapEAURC{.09}{.04}{.13} $\pm$ .01 & \heatmapEAURC{.03}{.02}{.04} $\pm$ .00 & \heatmapEAURC{.06}{.04}{.12} $\pm$ .01 & \heatmapEAURC{.12}{.06}{.15} $\pm$ .04 & 11.7 \\
AQA .75\, & \heatmapEAURC{.08}{.04}{.13} $\pm$ .01 & \heatmapEAURC{.03}{.02}{.04} $\pm$ .00 & \heatmapEAURC{.06}{.04}{.12} $\pm$ .01 & \heatmapEAURC{.12}{.06}{.15} $\pm$ .04 & 11.7 \\
AVG\, & \heatmapEAURC{.13}{.04}{.13} $\pm$ .01 & \heatmapEAURC{.03}{.02}{.04} $\pm$ .00 & \heatmapEAURC{.06}{.04}{.12} $\pm$ .01 & \heatmapEAURC{.12}{.06}{.15} $\pm$ .04 & 11.8 \\
\addlinespace[23pt]
\multicolumn{1}{c}{} & \multicolumn{1}{c}{\smash{\rotatebox[origin=lt]{55}{ARC-BC}}} & \multicolumn{1}{c}{\smash{\rotatebox[origin=lt]{55}{ARC-Nuc}}} & \multicolumn{1}{c}{\smash{\rotatebox[origin=lt]{55}{LIDC-Mal}}} & \multicolumn{1}{c}{\smash{\rotatebox[origin=lt]{55}{LIDC-Tex}}} & \\
\end{tabular}%
}
    \caption{\textbf{\Aggs{} performance on TTA heatmaps \Aggs{} in \ood{} and failure detection.} Color coding and metric interpretation are as detailed in Table A.1. (a)-(b) \Aggs{} are ranked from best (top) to worst (bottom) based on their average metric, first computed across 500 bootstrap samples per dataset, and then averaged across datasets to ensure stable rankings. Each value is reported with the std. deviation across bootstrap samples.}
    \label{tab:tta_bootstrapped_results}
\end{table}

\section{Details on Experimental Setup} 
\label{app:data}

\begin{figure*}[h]
    \centering
    \includegraphics[width=1\textwidth]{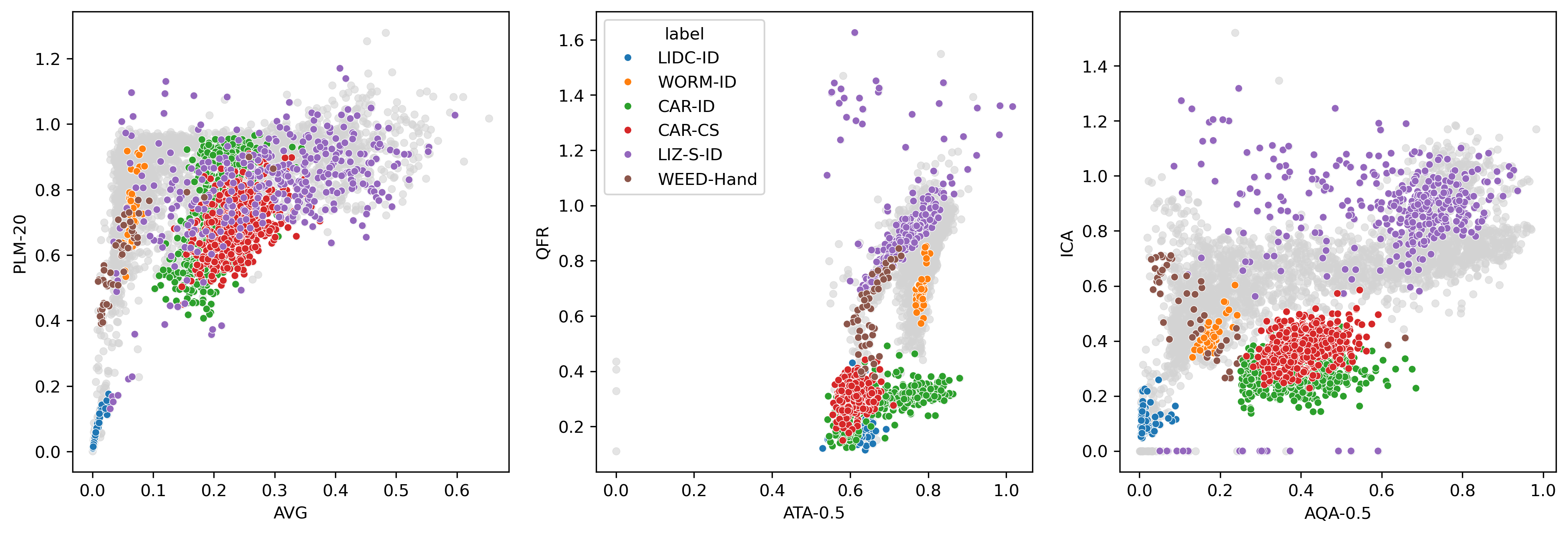}
    \caption{\textbf{Illustration of dataset diversity as captured by different aggregation measures.} Unlike the projection of uncertainty maps into EDS–MOR space (cf. Fig.~\ref{fig:datasets}, with non-highlighted datasets shown in gray), projections onto spaces defined by randomly chosen pairs of \textit{pixel-intensities} or \textit{prediction-aware} \Aggs{} do not reveal a clear separation of the datasets. Nevertheless, the two-dimensional configuration of CAR-ID and CAR-CS points suggests that a separation might emerge in a multidimensional space incorporating all \Aggs{}.}
    \label{fig:diversity_appendix}
\end{figure*}

The datasets were selected to cover a broad range of application domains for image segmentation (autonomous driving, agricultural monitoring, biomedical imaging) as well as maximum diversity in terms of image structure (\eg number, size, and shape of objects, as well as the number of semantic classes). This diversity is reflected in the distribution of aggregated uncertainty scores, as illustrated in ~\Cref{fig:datasets}.~\Cref{fig:diversity_appendix} further complements the illustration of the datasets diversity by showing the distribution of aggregated scores along additional pairs of \Aggs{}.

\subsection{Segmentation of nuclei in pathology images}
\paragraph{LIZ (\iid{})} The Lizard dataset~\cite{graham2021lizard} is a large-scale histopathology (H\&E) dataset with annotations for instance- and semantic segmentation of cell nuclei in histopathology images of colon tissue. The dataset consists of 238 images of varying sizes, from which we extract patches of size 256 $\times$ 256 pixels. Patches from the DigestPath, TCGA, PanNuke, CRAG and CoNSeP subsets are used as \iid{}.

\paragraph{LIZ-G (\ood{})} Based on established baselines ~\cite{rumberger2022panoptic,baumann2024hover}, we use the GlaS subset as the \ood{} dataset, consisting of 61 images, where we again extract patches of size 256 $\times$ 256 pixels. We evaluate both semantic and instance segmentation, referring to the resulting \ood{} sets as LIZ-SG (semantic) and LIZ-IG (instance). Here, the distribution shift arises from variations in acquisition and recording conditions (\eg, lighting, temperature, and focal plane), which affect tissue appearance and increase overall object uncertainty.

\paragraph{Segmentation} We train the HoVer-NeXt (HN) model~\cite{baumann2024hover}. HN builds upon HoVer-Net (HRNet, \cite{wang2020deep}) by simplifying the pipeline: it replaces the binary nuclei segmentation map with a 3-class center-background (BCB) prediction and merges the instance segmentation decoders into a single branch. The semantic arm predicts class labels for each pixel, while the instance arm outputs center-point vectors and BCB maps for individual nuclei. The architecture follows a U-Net~\cite{ronneberger2015u} with a ConvNeXt-v2 encoder \cite{woo2023convnextv2} (Tiny variant). \newline \noindent 
HN is trained from scratch for 200,000 steps with a batch size of 12 using AdamW (weight decay 0.0001) and a cosine-annealing learning rate schedule (1e-4 to 1e-8). The encoders use 50\% dropout; the decoder does not. Data augmentations include: HED color, hue/saturation/brightness, random noise, Gaussian blur, rotation, flipping, mirroring, zoom, scale, shear, translation, and elastic transforms. The training loss combines semantic and instance losses, summed and weighted by 0.02. The instance arm uses MSELoss for center-point vectors and cross-entropy for the BCB map, while the semantic arm employs a standard focal loss~\cite{lin2017focal} with $\gamma = 2.0$. Hyperparameter search and performance monitoring are conducted on the validation set, optimizing micro- and macro-F1 scores rather than panoptic quality, which has been shown to be suboptimal for histopathology data~\cite{foucart2023panoptic}.

\subsection{Synthetic Histopathological Images}
\paragraph{ARC (\iid{})} The Arctique dataset~\cite{franzen2024arctique} is a procedurally generated dataset modeled after histopathological colon tissue sections (inspired by the LIZ dataset), providing precise ground-truth masks for both semantic and instance segmentation. We use the standard training set which comprises 1,500 samples of $512 \times 512$-pixel RGB images.

\paragraph{ARC-Nuc/ARC-BC (\ood{})} Arctique includes controlled variations in targeted parameters, such as the presence of red blood cells (ARC-BC) and the intensity of nuclei staining (ARC-Nuc). In the ARC-BC setting, additional red blood cells are misclassified as eosinophil cells, leading to local increases in uncertainty at the locations of the blood cells for semantic segmentation. In the ARC-Nuc setting, treduced nuclei staining intensity makes it more difficult for the model to identify individual cells, thus increasing the uncertainty at object boundaries in instance segmentation.

\paragraph{Segmentation} We use the same training setup as described for the LIZ dataset.

\subsection{Binary segmentation of lung nodules in CT volumes}
\paragraph{LIDC (\iid{})} We use the LIDC-IDRI dataset \cite{armato2011lung}, with the \texttt{pylidc} library \cite{hancock2016lung}, following the pre-processing strategy of \citet{kahl2024values}. in \citet{kahl2024values}. In particular, we use only lung nodules $\geq$ 3mm; each annotated by up to four raters, and a consensus mask is computed as the union of all annotations. Images are cropped to 64×64×64 voxels and resampled to 1×1×1 mm resolution, yielding 901 samples. Finally, for better comparability with the other datasets, we convert the images and masks volumes to 2d by restricting evaluation to the central 50\% of the z-axis.

\paragraph{LIDC-Tex/LIDC-Mal (\ood{})} The LIDC-IDRI dataset includes nine metadata features, each with 4–6 categorical values assigned by the raters, which we use to define the \iid{} and \ood{} splits. Specifically, we consider textured (\iid{}) nodules vs. non-textured (\ood{}) nodules for the texture shift (LIDC-Tex) and benign (\iid{}) nodules vs. malignant (\ood{}) nodules for the malignancy shift (LIDC-Mal). Since malignant nodules are typically larger in size than benign nodules, the uncertainty in the LIDC-Mal shift is expected to increase along the object boundaries. In contrast, in the LIDC-Tex shift uncertainty will increase within the nodule, reflecting the change in texture.

\paragraph{Segmentation} We use a 3D U-Net as the segmentation backbone to detect lung nodules, with an initial filter size of 16 and four encoder–decoder blocks. The loss combines Dice and cross-entropy terms. Data augmentations include random flipping and Gaussian noise. Training uses the Adam optimizer (learning rate 3e-4 and weight decay 1e-5) with a batch size of 8 and 50\% dropout after each convolutional block. During training, validation performance is measured as the average Dice score between the predictions (or a single prediction for MSP) and each of the four reference annotations. Further details are provided in the Supp. of \citet{kahl2024values}.

\subsection{Binary segmentation of microorganisms in microscopy images}

\paragraph{WORM (\iid{})} The C.elegans live/dead assay from Broad Bioimage Benchmark Collection (BBBC010) contains 100 brightfield microscopy images of multiple C. elegans worms with pixelwise foreground and instance segmentation masks~\cite{ljosa2012annotated}. Images are single-channel and and cropped from  $696\times520$ pixels to $512\times512$ pixels to match the network input constraints. 

\paragraph{WORM-Nem/WORM-Pro (\ood{})} We use the Protist (1003 samples) and Nematode (468 samples) datasets from the SinfNet Microorganism Image Classifier dataset \cite{Sabban_SinfNet_Microorganism_image_2023}. Both contain microscope images of various microorganisms along with polygon annotations of the instances. For comparability with the \iid{} data, images are converted to grayscale, polygon annotations are converted to pixelwise masks, and both images and masks are cropped/resized to $512\times512$ pixels. In this binary segmentation setting, the main sources of uncertainty are the unfamiliar shapes of the other microorganisms and the background noise (\eg small particles misclassified as foreground). 

\paragraph{Segmentation} We train a standard U-Net with 5 convolutional layers, ReLU activations and 20\% dropout set after each convolutional layer. Input images are augmented with random resized crops, flips, Gaussian blur, sharpness, gamma adjustment, and intensity inversion. The network predicts binary foreground/background, trained with a combined Dice and cross-entropy loss. Optimization is performed with Adam (learning rate 1e-5, weight decay 1e-5), in batches of 16 for 20 epochs. Validation performance is monitored using the Dice score.

\subsection{Semantic segmentation of urban street scenes}
\paragraph{CAR-ID (\iid{})}
The GTA dataset~\cite{Richter_2016_ECCV} contains 24,966 ($1914 \times 1052$) of urban street scenes with dense semantic segmentation masks covering 19 classes from the computer game GTA V. Semantic masks are constructed by intercepting information between the game engine and graphics processor, and by reconstructing and annotating individual objects in scenes. These masks distinguish between 19 semantic classes.
To address ethical concerns, we verified that the dataset contains only urban street scenes with the 19 semantic classes matching standard driving benchmarks; no violent or sexualized content is present, and pedestrians appear incidentally while walking on sidewalks (0.36\% of pixels). We thank the anonymous reviewer for bringing this issue to our attention.

\paragraph{CAR-CS (\ood{}) }
The Cityscapes dataset~\cite{Cordts2016Cityscapes} includes 5,000 densely annotated urban street scenes from 27 cities and 20,000 coarsely annotated images from 23 cities. Images were captured via stereo cameras, thus the resolution being $1280 \times 720$, and annotations include instance and semantic segmentation masks (differientating 30 classes), which we map to the 19 GTA classes. The \ood{}-ness stems from a sim-to-real shift, affecting the entire image and resulting in more “blurry” uncertainty maps.

\paragraph{Segmentation} We perform semantic segmentation using HRNet~\cite{wang2020deep} as the backbone, pre-trained on ImageNet. The model is trained with cross-entropy loss and optimized using SGD (learning rate 0.01, weight decay 5e-4, momentum 0.9) with a batch size of 6. Data augmentations include random horizontal flipping, rotations, scaling, cropping, and Gaussian noise. 50\% dropout is applied at the end of each branch. Validation performance is monitored using the Dice score. Further training details are provided in the Supplement of \citet{kahl2024values}.


\subsection{Crop and Weed}

\paragraph{WEED (\iid{})} The Weedsgalore datase is a high-resolution drone-based multispectral imaging dataset with dense annotations for crop and weed segmentation in maize fields \cite{Celikkan_2025_WACV}. We use the training set comprising 104 images, each measuring $600\times600$ pixels, and only the RGB channels for prediction. 

\paragraph{WEED-Hand (\ood{})} The "Crop and Weed" dataset \cite{steininger2023cropandweed} contains 8,034 manually captured RGB images covering 16 species of crop and 58 species of weed over a variety of locations, soil types and lightning conditions. For comparability with Weedsgalore we collapse annotations into two classes, by considering maize as "crop" and all other plants as "weed". We further crop images to a a quadratic size of $600\times600$ pixels. 

\paragraph{Segmentation} We use publicly available checkpoints of a probabilistic DeepLabv3+ model \cite{chen2018encoder} trained on Weedsgalore for 3-class semantic segmentation (background, crops, weeds).



\subsection{Training and Hardware}\label{app:train}
When training was performed from scratch, it was carried out on single NVIDIA H100 and A40 GPUs using PyTorch~\cite{paszke2019pytorch}.

\subsection{Evaluation sets}\label{app:evaluation-sets}

\begin{table}[h!]
\centering
\begin{tabular}{lcc}
\toprule
\textbf{Dataset} & \textbf{\iid{} Samples} & \textbf{\ood{} Samples} \\
\midrule
ARC-BC        & 25  & 50  \\
ARC-Nuc        & 25  & 50  \\
LIDC-Mal          & 53  & 93  \\
LIDC-Tex             & 84  & 20  \\
WORM-Nem           & 25  & 47  \\
WORM-Pro            & 25  & 82  \\
CAR-CS          & 300 & 300 \\
WEED-HAND             & 26  & 159 \\
LIZ-SG     & 193 & 356 \\
LIZ-IG    & 193 & 356 \\
\bottomrule
\end{tabular}
\caption{Sample counts for \iid{} and \ood{} splits across evaluation datasets.}
\label{tab:samples}
\end{table}

\noindent As evident from the dataset descriptions in this Supp. Section, dataset sizes vary widely, from roughly 100 training samples in WORM to tens of thousands in CAR. For comparability between \iid{} and \ood{} sets, as well as for processing efficiency, we use randomly selected subsets of the larger datasets for evaluation experiments. Final sample counts are summarized in Table \ref{tab:samples}.



\section{Details on Downstream Tasks}
\label{app:supplement_downstream_tasks}

\subsection{Out-of-Distribution Detection} \label{app:ood}
To empirically evaluate the theoretical properties of \Agg{}s, we focus on Out-of-Distribution (\ood{}) detection at the image level rather than the pixel level. Intuitively, humans perceive an image as \ood{} as a whole rather than based on individual pixels. Moreover, when part of an image is OoD, it can affect all predictions, making them unreliable.

\paragraph{Metric: Area Under the Receiver Operating Characteristic Curve (AUROC)}
We use AUROC to assess an \Agg{}'s ability to detect \ood{} images. Each image is labeled as 1 for \ood{} and 0 for \iid{}. The True Positive Rate (TPR) is the proportion of correctly identified \ood{} images, i.e., the fraction of \ood{} samples whose aggregated uncertainty score exceeds a given threshold. The False Positive Rate (FPR) reflects the proportion of \iid{} images incorrectly classified as \ood{}, i.e., the fraction of \iid{} samples whose aggregated uncertainty score surpasses the same threshold. By varying this threshold, we calculate the TPR and FPR at different levels to construct the ROC curve and subsequently compute the AUROC. We use the \texttt{sklearn} library \cite{scikit-learn} to first compute the TPRs and FPRs of the ROC curve
with the ground truth input ($0$ or $1$) and the aggregated uncertainty scores as target values. From this we then calculate the AUC (area under the curve).

\subsection{Failure Detection} \label{app:sel-cl}
To assess the impact of \Agg{}s on model performance in real-world applications, we define a continuous failure signal based on two segmentation accuracy metrics: (1) micro-Dice, which captures the accuracy of object and boundary detection in the instance-3-label segmentation task (\eg nuclei and boundary detection), and (2) macro-Dice, which reflects class-wise performance in semantic segmentation tasks. The motivation behind this approach is that, while automated decision-making requires a holistic view at image level, the performance of a panoptic, instance-based three-label, or semantic segmentation model must ultimately be evaluated at the instance level.

\paragraph{Metric: Excess-Area Under the Risk-Coverage Curve (E-AURC)} 
As analyzed by \citet{jaeger2023reflectevaluationpracticesfailure} and originally proposed by \citet{geifman2018biasreduced}, we use the Excess-Area Under the Risk-Coverage Curve (E-AURC) as the evaluation metric for the SC experiment. In this downstream task, E-AURC measures the quality of \Aggs{}-based ranking while remaining independent of the underlying model’s absolute performance. Moreover, we compute E-AURC for each employed uncertainty-based predictive model, making the segmentation model’s absolute performance irrelevant and further confirming the suitability of E-AURC for our benchmarking. Like AURC \cite{geifman2017selective}, E-AURC balances two objectives:
minimizing risk (i.e., ensuring strong classifier performance) while maximizing coverage (i.e., reducing the fraction of cases requiring manual review). Both AURC and E-AURC are computed following the implementation of \citet{jaeger2023reflectevaluationpracticesfailure}, adapted here for multi-class segmentation as detailed below. \newline \noindent 
Let $\phi$  denote a predictive model that maps an image to a multi-class segmentation mask. We define the evaluation dataset of images $I$ and segmentation masks $M$ as  $\widetilde{\mathscr{D}} = \{I_\ell, M_\ell\}^{L}_{\ell=1}$, and the \textit{confidence scoring function} (CSF) $g(I_\ell)$ as the negative aggregated uncertainty score, $- f (U_\ell)$. Furthermore, we choose the inverted micro-Dice (macro-Dice) as the \textit{risk} $s$ associated with a prediction,
\begin{equation}
    s(I,M,\phi) = 1 - \operatorname{Dice}(\phi(I), M).
\end{equation}
The risk-coverage curve is obtained by introducing a confidence threshold $\rho$, which leads to the selective risk
\begin{equation}
    \operatorname{Sel. Risk}(\rho \vert \phi,g,\widetilde{\mathscr{D}}) = \frac{\sum_{\ell} s(I_\ell,M_\ell, \phi) \cdot \mathbb{I}(g(I_\ell) \geq \rho)}{\sum_{\ell} \mathbb{I}(g(I_\ell) \geq \rho)}
\end{equation}
and coverage, defined as the ratio of cases remaining after selection,
\begin{equation}
    \operatorname{Coverage}(\rho \vert g, \widetilde{\mathscr{D}}) = \frac{\sum_{\ell} \mathbb{I}(g(I_\ell) \geq \rho)}{L}.
\end{equation}
The AURC based on a threshold list $\{\rho_r\}^R_{r=1}$ with $R$ values of a CSF that are sorted ascending can then be computed as,
\begin{align}
    \operatorname{AURC}(f,g,\widetilde{\mathscr{D}}) &= \sum_{r} (\operatorname{Coverage}(\rho_{r}) - \operatorname{Coverage}(\rho_{r-1}) \cdot \\
    &\cdot \frac{(\operatorname{Sel. Risk}(\rho_{r}) + \operatorname{Sel. Risk}(\rho_{r-1}))}{2} \notag
\end{align}
where we omit the conditioning on $\phi$, $g$, $\widetilde{\mathscr{D}}$ on the RHS for brevity.
We now derive the E-AURC as follows,
%
\begin{align}
    \operatorname{E-AURC} &= \operatorname{AURC}(\phi,g,\widetilde{\mathscr{D}}) - \operatorname{AURC}(\phi,g^*,\widetilde{\mathscr{D}}) 
\end{align}
where the second term represents the optimal AURC achievable. This optimal CSF can be defined, for example, by an oracle that assigns to each prediction a confidence equal to the negative risk: $g^{*}(x) = -s(x, y, \phi)$. In practice, this corresponds to perfectly ranking the predictions by their risk, in our case, in ascending order of Dice score.
For a formal discussion of the applicability and limitations of this metric, please refer to \citet{jaeger2023reflectevaluationpracticesfailure}.

\section{Details on UQ map generation}
\label{app:supp_uq_maps}


For UQ, we use Monte Carlo Dropout for all results shown in the main text. All segmentation models employ dropout during training, and the dropout layers remain active at test time. By passing each input image through the model for $L=10$ times with different dropout masks 
%
we generate $L$ samples of the pixel-wise class probabilities $p_{i}^{(l)}(c)$, where $l = 1, \dots, L$ indexes the dropout samples and $i=1,\dots,mn$ refers to the pixel index. We then compute the mean class probability across samples as:

\begin{equation} \bar{p}_{i}(c) = \frac{1}{L} \sum_{l=1}^L p_{i}^{(l)}(c). \end{equation}

\noindent Finally, pixelwise uncertainty is calculated using Shannon entropy over the averaged probabilities:

\begin{equation} u_{i} = -\sum_{c=1}^K \bar{p}_{i}(c) \log \bar{p}_{i}(c). \end{equation}

\noindent Shannon Entropy is maximal when all $K$ classes have equal probability i.e $p(c) = 1/K$ for $c=1\dots K$. In that case $-\sum_{c=1}^Kp(c)\log(p(c)) = \log(K)$ and thus the pixelwise uncertainty scores $u_{i}$ fall in $[0, \log(K)]$ where $K$ is the total number of (semantic) classes. \newline \noindent
The same generation process applies to the pixel-wise class probabilities produced by other UQ methods when replicating the \Aggs{} benchmarking for \ood{} and FD detection. What differs is how and how many samples $p_i^{(l)}(c)$ are generated:
\begin{itemize}
    \item TTA: $L=16$ samples are generated by applying various combinations of the augmentations used during training to the test input, then inverting them before applying Softmax.
    \item DE: $L=5$ samples are obtained from predictions of five segmentation models with identical architecture and training setup, but initialized with different random seeds.
    \item CE: $L=10$ samples are generated using nine model checkpoints selected near convergence in the loss basin.
\end{itemize}
we generate $L=10$ by using 9 checkpoints of the model registered near convergence to loss basin. \newline \noindent
To ensure comparability across aggregated uncertainty maps, we divide the pixelwise scores from all UQ methods except MSP by $\log(K)$ to make sure all maps are normalized such that $U \in \left[0,1\right]^{m \times n}$.





\section{Details on Meta-Aggregation via GMM}
\label{sec:supplement_gmm}

\begin{figure}[t]
    \centering
    \includegraphics[width=\linewidth]{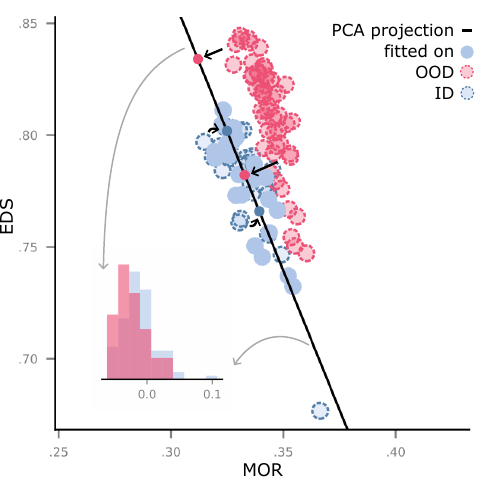}
    \caption{\textbf{Effect of PCA.} Example of feature collapse resulting from computing a 1D PCA projection in 2D space using only \iid{} samples, and subsequently applying this projection to both \iid{} and \ood{} samples. The resulting histogram (inset) illustrates that this approach yields poor separability between \iid{} and \ood{} data.}  
    \label{fig:pca}
\end{figure}

\subsection{Gaussian Mixture Models}

The meta-aggregation approach is motivated by the task of \ood{} detection as illustrated in 
Fig.~\ref{fig:aggregation_strategies}. Here, we assume access to a representative collection of uncertainty maps generated from \iid{} data, i.e. data drawn from the same distribution the model was trained on. To summarize the overall uncertainty present in these maps, we can apply one of the various aggregation strategies (\Aggs{}) introduced in Sec. \ref{sec:common_agg}, \ref{sec:spatial_agg} and \ref{sec:meta_agg}. \newline \noindent
When deploying the same trained model on a new, potentially out-of-distribution sample our experimental results in Fig.~\ref{fig:ood_results} and Fig.~\ref{fig:fd_results} reveal a key challenge: it remains unclear which specific \Agg{} is most suitable for determining whether a given sample is \iid{} or \ood{} for any particular task. \newline \noindent
Therefore, instead of relying exclusively on any single \Agg{} we aim to combine multiple \Aggs{} each focusing on different aspects of the uncertainty map.
To this end, we represent each uncertainty map $U$ by its aggregated feature vector $\mathbf{f}_{U} \in\mathbb{R}^d$, defined as $\mathbf{f}_U = (f_1(U), \dots, f_d(U))$ where $d$ denotes the number of aggregation strategies. We then model the distribution of these features using a Gaussian Mixture Mode Gaussian Mixture Model (GMM) \cite{dempster1977maximum}:
\begin{equation}
p_{\text{GMM}}\!\bigl(\mathbf{f}_U;\Theta\bigr)
=\sum_{k=1}^{K}
      \pi_k\,
      \mathcal{N}\!\bigl(\mathbf{f}_U\mid
            \boldsymbol{\mu}_k,\boldsymbol{\Sigma}_k\bigr). 
\end{equation}

\noindent Here, $\Theta=\{\pi_k,\boldsymbol{\mu}_k,\boldsymbol{\Sigma}_k\}_{k=1}^{K}$ are the parameters of the mixture, with $\pi_k$, $\boldsymbol{\mu}_k$, and $\boldsymbol{\Sigma}_k$ representing the mixing coefficient, mean vector, and covariance matrix of the $k$-th component, respectively. The number of components $K$ is selected via the Bayesian Information Criterion (BIC) \cite{schwarz1978estimating}. \newline \noindent
Each Gaussian component is defined by
\[
\mathcal{N}\!\bigl(\mathbf{f}_U\mid\boldsymbol{\mu},\boldsymbol{\Sigma}\bigr)
  \;=\;
  \frac{1}{(2\pi)^{d/2}\,|\boldsymbol{\Sigma}|^{1/2}}
  e ^{
        -\frac{1}{2}
        (\mathbf{f}_U-\boldsymbol{\mu})^{\!\top}
        \boldsymbol{\Sigma}^{-1}
        (\mathbf{f}_U-\boldsymbol{\mu})
       },
\]
And the mixture weights $\pi_k$ are constrained by: 
\begin{equation}
\sum_{k=1}^{K}\pi_k=1,\;
\pi_k\ge 0.
\label{eq:gmm_def}
\end{equation}
We estimate the mode-parameters $\Theta$ on a held-out \iid{} subset and evaluate on unseen \iid{} and \ood{} data points. \newline \noindent 
The final meta-aggregator score is defined as the negative log-likelihood (NLL): 
\begin{equation}
f_{\text{meta}}(U) \;=\; -\log p_{\text{GMM}}\!\bigl(\mathbf{f}_U\bigr),
\label{eq:meta_score}
\end{equation}
which serves as a unified uncertainty score: larger values indicate greater deviation from the learned \iid{} distribution and thus flag the corresponding sample as likely \ood{}. \newline \noindent
An example of the effectiveness of this approach is illustrated in Fig.~\ref{fig:ood_qual}d-e, where EDS and MOR individually are unable to separate between \iid{} and \ood{} data. Yet, fitting a GMM on both allows for an improved separation. 

\begin{figure*}[t]
    \centering
    \includegraphics[width=\linewidth]{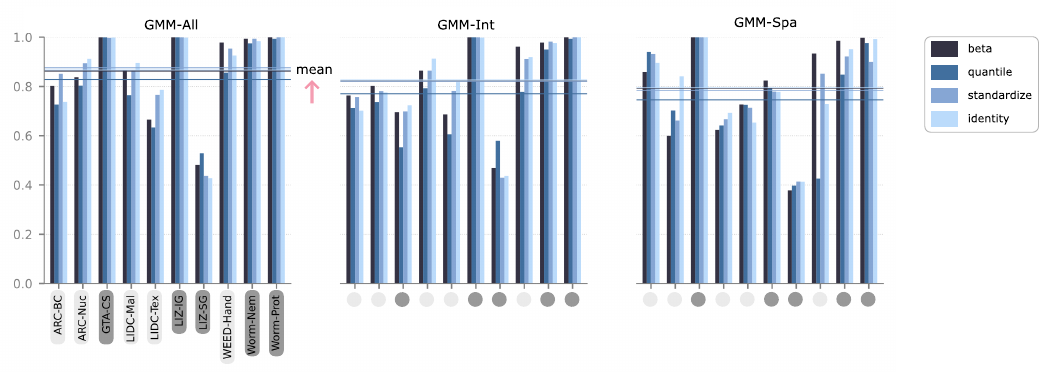}
    \caption{\textbf{Data Normalization.} AUROC scores from our OOD detection experiment are shown for four different normalization methods: Beta CDF Gaussianizer, Standard Scaler, Quantile Transformer, and no normalization. The three bar plots represent the performance of the methods \textit{GMM-All}, \textit{GMM-Int}, and \textit{GMM-Spa} under each normalization setting.}
    \label{fig:preprocessing}
\end{figure*}

\subsection{Data preprocessing}\label{app:datapreproc}
We compare three different aggregation sets to fit a GMM on: (1) \emph{GMM-Int}, based on 13 classical intensity-based \Aggs{}; \emph{GMM-Spa}, which includes our 3 novel spatial \Aggs{}; and (3) \emph{GMM-All}, which combines all 16 aggregated scores.  \newline \noindent
The relatively high dimensionality of the aggregated uncertainty feature vectors would typically motivate dimensionality reduction prior to fitting the GMM. However, for the task of \ood{} detection, such an approach can actually be detrimental. As illustrated in Fig.~\ref{fig:pca} for the Principled-Component Analysis (PCA) technique: projecting the \ood{} samples onto the two lower-dimensional PCA subspace leads to a strong overlap between \iid{} and \ood{} samples, making their separation more challenging. \newline \noindent
A second preprocessing issue concerns the bounded range of the individual \Aggs{}. These scores lie in $[0,1]$, whereas GMMs are defined over $[-\infty,\infty]$. However, we find that non-linear 
transformations can lead to distortions, by exaggerating differences between data points that are close to 0 or 1, potentially affecting the GMM's ability to accurately detect \ood{} samples. Hence, following common practice, we standardize our features $\mathbf{f}_U$ by $\mathbf{f}'_U = (\mathbf{f}_U- \mu)/\sigma$, where $\mu$ and $\sigma$ denote the mean and standard deviation, respectively. This preprocessing is used for the results shown in Figure ~\ref{fig:ood_results} and \ref{fig:fd_results}. \newline \noindent
To evaluate the impact of the normalization methods on the GMM performance, in Fig.~\ref{fig:preprocessing} we test four different strategies (including no normalization) for the \ood{} detection benchmark (note that prior to normalization, we rescale scores as $\mathbf{f}_U = (1 - 2  \epsilon)(\mathbf{f}_U - 0.5) + 0.5$ with a small $\epsilon$ to avoid values at 0 or 1). Two key takeaways emerge regardless of the normalization method: (1) the mean AUROC consistently increases when using all the features (\emph{GMM-All}); and (2) the meta-aggregators outperform in most cases (7 out of 10) the individual \Aggs{} (cf. ~\Cref{fig:ood_results}). Furthermore, while performance varies across individual datasets, the mean AUROC indicates that the Quantile Transformer is the least effective among the tested preprocessing methods.

\subsection{Meta-aggregation ablations} \label{app:gmm-ablations}
\begin{figure}[t]
    \centering
    \includegraphics[width=\columnwidth]{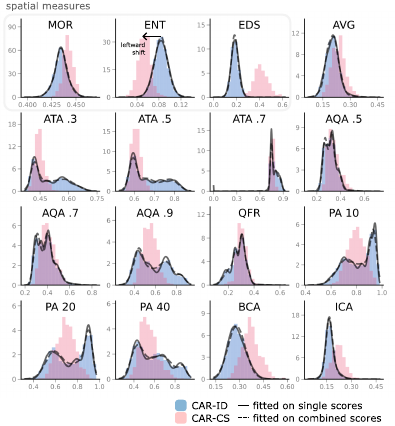}
    \caption{\textbf{Utilizing spatial measures as \Aggs{}.} Histograms of individual \Aggs{} for CAR-ID and CAR-CS are shown along with: 1. a GMM fitted on an an \iid{} subset of the individual score (solid line), and 2. the diagonal component of a GMM fitted on all scores (dotted line). This empirically shows that spatial measures can be leveraged by a GMM to detect \ood{} samples, even when the shift from \iid{} to \ood{} is toward lower aggregated values.}
    \label{fig:fitting_single_vs_combined}
\end{figure}

\begin{figure*}[t]
    \centering
    \includegraphics[width=\linewidth]{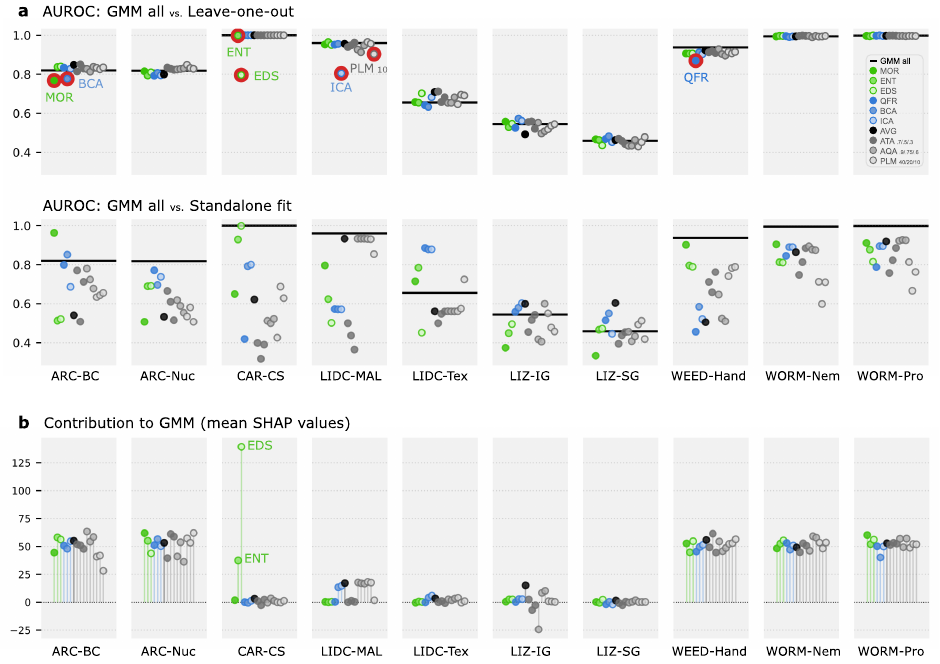}
    \caption{\textbf{GMM Robustness for \ood{} Detection for all datasets.} (a) \textbf{\emph{GMM-All} AUROC: leave-one-out (top) and individual fitting (bottom).} Using all \Aggs{} generally matches or outperforms individual ones in 6 of 10 cases, with minimal impact from removing specific \Aggs{}. Exceptions occur when a feature dominates (\eg, EDS for CAR-CS or ICA in LIDC-MAL, marked in red as Tukey outliers) or when features lack discriminative power between \iid{} and \ood{} samples (\eg, LIZ-SG). The most discriminative \Aggs{} for ARC-BC, ARC-Nuc, CAR-CS, LIDC-MAL, and WEED-Hand are typically \textit{spatial mass ratios} or \textit{prediction-based} (proportion-invariant). (b) \textbf{\emph{GMM-All} AUROC: absolute SHAP values.} Positive values indicate that an \Agg{} helps \emph{GMM-All} separate \iid{}–\ood{}, confirming the results observed in (a); bi-directional or null contributions reduce performance (e.g., in LIZ-IG).}
    \label{fig:gmm_stab_rebuttal}
\end{figure*}

\Cref{fig:gmm_stab_rebuttal} provides a comprehensive overview of the ablation studies conducted for the meta-aggregation \emph{GMM-All} in \ood{} detection. We focus on this downstream task, as its evaluation metric is more intuitively interpretable than that of FD. 
The top panel of \Cref{fig:gmm_stab_rebuttal}a compares the AUROC of \emph{GMM-All} (shown as a horizontal black line) to leave-one-out variants, in which one \Agg{} is omitted at a time during fitting. Results are visualized as jittered points obtained by evaluating across test samples spanning the 10 benchmarking datasets and following the bootstrapping protocol described in Supp.~\ref{app:datapreproc}. 
The \iid{} and \ood{} samples used to compute the AUROC are the ones reported in Supp.~\ref{app:evaluation-sets} and obtained via a fixed split; the remaining \iid{} samples are those used for fitting the GMM.
The bottom panel of \Cref{fig:gmm_stab_rebuttal}a, in turn, shows the performance of GMMs fitted on each individual \Agg{} relative to the combined approach. \newline \noindent
The two panels indicate that fitting \emph{GMM-All} on all \Aggs{} either outperforms or matches the performance of individual-feature fits in 6 of 10 datasets, with generally minimal sensitivity to the removal of specific features. Exceptions arise when: a particular feature dominates (\eg, EDS for CAR-CS and QFR for WEED-Hand, marked as Tukey outliers \cite{tukey1977exploratory}), or when most features lack clear discriminatory power between \iid{} and \ood{} samples (\eg, in LIZ-SG). Notably, the most discriminative \Aggs{} for ARC-BC, ARC-Nuc, CAR-CS, LIDC-MAL, and WEED-Hand are either \textit{spatial mass ratios} or \textit{prediction-based} (proportion-invariant), consistent with the findings reported in Section~\ref{sec:overall-res}. \newline \noindent
\Cref{fig:fitting_single_vs_combined} provides distributional insights into the individual fitting for the CAR-ID and CAR-CS datasets by illustrating the distinct \iid{} and \ood{} distributions of the \Aggs{}. It further depicts the \emph{GMM-All} density when fitted on each individual \Agg{} (solid line), alongside its corresponding diagonal component (dotted line) when fitted on all \Aggs{}. While the \ood{} shift induced by real CAR-CS images relative to synthetic CAR-ID images is clearly reflected in the EDS, this is not the case for MOR, where the support and likelihood of the \iid{} and \ood{} distributions do not exhibit a clear separation, nor for ATA .3, whose distribution is broader and bi-modal, making an individual GMM fit insufficient. In contrast, ENT highlights a key advantage of the meta-aggregator: using a density-based \ood{} detector allows us to identify \ood{} samples, even when their aggregate uncertainty for that score is lower than that of \iid{} samples. Overall, this illustrates that spatially aware scores can be incorporated as \Aggs{} in a GMM-based \ood{} detector, potentially allowing us to capture distributional structure beyond simple shifts in aggregate uncertainty. \newline \noindent
To complement our ablations and provide a mechanistic explanation for the results in \Cref{fig:ood_results}, we first fit the \emph{GMM-All} on all aggregated values using the protocol and sample splits described above. We then compute NLL scores for \iid{} and \ood{} test samples, followed by SHAP values \cite{NIPS2017_7062} to assess how individual features contribute to the \emph{GMM-All} predictions (cf. Supp.~\ref{app:evaluation-sets}). These SHAP values quantify the contribution of each \Agg{} to the \emph{GMM-All}’s performance in detecting \ood{} samples. \newline \noindent
\Cref{fig:gmm_stab_rebuttal}b reports these SHAP values averaged across samples from the 10 benchmarking datasets (absolute SHAP), highlighting that, for datasets where \emph{GMM-All} achieved a particularly high AUROC, all individual \Agg{} components contributed positively, producing a clear separation between \ood{} samples and the estimated GMM modes. The positive contributions are especially pronounced for datasets exhibiting semantic shifts (WEED-Hand, WORM-Nem, WORM-Pro). By contrast, for synthetic datasets with covariate shifts (ARC-BC and ARC-Nuc), uncertainty concentrated at cell borders reduces the effectiveness of intensity-based \Aggs{} such as PLM and AQA. These methods rely on hyperparameters tuned to localized uncertainty distributions in the simulated tissue slices, slightly limiting overall \emph{GMM-All} performance. \newline \noindent
The performance of \emph{GMM-All} on CAR-CS is entirely explained by EDS, which alone suffices as a discriminator, followed by ENT. For lung-nodule detection, even if prediction-unaware \Aggs{} are favored by the larger malignant nodules in LIDC-Mal, a few positive outliers from \emph{prediction-aware} and other intensity-based \Aggs{} still contribute meaningfully within \emph{GMM-All}, despite generally offset contributions from the remaining \Aggs{}. A similar pattern occurs for partly solid tumors in LIDC-Tex, although the effect of these outliers is milder. Consequently, the \emph{GMM-All} performance is somewhat lower in LIDC-Tex than in LIDC-Mal, as individual \emph{prediction-aware} \Aggs{} achieve high performance when fitted with a single GMM (bottom panel of \Cref{fig:gmm_stab_rebuttal}a). Finally, the absolute SHAP values help explain the relatively poor performance of the meta-aggregator on the LIZ dataset. For both \ood{} variants, LIZ-IG and LIZ-SG, we observe an offset in \Agg{} contributions, resulting from the presence of both high and low values for certain features in \iid{} and \ood{} samples. This variability arises from tissue folds, background noise, and variable cell positioning, independently of the distribution shift introduced by the different recording technique.




\begin{figure*}[ht]
    \centering
    \includegraphics[width=\linewidth]{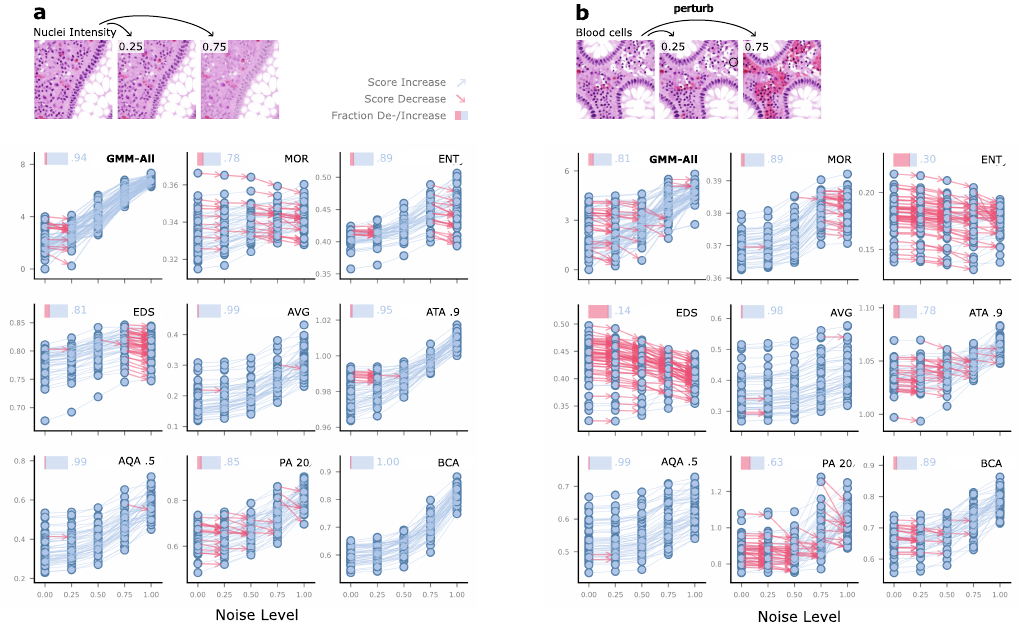}
    \caption{\textbf{\Aggs{} behavior under gradual perturbations.} For varying perturbation levels of a) ARC-Nuc and b) ARC-BC we display the sample-wise behavior of selected \Aggs{}, i.e. an increase with increasing perturbation parameter manipulation is marked as light blue connection, a decrease as red arrow. The overall fraction between de- and increases is inset in the upper left corner of each \Aggs{} plot.}
    \label{fig:mono_experiment}
\end{figure*}

\subsection{Gradual distribution shifts}
In this paper, we demonstrated that fitting a Gaussian Mixture Model (GMM) on multiple aggregation scores yields a robust \ood{} detection method. Our experiments were designed by pairing an \iid{} dataset with an \ood{} counterpart, showing that the GMM effectively distinguishes between the two in this binary setting. However, an ideal score should not only support binary classification but also capture gradual distribution shifts. \newline \noindent
To explore this, we used the synthetic ARC dataset, which enables the study of such progressive changes: ARC-Nuc involves perturbations of nuclei intensity, while ARC-BC involves an increased presence of blood cells (see Fig.~\ref{fig:mono_experiment}). Note, in our main analysis, we only focused on the ARC-Nuc 0.5 and ARC-BC 0.75 variations. \newline \noindent
Fig.~\ref{fig:mono_experiment} illustrates whether the aggregated scores increase or decrease in response to the respective perturbations. Notably, the AVG score shows almost strictly monotonic increases. However, its separation between perturbed and unperturbed samples is less pronounced than that of the GMM score for instance. While the GMM score does not increase perfectly monotonically, it still exhibits a high fraction of increases—0.94 for ARC-Nuc and 0.84 for ARC-BC. Also \eg the \Aggs{} exhibits promising behaviour in that regard.

\section{Details on Limitations}
\label{app:limitations}

\noindent\paragraph{Uncertainty Types}
While disentangling epistemic and aleatoric uncertainty is important—particularly for \ood{} detection—such a separation would primarily affect pixel-wise scores rather than the aggregated outputs. Moreover, the impact on failure detection (FD) performance is expected to be limited, as this task may be conducted on \iid{} test data only, depending on the scenario under consideration. Nevertheless, investigating how uncertainty type disentanglement influences aggregation remains an interesting direction for future work.

\noindent\paragraph{Diversity of datasets and \ood{} scenarios}
We aimed for maximal diversity in our datasets, focusing on realistic shifts following \citet{taori2020measuring}. Nevertheless, our selection is naturally non-exhaustive. Other types of \ood{} shifts, such as image augmentations (e.g., blurring) or new-class shifts in label space, could also be explored. Furthermore, our study focuses on the aggregation of uncertainty in 2D. While we expect many of our findings to generalize to 3D, additional challenges may arise in volumetric settings, such as when uncertainty is concentrated at object surfaces.

\noindent\paragraph{Diversity of \Aggs{}}
Our work emphasizes the most common intensity-based \Aggs{}. Future investigations could incorporate additional sources of information beyond prediction masks, such as ground truth annotations, error maps for calibration tasks, or auxiliary modalities (\eg depth). Another direction is the extension of the pipeline to unbounded logit-based scores, such as DDU~\cite{Mukhoti_2023_CVPR, mukhoti2018evaluating} or energy scores~\cite{liu2020energy}. However, this extension is non-trivial, as extreme pixel values can dominate aggregation and may require specifically tailored \Aggs{}. For the \Aggs{} explored in this study, hyperparameters were chosen based on standard practices in the literature; a more comprehensive parameter sweep could reveal more optimal settings.

\noindent\paragraph{Gaussian Mixture Models}
The GMM approach demonstrates the effectiveness of combining individual \Aggs{} for \ood{} and Failure Detection. In Supp. \ref{app:datapreproc}, we already discuss the potential issue arising from data transformations. Another prominent limitation occurs in low-data, high-dimensional settings, where a GMM may struggle to accurately fit the sample distribution.
This highlights the need for more robust \ood{} detection methods or a reduction in the number of aggregation measures in these particular settings. Alternatives to GMMs could be explored in future work, such as Vine Copulas~\cite{czado2022vine}, which can handle arbitrary distributions within the [0,1] range without requiring transformations and may provide greater robustness in a higher dimensional settings. \newline \noindent
Additionally, further studies could investigate ablations with fewer or better-chosen parameterized \Aggs{} (\eg PLM), potentially revealing an optimal dimensionality for the GMM and more stable performance in certain datasets (e.g., LIZ). For completeness, it would be interesting to compare our non-learned, interpretable meta-aggregator with \ood{} detectors based on learned representations, such as those obtained through self-supervised learning for modeling sample distributions. Our approach retains interpretability, as it relies on spatial and intensity-based scores whose individual meanings directly reflect the structure and magnitude of uncertainty.

\end{document}